\pdfoutput=1

\documentclass[journal]{IEEEtran}
\usepackage{graphicx}
\usepackage{comment}
\usepackage{amsmath,amssymb} % define this before the line numbering.
\usepackage{caption}
\usepackage{subfigure}
\usepackage{multirow}
\usepackage{booktabs}
\usepackage{verbatim}
\usepackage{color, colortbl}
\usepackage{xspace, xcolor}
\usepackage{cite}
\usepackage[export]{adjustbox}
\usepackage{dsfont}
\usepackage{physics}
\usepackage{threeparttable}
\usepackage{afterpage}
\usepackage{url}
\usepackage{xspace}
\def\eg{\textit{e.g.,}}
\def\ie{\textit{i.e.}}

\def\etal{\textit{et al. }}

\usepackage[normalem]{ulem}

\definecolor{Gray}{gray}{0.9}
\definecolor{Red}{RGB}{245,220,220}

\ifCLASSINFOpdf
\else
\fi
\begin{document}
%\title{Guided Filtering with Intercept Prior}
\title{Unsharp Mask Guided Filtering}

\author{Zenglin~Shi, %~\IEEEmembership{Member,~IEEE,}
        Yunlu~Chen, %~\IEEEmembership{Member,~IEEE,}
        Efstratios~Gavves, %~\IEEEmembership{Member,~IEEE,}
        Pascal~Mettes, %~\IEEEmembership{Member,~IEEE,}
        and~Cees~G.~M.~Snoek,~\IEEEmembership{Senior Member,~IEEE}% <-this % stops a space
\thanks{The authors are with the Informatics Institute
of the University of Amsterdam, Amsterdam, the Netherlands.}% <-this % stops a space
}


\markboth{IEEE Transactions on Image Processing,~Vol.~X, No.~Y, Month~2020}%
{Shell \MakeLowercase{\textit{et al.}}: Bare Demo of IEEEtran.cls for IEEE Journals}



\maketitle

\begin{abstract}
The goal of this paper is guided image filtering, which emphasizes the importance of structure transfer during filtering by means of an additional guidance image.
Where classical guided filters transfer structures using hand-designed functions, recent guided filters have been considerably advanced through parametric learning of deep networks. The state-of-the-art leverages deep networks to estimate the two core coefficients of the guided filter. In this work, we posit that simultaneously estimating both coefficients is suboptimal, resulting in halo artifacts and structure inconsistencies. 
Inspired by unsharp masking, a classical technique for edge enhancement that requires only a single coefficient, we propose a new and simplified formulation of the guided filter. Our formulation enjoys a filtering prior from a low-pass filter and enables explicit structure transfer by estimating a single coefficient. Based on our proposed formulation, we introduce a successive guided filtering network, which provides multiple filtering results from a single network, allowing for a trade-off between accuracy and efficiency. 
Extensive ablations, comparisons and analysis show the effectiveness and efficiency of our formulation and network, resulting in state-of-the-art results across filtering tasks like upsampling, denoising, and cross-modality filtering. 
Code is available at \url{https://github.com/shizenglin/Unsharp-Mask-Guided-Filtering}.

\end{abstract}

% \fi
\IEEEpeerreviewmaketitle

\section{Introduction}
%------------------------------------------------------------------------

\IEEEPARstart{I}{mage} filtering has been widely used to suppress unwanted signals (\eg~noise) while preserving the desired ones (\eg~edges) in image processing tasks like image restoration \cite{elad1999superresolution,banham1996spatially,awate2006unsupervised}, boundary detection \cite{ma2000edgeflow, kang2012lidar, jacob2004design}, texture segmentation \cite{dunn1995optimal,weldon1996efficient,randen1999texture}, and image detail enhancement \cite{aysal2006quadratic, guo2016lime, min2011depth}. Standard filters, such as Gaussian filters and box mean filters, swiftly process input imagery but suffer from content-blindness, \ie,~they treat noise, texture, and structure identically. To mitigate content-blindness, guided filters \cite{he2012guided,kopf2007joint,pham2011adaptive,liu2013joint,gharbi2017deep,xie2015edge}, have received a great amount of attention from the community. The key idea of guided filtering is to leverage an additional guidance image as a structure prior and transfer the structure of the guidance image to a target image. By doing so, it strives to preserve salient features, such as edges and corners, while suppressing noise. The goal of this paper is guided image filtering. 

\begin{figure*}[t!]
\centering 
\includegraphics[width=\linewidth]{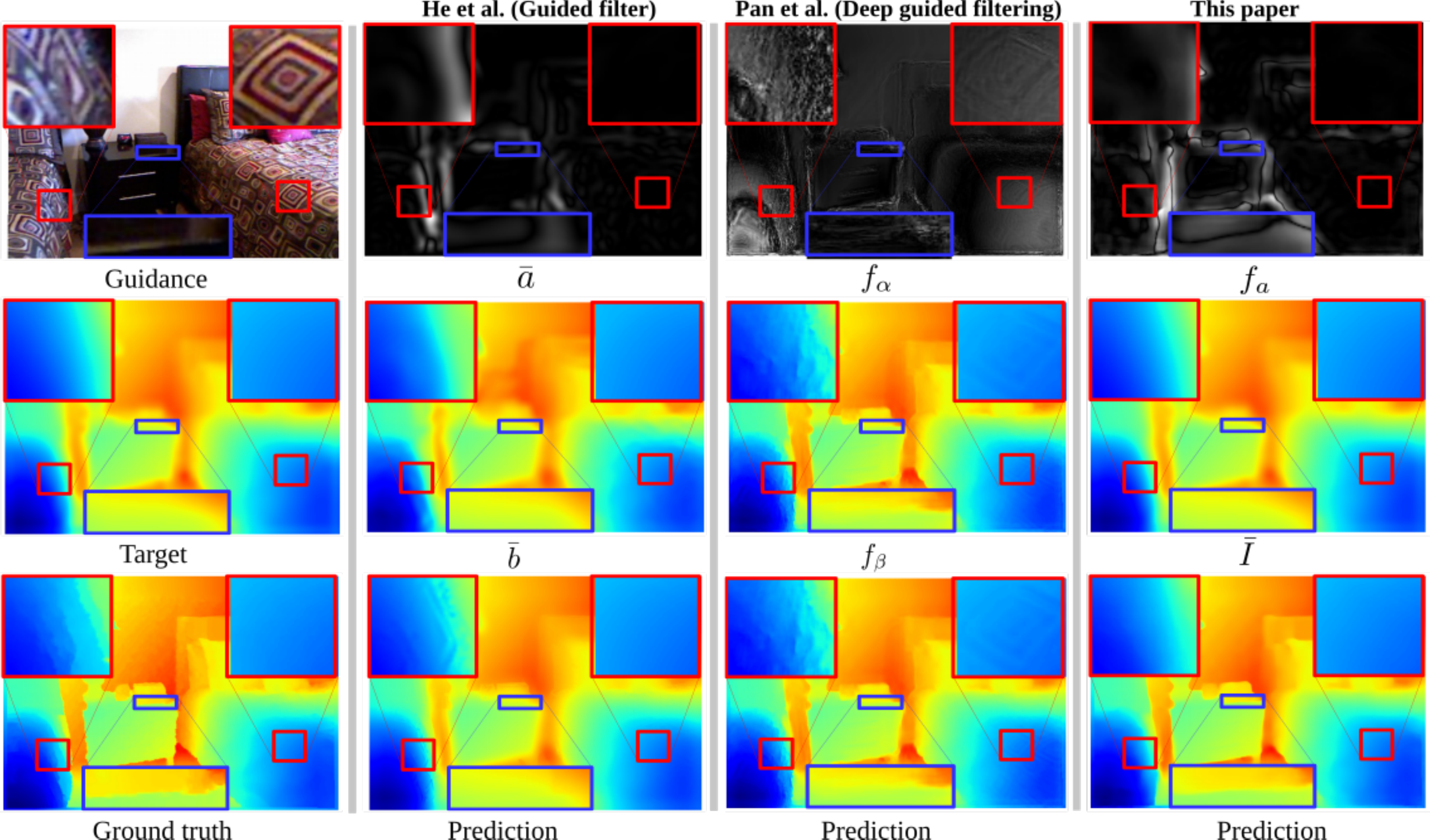}
\caption{\textbf{Motivation of this paper.} We show an example of depth upsampling ($16 \times$) using an RGB image as guidance. Both the conventional guided filter by He \etal \cite{he2012guided} and the state-of-the-art deep guided filter by Pan \etal \cite{pan2019spatially} explicitly estimate two coefficients, respectively $(\bar{a},\bar{b})$ and $(f_{\alpha},f_{\beta})$. In their current formulation, however, both methods are likely to over-smooth edges (compare edges in blue boxes) and transfer unwanted textures (compare highlighted details in red boxes). Our proposed guided filter, taking inspiration from unsharp masking, only requires learning a single coefficient $f_a$ (notation details provided in Sections 2 and 3). As a result, we obtain a more desirable upsampling result, free of undesirable structures and textures from the guidance image. 
}
\label{exp_fig_motivation}    
\end{figure*}

Classical guided filtering, \eg~\cite{he2012guided,kou2015gradient,li2014weighted,sun2019weighted}, performs structure-transferring by relying on hand-designed functions. Nonetheless, it is known to suffer from halo artifacts and structure inconsistency problems (see Fig. \ref{exp_fig_motivation}), and it may  require a considerable computational cost. In recent years, guided image filtering has advanced by deep convolutional neural networks. Both Li \etal \cite{li2016deep} and Hui \etal \cite{hui2016depth} demonstrate the benefits of learning-based guided filtering over classical guided filtering. These works and their follow-ups, \eg~\cite{AlBahar_2019_ICCV,Su_2019_CVPR,li2019joint}, directly predict the filtered output by means of feature fusion from the guidance and target images. Yet, this implicit way of structure-transferring may fail to transfer the desired edges and may suffer from transferring undesired content to the target image \cite{li2019joint,pan2019spatially}. 

Pan \etal \cite{pan2019spatially} propose an alternative way to perform deep guided filtering. Rather than directly predicting the filtered image, they leverage a shared deep convolutional neural network to estimate the two coefficients of the original guided filtering formulation \cite{he2012guided}. 
While their approach obtains impressive filtering results in a variety of applications, we observe their network has difficulty disentangling the representations of the two coefficients, resulting in halo artifacts and structure inconsistencies, see Fig. \ref{exp_fig_motivation}.
Building on the work of Pan \etal \cite{pan2019spatially}, we propose a new guided filtering formulation, which depends on a \textit{single} coefficient and is therefore more suitable to be solved by a single deep convolutional neural network. 

We take inspiration from another classical structure-transferring filter: unsharp masking \cite{morishita1988unsharp,polesel2000image,deng2010generalized,ye2018blurriness}. From the original guided filter by He \etal \cite{he2012guided} we first derive a simplified guided filtering formulation by eliminating one of its two coefficients. So there is only one coefficient left to be estimated for deciding how to perform edge enhancement, akin to unsharp masking. To arrive at our formulation, we rely on the filtering prior used in unsharp masking and perform guided filtering on the unsharp masks rather than the raw target and guidance images themselves. The proposed formulation enables us to intuitively understand how the guided filter performs edge-preservation and structure-transferring, as there is only one coefficient in the formulation, rather than two in \cite{pan2019spatially}. The coefficient explicitly controls how structures need to be transferred from guidance image to target image and we learn it adaptively by a single network. To that end, we introduce a successive guided filtering network. The network obtains multiple filtering results by training a single network. It allows a trade-off between accuracy and efficiency by choosing different filtering results during inference. This leads to fast convergence and improved performance. Experimental evaluation on seven datasets shows the effectiveness of our proposed filtering formulation and network on multiple applications, such as upsampling, denoising, and cross-modality filtering.

\section{Background and Related Work}
\label{sec:related}
In guided filtering, we are given an image pair $(I, G)$, where the image pair has been properly registered by default.
Image $I$ needs to be filtered, \eg~due to the presence of noise or due to low resolution because it is captured by a cheap sensor. Guidance image $G$ contains less noise and is of high resolution with sharp edges, \eg~because it is captured by accurate sensors under good light conditions. The low-quality image $I$ can be enhanced by filtering under the guidance of high-quality image $G$. Such a guided filtering process is defined as $\hat{I} = \mathcal{F}(I, G)$, where $\mathcal{F}(\cdot)$ denotes the filter function and $\hat{I}$ denotes the filtered output image. Below, we first review the benefits and weaknesses of classical guided filter functions and existing deep guided filter functions. We then present how our proposal can be an improved guided filtering solution by establishing a link to unsharp masking.   

\subsection{Classical guided filtering}
The guided image filter \cite{he2012guided} assumes that the filtered output image $\hat{I}$ is a linear transform of the guidance image $G$ at a window $w_k$ centered at pixel $k$:
\begin{equation}
\hat{I}_i = a_kG_i+b_k, \quad \forall i \in w_k, 
\label{eq_local}
\end{equation}
where $a_k$ and $b_k$ are two constants in window $w_k$. Their values can be obtained by solving:
\begin{equation}
E(a_k, b_k) = \sum_{i \in w_k}((a_kG_i+b_k-I_i)^2+ \epsilon a_k^2).
\label{eq_cost}
\end{equation}
Here, $\epsilon$ is a regularization parameter penalizing large values for $a_k$. The optimal values of $a_k$ and $b_k$ are computed as:
\begin{equation}
a_k = \frac{\frac{1}{|w|}\sum_{i\in w_k}I_iG_i-\bar{I}_k\bar{G}_k}{\sigma _k^2+\epsilon},
\label{eq_a}
\end{equation}
\begin{equation}
b_k = \bar{I}_k-a_k\bar{G}_k.
\label{eq_b}
\end{equation}
Here, $\bar{G}_k$ and $\sigma _k^2$ are the mean and variance of $G$ in $w_k$, $\bar{I}_k$ is the mean of $I$ in $w_k$, and $|w|$ is the number of pixels in $w_k$. 
For ease of analysis, the guidance image $G$ and filtering target image $I$ are assumed to be the same \cite{he2012guided}, although the general case remains valid. 
As a result, we can obtain:
\begin{equation}
a_k = \sigma _k^2/(\sigma _k^2+\epsilon), \quad b_k=(1-a_k)\bar{G}_k.
\label{eq_ab}
\end{equation}
Based on this, the regions and edges with variance ($\sigma _k^2$) much larger than $\epsilon$ are preserved, whereas those with variance ($\sigma _k^2$) much smaller than $\epsilon$ are smoothed. Hence, $\epsilon$ takes control of the filtering. However, the value of $\epsilon$ in the guided image filter \cite{he2012guided} is fixed. As such, halos are unavoidable when the filter is forced to smooth some edges \cite{he2012guided,kou2015gradient,li2014weighted,sun2019weighted}. An edge-aware weighted guided image filter is proposed in \cite{li2014weighted} to overcome this problem, where $\epsilon$ varies  spatially rather than being fixed:
\begin{equation}
\epsilon = \frac{ \lambda }{\Gamma_G}, \quad \Gamma_G = \frac{ \sigma _{k}^2 + \varepsilon}{\frac{1}{N}\sum_{k=1}^N \sigma _k^2 + \varepsilon},
\end{equation}
where $\lambda$ and $\varepsilon$ are two small constants and $\sigma _k^2$ is the variance of $G$ in a $3 \times 3$ window centered at the pixel $k$. $\Gamma_G$ measures the importance of a pixel $k$ with respect to the whole guidance image. Kou~\etal \cite{kou2015gradient} propose a multi-scale edge-aware weighted guided image filter, in which $\Gamma_G$ is computed by multiplying the variances of multiple windows. By taking the edge direction into consideration, Sun \etal~\cite{sun2019weighted} further improve the filter’s performance. However, predefined parameters still remain in all these three methods. 

Another limitation of classical guided image filters is their assumption that the target image and the guidance image have the same structure. In practice, there are also situations conceivable where an edge appears in one image, but not in the other. To address this issue, recent works \cite{shen2015mutual, Jevnisek_2017_CVPR, guo2018mutually, ham2018robust,Yin_2019_CVPR} enhance guided image filters by considering the mutual structure information in both the target and the guidance images. These methods typically build on iterative algorithms that minimize global objective functions. The guidance signals are updated at each iteration to enforce the outputs to have similar structure as the target images.
These global optimization methods generally use hand-crafted objectives that usually involve fidelity and regularization terms. The fidelity term captures the consistency between the filtering output and the target image. The regularization term, typically modeled using a weighted L2 norm \cite{farbman2008edge}, encourages the filtering output to have a similar structure as the guidance image. However, such hand-crafted regularizers may not transfer the desired structures from the guidance image to the filtering output \cite{li2019joint,pan2019spatially}.
For this reason we prefer a guided filtering solution based on end-to-end deep representation learning.

\subsection{Deep guided filtering}
Li \etal \cite{li2016deep,li2019joint} introduce the concept of deep guided image filtering based on an end-to-end trainable network. Two independent convolutional networks $f_I$ and $f_G$ first process the target image and guidance image separately. Then their outputs from the last layers are concatenated and forwarded to another convolutional network $f_{IG}$ to generate the final output. We define the method as:
\begin{equation}
\hat{I} = f_{IG}\big(f_I(I) \oplus f_G(G)\big),
\label{eq_djf}
\end{equation}
where $\oplus$ denotes the channel concatenation operator. Several works follow the same definition but vary in their feature fusion strategies \cite{AlBahar_2019_ICCV,hui2016depth,Su_2019_CVPR}. AlBahar \etal \cite{AlBahar_2019_ICCV} introduce a bi-directional feature transformer to replace the concatenation operation. Su \etal \cite{Su_2019_CVPR} propose a pixel-adaptive convolution to fuse the outputs of networks $f_I$ and $ f_G$ adaptively on the pixel level. Hui \etal \cite{hui2016depth} propose to perform multi-scale guided depth upsampling based on spectral decomposition, where low-frequency components of the target images are directly upsampled by bicubic interpolation and the high-frequency components are upsampled by learning two convolutional networks. The high-frequency domain learning leads to an improved performance.
Wu \etal \cite{wu2018fast} alternatively combine convolutional networks and traditional guided image filters. Two independent convolutional networks $f_I$ and $ f_G$ first amend the target image and the guidance image, and then feed their outputs into the traditional guided image filter $\mathcal{F}_{IG}$ \cite{he2012guided}: 
\begin{equation}
\hat{I} = \mathcal{F}_{IG} \big(f_I(I), f_G(G)\big).
\label{eq_tgf}
\end{equation}
Rather than directly predicting the filtered image, Pan \etal \cite{pan2019spatially} leverage deep neural networks to estimate the two coefficients of the original guided filtering formulation \cite{he2012guided} based on a spatially-variant linear representation model, leading to impressive filtering results. It is defined as:
\begin{equation}
\hat{I} = f_{\alpha}(I,G) \odot G + f_{\beta}(I,G),
\label{eq_svlrm}
\end{equation}
where $f_{\alpha}$ and $f_{\beta}$ are two convolutional networks, and $\odot$ denotes element-wise multiplication.
To reduce the complexity of learning, in the implementation Pan \etal \cite{pan2019spatially} learn a single network and predict an output with two channels, one channel for $f_{\alpha}$ and another channel for $f_{\beta}$.
However, we observe that the shared network has difficulty disentangling the representations of the two coefficients, resulting in halo artifacts and structure inconsistencies. Differently, we propose a new guided filtering formulation, which depends on a \textit{single} coefficient only and is therefore more suitable to be solved by a single deep convolutional neural network.

The aforementioned existing deep guided filtering works implicitly perform structure-transferring by learning a joint filtering network, usually resulting in undesired filtering performance~\cite{pan2019spatially,deng2020deep,marivani2020multimodal}. Recent works~\cite{marivani2019learned,marivani2020joint,marivani2019multimodal,marivani2020multimodal,marivani2021designing,deng2020deep,deng2019deep} take inspiration from coupled dictionary learning~\cite{song2019multimodal}, and incorporate sparse priors into their deep networks for explicit structure-transferring. Marivani \etal \cite{marivani2019learned,marivani2020joint,marivani2019multimodal,marivani2020multimodal} propose a learned multimodal convolutional sparse coding network with a deep unfolding method for explicitly fusing information from the target and guidance image modalities. Deng \etal propose a deep coupled ISTA network with a multimodal dictionary learning algorithm~\cite{deng2019deep}, and a common and unique information splitting network with multi-modal convolutional sparse coding~\cite{deng2020deep}, for the sake of explicitly modeling the knowledge from the guidance image modality. Most of these works focus on guided image super-resolution. In this work, we propose an explicit structure-transferring method for general guided filtering problems. In particular, we propose a guided filtering formulation with a single coefficient, and we learn to estimate the coefficient to explicitly decide how to transfer the desirable structures from guidance image to target image. As we will demonstrate, this leads to more desirable filtering results.
\subsection{Unsharp masking}
Our formulation is inspired by the classical sharpness enhancement technique of unsharp masking \cite{morishita1988unsharp,polesel2000image,deng2010generalized,ye2018blurriness}, which can be described by the equation:
\begin{equation}
\hat{I} = \lambda (I-\mathcal{F}_L(I)) + I,
\label{eq_unsharp}
\end{equation}
where an enhanced image is represented by $\hat{I}$, an original image by $I$, an unsharp mask by $(I-\mathcal{F}_L(I))$ where $\mathcal{F}_L$ denotes a low-pass filter like Gaussian filters or box mean filters, and an amount coefficient by $\lambda $ which controls the volume of enhancement achieved at the output.
Essentially, guided filtering shares the same function of edges enhancement as unsharp masking by means of the structure-transferring from an additional guidance image. Based on this viewpoint, we derive a simplified guided filtering formulation from the original guided filter \cite{he2012guided},
%where there is
with only one coefficient to be estimated, akin to the formulation of unsharp masking in Eq.~(\ref{eq_unsharp}). 
\section{Filtering formulation}
\label{sec:formulation}

Here, we outline our new guided filtering formulation, in which only one coefficient needs to be estimated. Compared to estimating two coefficients ($a$, $b$) as in the original guided filtering formulation and subsequent deep learning variants, our formulation is more suitable to be solved with one single deep network.
We start the derivation of our guided filtering formulation from the classical guided filter \cite{he2012guided}, summarized in Eq. (\ref{eq_local}), Eq. (\ref{eq_a}) and Eq. (\ref{eq_b}).
In Eq. (\ref{eq_local}), $\hat{I}$ is a linear transform of $G$ in a window $w_k$ centered at the pixel $k$. When we apply the linear model to all local windows in the entire image, a pixel $i$ is involved in all the overlapping windows $w_k$ that covers $i$. In this case, the value of $\hat{I}_i$ in Eq. (\ref{eq_local}) is not identical when it is computed in different windows. So after computing $(a_k,b_k)$ for all windows $w_k$ in the image, we compute the filtered output image $\hat{I}_i$ by averaging all the possible values of $\hat{I}_i$ with:
\begin{equation}
\hat{I}_i = \frac{1}{|w|}\sum_{k\in w_i}(a_kG_i+b_k).
\label{eq_local_mean}
\end{equation}
Similar in spirit to unsharp masking, summarized in Eq. (\ref{eq_unsharp}), we want to maintain only the coefficient $a$ to control the volume of structure to be transferred from guidance $G$ to the filtered output image $\hat{I}$. Thus, we put Eq. (\ref{eq_b}) into Eq. (\ref{eq_local_mean}) to  eliminate $b$, and obtain:
\begin{equation}
\hat{I}_i = \frac{1}{|w|}\sum_{k\in w_i}a_kG_i+\frac{1}{|w|}\sum_{k\in w_i}(\bar{I}_k-a_k\bar{G}_k).
\label{eq_new_form1}
\end{equation}
Next, we rewrite the formulation as:
\begin{equation}
\hat{I}_i = \frac{1}{|w|}\sum_{k\in w_i}a_k(G_i-\bar{G}_k)+\widetilde{I}_i,
\label{eq_new_form2}
\end{equation}
where $\widetilde{I}_i=\frac{1}{|w|}\sum_{k\in w_i}\bar{I}_k$. Since $\bar{G}_k$ is the output of a mean filter, it's assumed that $\bar{G}_k$ is close to its mean in the window $w_i$. Next we rewrite Eq. (\ref{eq_new_form2}) as follows
\begin{equation}
\hat{I}_i = \bar{a}_i(G_i-\widetilde{G}_i)+\widetilde{I}_i,
\label{eq_new_form3}
\end{equation}
where $\bar{a}_i=\frac{1}{|w|}\sum_{k\in w_i}a_k$, and $\widetilde{G}_i=\frac{1}{|w|}\sum_{k\in w_i}\bar{G}_k$. For convenience, we will omit subscript $i$ in the following. 

The formulation in Eq. (\ref{eq_new_form3}) enables us to intuitively understand how the guided filter performs edge-preservation and structure-transferring. Specifically, the target image $I$ is first smoothed to remove unwanted components like noise/textures, and the smoothing result is denoted by $\widetilde{I}$. However, the smoothing process usually suffers from the loss of sharp edges, leading to a blurred output. To enhance the edges, an unsharp mask $(G-\widetilde{G})$ with fine edges generated from the guidance image $G$ is added to $\widetilde{I}$ under the control of the coefficient $a$, leading to the structure being transferred from the guidance image to the filtered output image $\hat{I}$. Finally, we rewrite Eq. (\ref{eq_new_form3}) to obtain a more general formulation for deep guided filtering: 
\begin{equation}
\hat{I} = f_a(I_m,G_m) \odot G_m+\mathcal{F}_L(I),
\label{eq_new_form4}
\end{equation}
where $I_m=I-\mathcal{F}_L(I)$ and $G_m=G-\mathcal{F}_L(G)$ denote the unsharp masks of the target image and the guidance image,
which contain the structures of the guidance and the target images. $\mathcal{F}_L$ denotes a linear shift-invariant low-pass filter like the Gaussian filter or the box mean filter. $f_a$ denotes the amount function, which controls the volume of structure to be transferred from the guidance image to the filtered output image. Next, we will elaborate on this function. 

\textbf{Amount function $f_a$.}
The output of $f_a$ is the volume of the structure to be transferred from the guidance image to the filtered output image. Thus, the input of $f_a$ should involve the structure of both the target and the guidance image, which together determine the output, \ie, $f_a(I_m,G_m)$.
Ideally, $f_a$ should determine the structure-transferring in a pixel-adaptive fashion. It can be a manually designed function as the function $a$ of the guided filter in Eq. (\ref{eq_a}). It also can be estimated by learning a deep neural network. Compared to hand-crafted functions, learnable functions are more flexible and allow for a better generalization to various image conditions. 

\textbf{Successive filtering.} Successive operations of the same filter generally result in a more desirable output, thus we develop a successive guided filtering based on our formulation in Eq. (\ref{eq_new_form4}). Instead of directly iterating the filtering output $\hat{I}$, we iterate the outputs of $f_a$ as the function decides the effect of filtering. Let $f_a^\star$ be a composition of a set of basic functions $\{f_a^{(l)}\}_{l=1}^{L}$: 
\begin{equation}
 f_a^{\star} =f_a^{(L)} \circ f_a^{(L-1)} \circ \cdots \circ f_a^{(1)}  
 \label{eq_new_form5}
\end{equation}
in which $\circ$ denotes the function composition operation, such as $\big(f \circ u \big) (\cdot)= f(u(\cdot))$. With $f_a^\star$ we obtain a successive filtering formulation from Eq. (\ref{eq_new_form4}),
\begin{equation}
\hat{I} = f^\star_a(I_m,G_m) \odot G_m + \mathcal{F}_L(I). 
\label{eq_successive}
\end{equation}
In the next section we will detail how to implement our filters with deep convolutional neural networks.

\section{Filtering network}
\label{sec:network}
There is a function $f_a$ in our formulation, which governs a guided filtering coefficient. We propose to solve this function with a single convolutional neural network. 

\textbf{Network for amount function $f_a$.} The function $f_a$ decides how to transfer the structure of the guidance image to the filtered output image. There are two inputs $G_m$ and $I_m$ for this function. Two options are available for the network architecture. Like \cite{li2019joint,hui2016depth}, we can separately process these two inputs with two different sub-networks at first, and then fuse their outputs with another sub-network. Alternatively, we can concatenate these two inputs together and forward the concatenation into a single network, similar to the framework of \cite{pan2019spatially}. Empirically, we find that the second option is easily implemented and achieves a desirable filtering accuracy and efficiency. Thus, we design the network of $f_a$ with the second option in this work. 

In our approach the unsharp masks, rather than the raw images themselves, are used as the inputs of the network. The unsharp mask is more sparse than the raw image itself, since most regions in the unsharp mask are close to zero. The learning of spatially-sparse data usually requires a large receptive field. The dilated convolution is a popular technique to enlarge the receptive fields in convolutional networks \cite{yu2015multi,chen2017deeplab}. We design our network by cascading four dilated convolutional layers with increasing dilation factors, where all the dilated convolutional layers have the same channel number of 24 and the same kernel size of $3 \times 3$. Their dilation factors are set to \{1, 2, 4, 8\}. Leaky ReLU with a negative slope of $0.1$ is used as an activation function after each dilated convolutional layer. Finally, a $1 \times 1$ convolution layer generates the target output for $f_a$. 

\begin{figure}[t]
\centering 
\subfigure[Amount block]{
\begin{minipage}{8.5cm}
\centering
\includegraphics[width=0.98\linewidth,left]{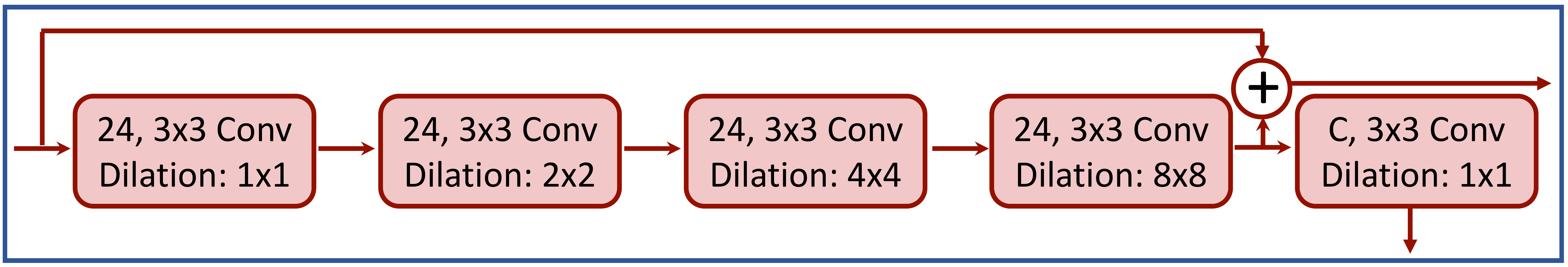}
\end{minipage}
}
\subfigure[Successive guided filtering network]{
\begin{minipage}{8.5cm}
\centering
\includegraphics[width=0.98\linewidth,left]{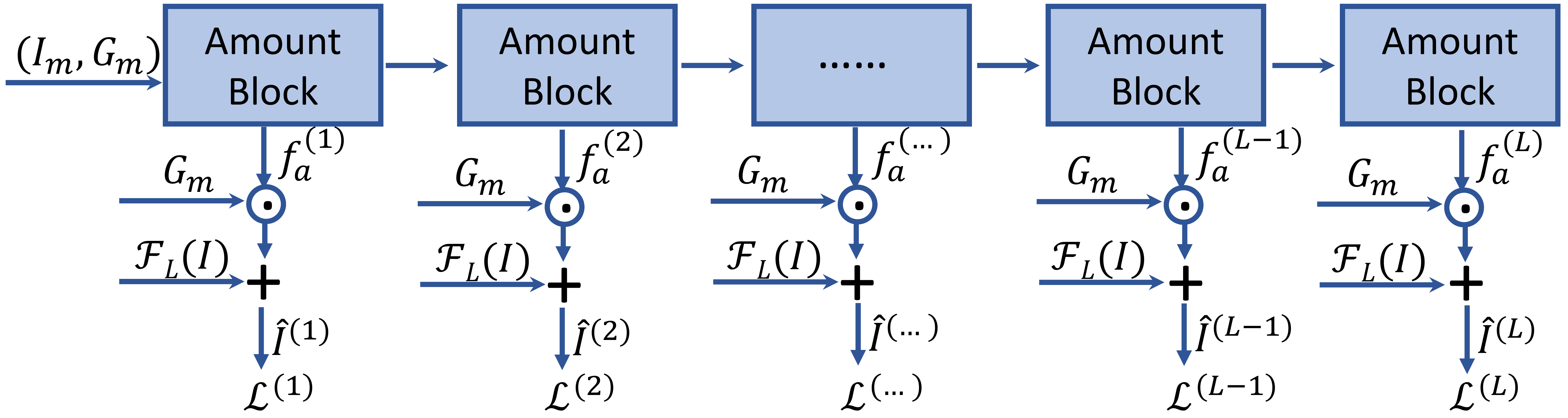}
\end{minipage}
}
\caption{\textbf{Network architecture for unsharp-mask guided filtering}. a) Dilated convolutional block for amount function $f_a$; b) Network architecture for successive unsharp-mask guided filtering. Here, $\odot$ denotes the element-wise product, $\oplus$ denotes the concatenation operation, $+$ denotes the element-wise sum. Leaky ReLU is used as activation function after each convolutional layer. There are $(L-4)$ amount blocks in the box, indicated with dots. $\mathcal{L}$ denotes the loss function in Eq. (\ref{equ_loss}). Here, $G_m$ and $\mathcal{F}_L(I)$ are shared by $\{f_a^{(l)}\}_{l=1}^{L}$.}
\label{fig_network}  
\end{figure}

\textbf{Network for successive filtering.} To develop a network for the successive filtering formulation in Eq. (\ref{eq_successive}), we consider the network of the basic functions $\{f_a^{(l)}\}_{l=1}^{L}$ as a convolutional block, as shown in Fig. \ref{fig_network} (a). Then stacking this block multiple times results in a deep network for $f^\star_a$. There are two outputs in the block. By concatenating its input and its feature maps from the last layer results in the first output. The concatenation output allows feeding the previous multi-level outputs to the following blocks, leading to improved performance. We develop the second output by using a convolutional layer on top of the last layer of the block, for the sake of balancing accuracy and efficiency.

Stacking more blocks results in higher accuracy at the expense of an increased computational complexity. To allow users to choose between accuracy and computational complexity, we generate filtering outputs from each block. If users want to obtain filtering results fast, the filtering results from the first blocks can be used. When the accuracy is more important, the filtering results from the later blocks can be used. In short, we obtain multiple filtering results while training one single network. The overall network architecture is visualized in Fig.~\ref{fig_network} (b).

\textbf{Optimization.} During training, we are given $N$ samples $\{(I_n, G_n, Z_n)\}_{n=1}^{N}$, with $I_n \in \mathcal{I}$ the target image, $G_n \in \mathcal{G}$ the guidance image and $Z_n \in \mathcal{Z}$ the task-dependent ground-truth output image. Our goal is to learn the parameters $\theta$ of the network for $f_a$. Two types of loss, $L_1$ loss and $L_2$ loss, have been widely used in deep guided filtering works. Early works, like~\cite{li2016deep,hui2016depth}, have adopted a $L_2$ loss, while recent works, like~\cite{pan2019spatially,AlBahar_2019_ICCV}, prefer a $L_1$ loss because it is less sensitive to outliers and leads to less blurry results compared to a $L_2$ loss \cite{belagiannis2015robust,isola2017image}. Following these recent works, we minimize the difference between filtered output image $\hat{I}$ and its ground-truth $Z$ using the $L_1$ loss, which is defined as: 
\begin{equation}
 \begin{split}
\mathcal{L}(I,G,Z; \theta) = 
\frac{1}{N} \sum_{n=1}^{N}\parallel \hat{I}_n(I_n,G_n; \theta)-Z_n \parallel_{1}.
\end{split}
\label{equ_loss}
\end{equation}

\section{Experiments}
In this section, we provide extensive experimental evaluations. Section~\ref{sec:setup} introduces the experimental setup. Sections~\ref{sec:exp1},~\ref{sec:exp2},~\ref{sec:exp3}, and~\ref{sec:exp4} emphasize ablations, comparisons and analysis. Sections~\ref{sec:exp5},~\ref{sec:exp6}, and~\ref{sec:exp7} show further qualitative and quantitative results, and state-of-the-art comparisons on various applications, including upsampling, denoising, and cross-modality filtering.

\subsection{Experimental setup}
\label{sec:setup}
%\subsubsection{Datasets} 

\textbf{Image upsampling datasets.} We perform upsampling experiments on NYU Depth V2 \cite{silberman2012indoor}, and Sintel optical flow \cite{butler2012naturalistic}. 
For NYU Depth V2 we follow \cite{li2019joint}. We use the first 1000 RGB/depth pairs for training and the remaining 449 pairs for testing, where each low-resolution depth image is generated by applying a nearest-neighbor interpolation.
For Sintel, following \cite{Su_2019_CVPR}, 908 samples, and 133 samples from clean pass are used for training, and testing, where each low-resolution flow image is generated by applying a bilinear interpolation. 

\textbf{Image denoising datasets.} We perform denoising experiments also on NYU Depth V2 \cite{silberman2012indoor}, as well as on BSDS500 \cite{martin2001database}. For NYU Depth V2, we use the same split as for upsampling. BSDS500 contains 500 natural images. We train on the training set (200 images) and the validation set (100 images). We evaluate on the provided test set (200 images). Following \cite{zhang2017beyond}, to train a blind Gaussian denoiser, random Gaussian noise is added to the clean training depth images in NYU Depth V2 and the clean RGB images in BSDS500, with a noise level $\sigma \in [0,55]$. For testing, we consider three noise levels, $\sigma= \{15, 25, 50\}$. Thus, three separate noisy test images are generated for each original test image.

\textbf{Pre-processing.} For all datasets, we normalize the input images by scaling their pixel values to the range $[0,1]$. According to Eq. (\ref{eq_new_form4}), the guidance image $G$ and target image $I$ should have the same number of channels. In the depth/RGB dataset, the target is the 1-channel depth image. Thus, rgb2gray operation is used to transform a 3-channel RGB image into a 1-channel grayscale image as guidance.
During training, we augment the images by randomly cropping $256 \times 256$ patches. No cropping is performed during testing. 

\textbf{Network and optimization.}
The proposed network is optimized in an end-to-end manner. We implement the network with TensorFlow on a machine with a single GTX 1080 Ti GPU. The optimizer is Adam with a mini-batch of 1. We set $\beta_1$ to 0.9, $\beta_2$ to 0.999, and the initial learning rate to 0.0001. Optimization is terminated after 1000 epochs. %We will make our code publicly available.

\textbf{Evaluation metrics.}
To evaluate the quality of the predicted images we report four standard metrics: RMSE (Root Mean Square Error) for depth upsampling, EPE (End-Point-Error) for flow upsampling, PSNR (Peak Signal-to-Noise Ratio) for denoising, and SSIM (Structural Similarity Index Measure) for all applications.

%%%%%%%%%%%%%%%%%%%%%%%%
% RESULTS
%%%%%%%%%%%%%%%%%%%%%%%%
\subsection{Unsharp-mask guided filtering without learning.}
\label{sec:exp1}
The goal of the first experiment is to validate our formulation as a valid guided filter. Here we do not rely on any deep learning for estimating $f_a$. Instead, we use the function $a$ of the guided filter \cite{he2012guided} and the weighted guided filter (WGF) \cite{li2014weighted} as $f_a$. $\mathcal{F}_L(G)$ and $\mathcal{F}_L(I)$ are generated by cascaded box mean filters. Using our formulation as a conventional guided filter, we provide qualitative and quantitative results to demonstrate that our filter performs as good as, or even better than the guided filter \cite{he2012guided} and the weighted guided filter \cite{li2014weighted}. 

\begin{figure}[t!]
\centering 
\subfigure[\textbf{Input $I$($G$)}]{
\begin{minipage}{2.3cm}
\centering
\includegraphics[width=1.1\linewidth,left]{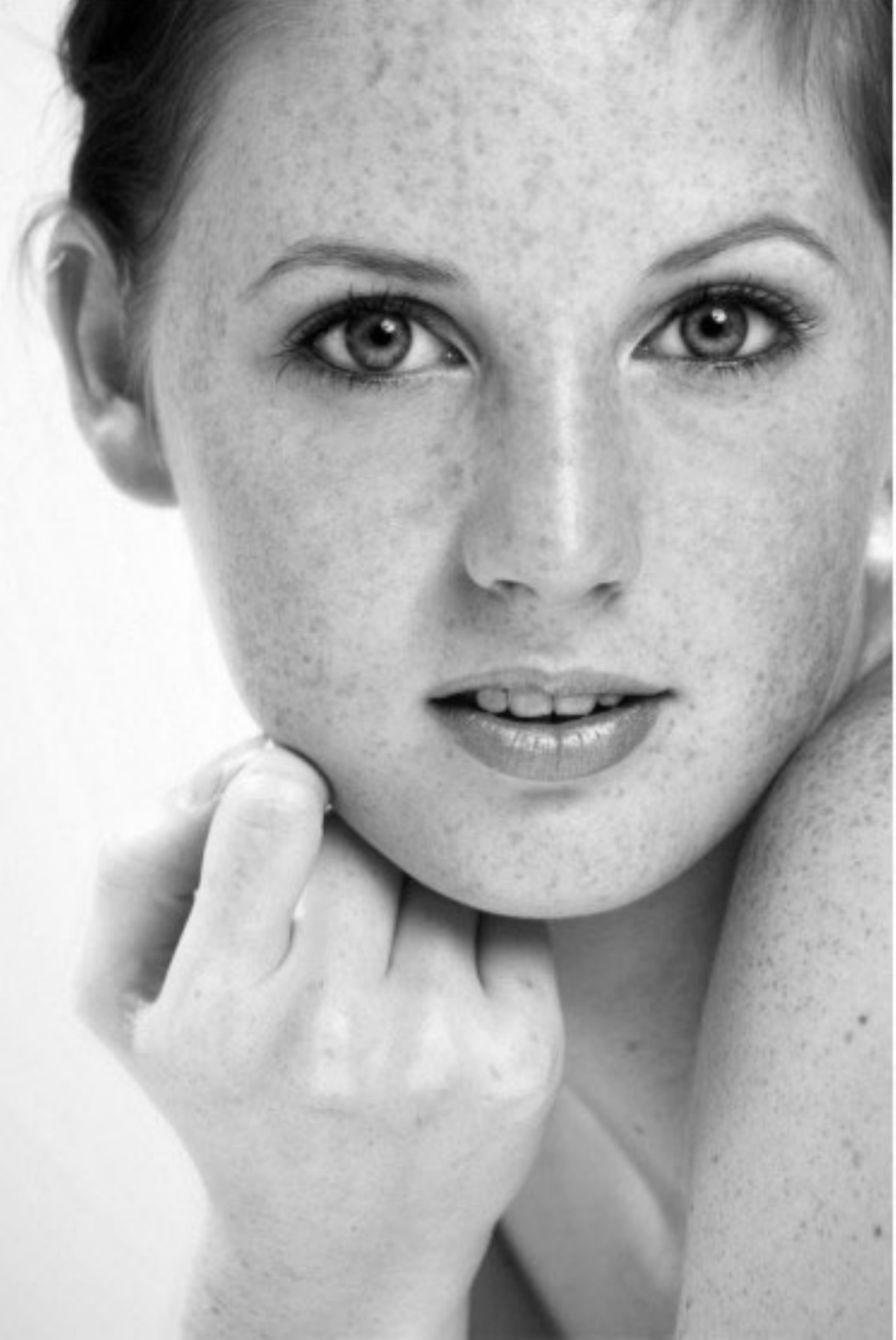}
\end{minipage}
}
\subfigure[\textbf{$G_m$}]{
\begin{minipage}{2.3cm}
\centering
\includegraphics[width=1.1\linewidth,left]{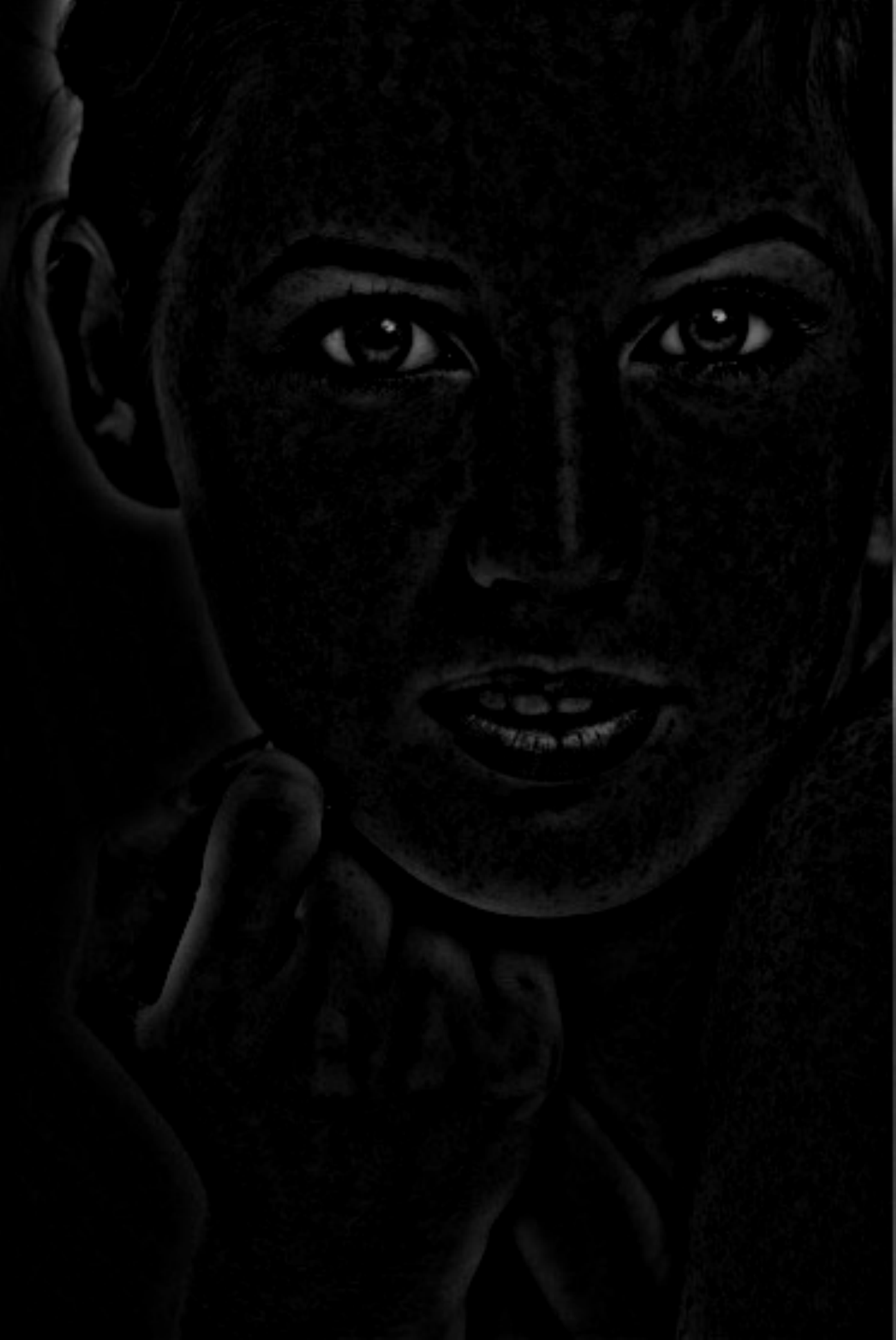}
\end{minipage}
}
\subfigure[\textbf{$f_a(I_m,G_m)$}]{
\begin{minipage}{2.3cm}
\centering
\includegraphics[width=1.1\linewidth,left]{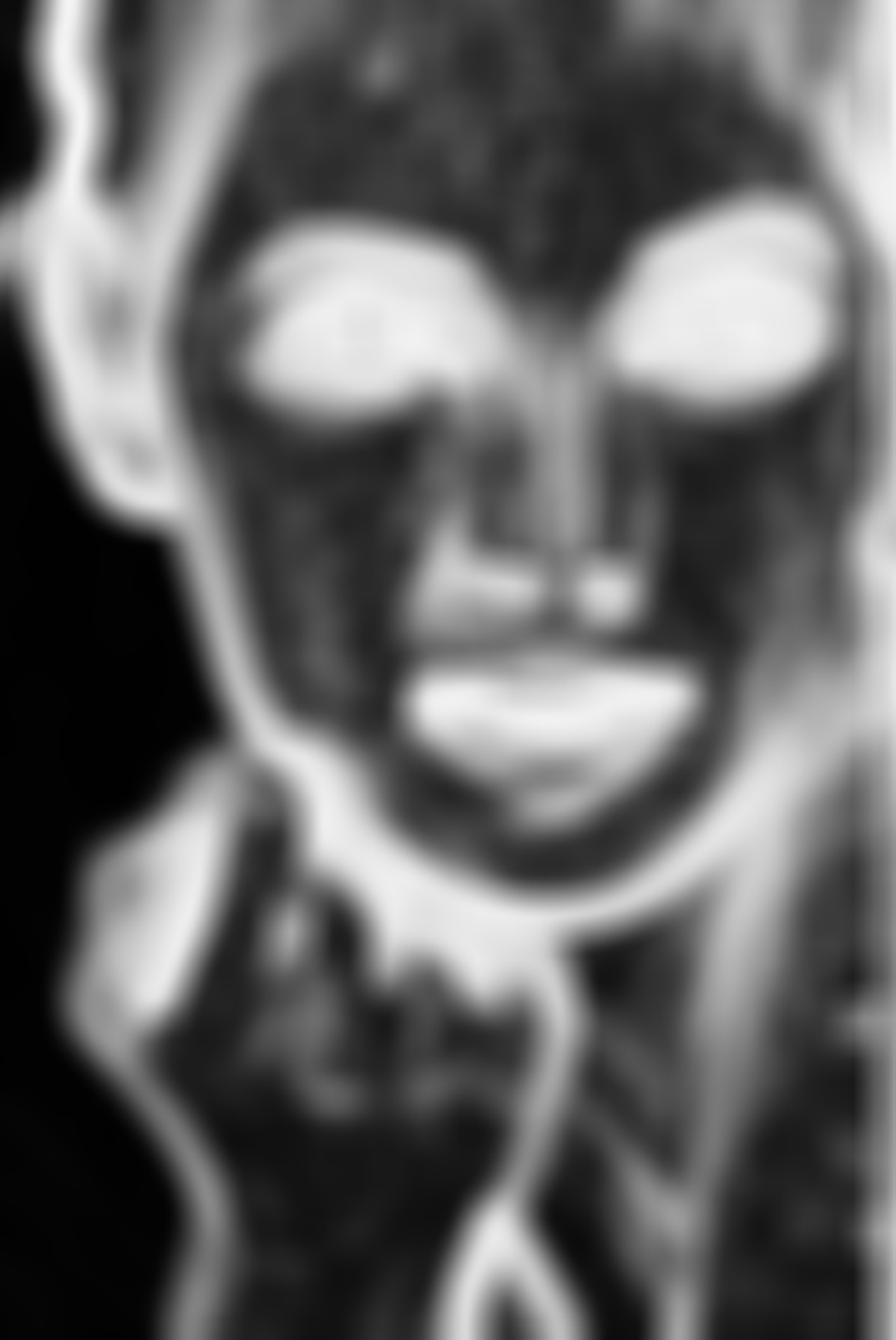}
\end{minipage}
}
\subfigure[\textbf{$\mathcal{F}_L{(I)}$}]{
\begin{minipage}{2.3cm}
\centering
\includegraphics[width=1.1\linewidth,left]{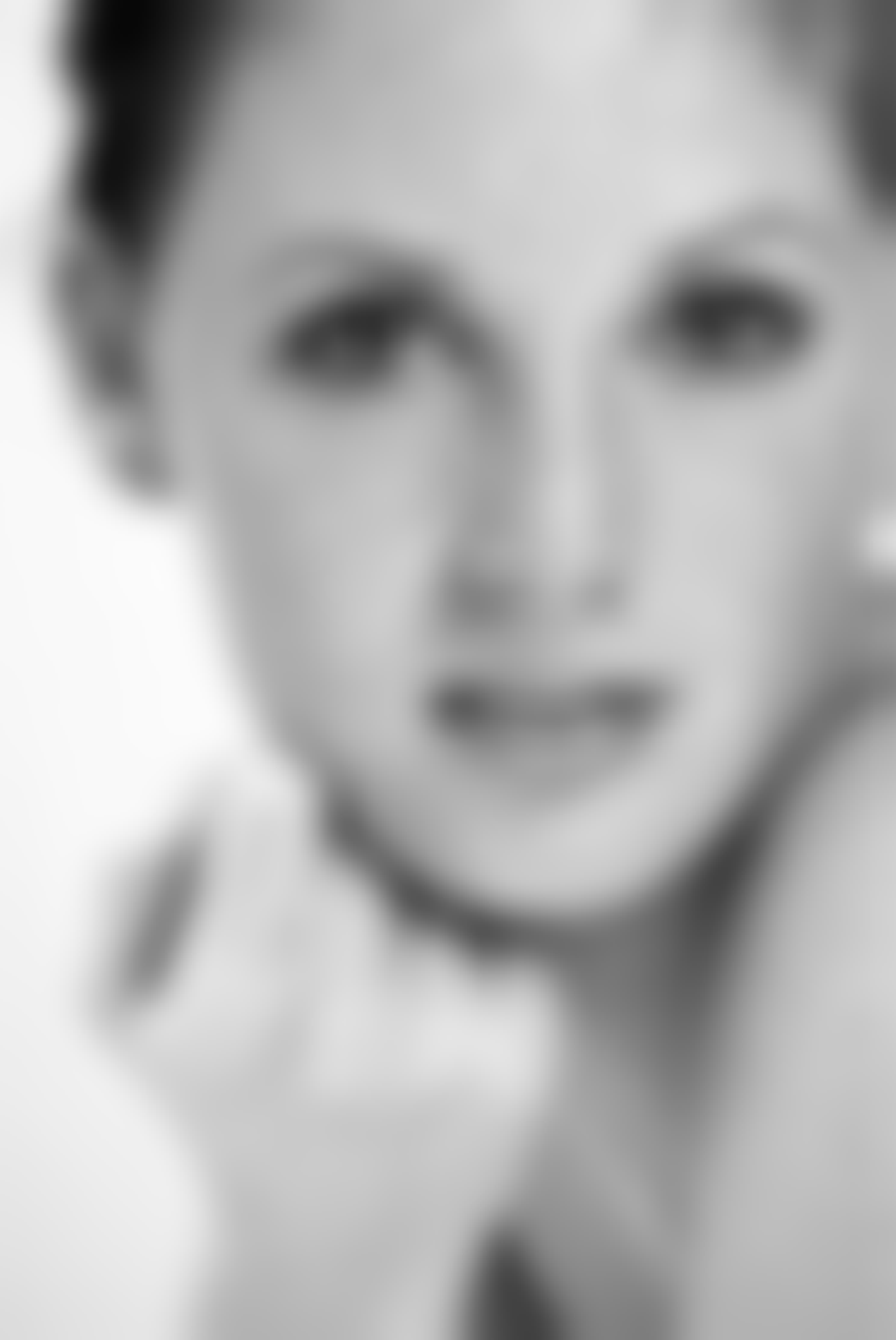}
\end{minipage}
}
\subfigure[\textbf{GF \cite{he2012guided}}]{
\begin{minipage}{2.3cm}
\centering
\includegraphics[width=1.1\linewidth,left]{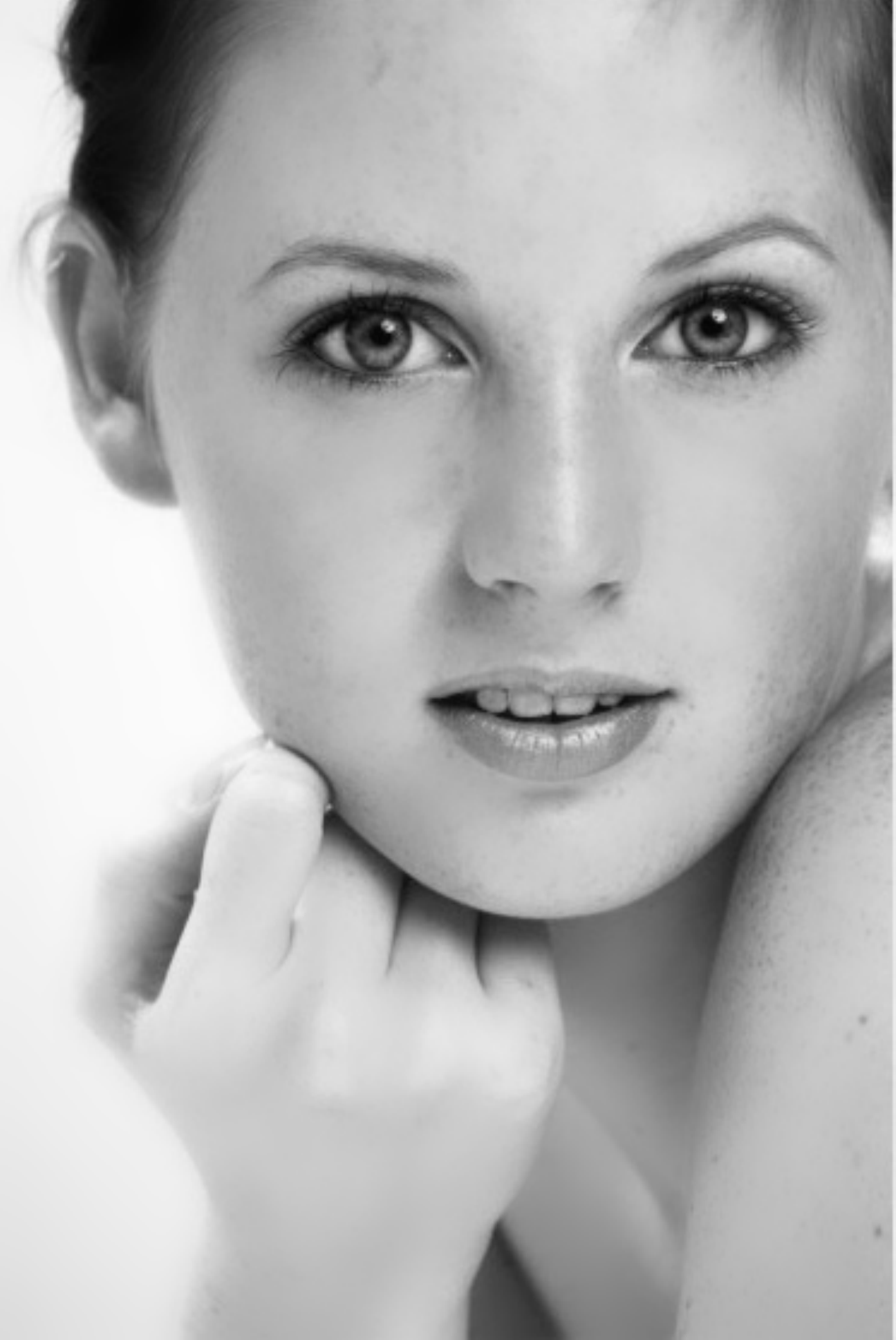}
\end{minipage}
}
\subfigure[\textbf{\textit{This paper}}]{
\begin{minipage}{2.3cm}
\centering
\includegraphics[width=1.1\linewidth,left]{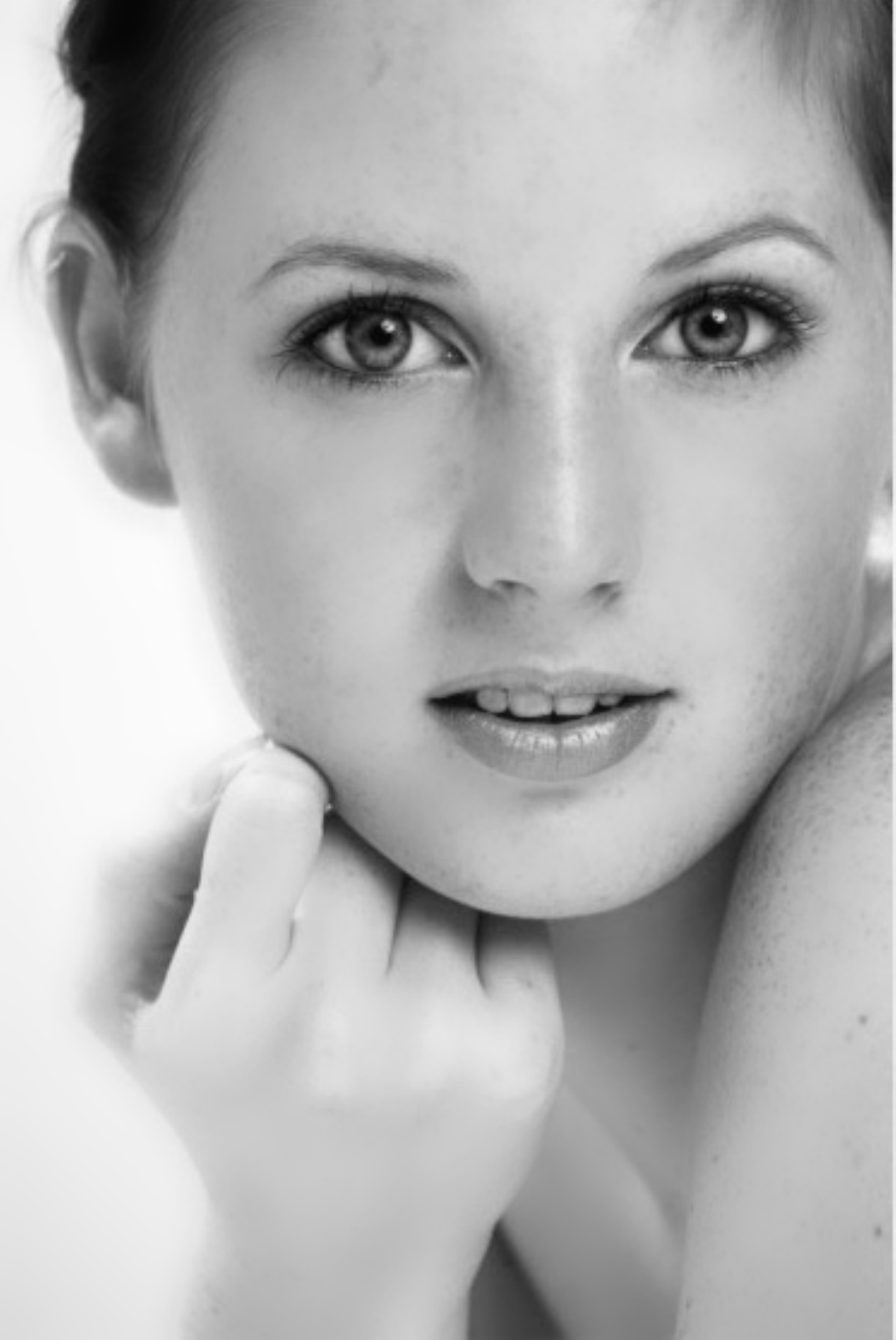}
\end{minipage}
}
\caption{\textbf{Edge-preserving filtering.} a) Target and guidance image $I=G$; b) $G_m$ containing the important structures.
c) $f_a(I_m,G_m)$ estimated by Eq. (\ref{eq_a}) with $\epsilon=0.05^2$; d) $\mathcal{F}_L{(I)}$ obtained by a cascade of two box filters with radius $r=8$; e) The smoothing result obtained by the guided filter \cite{he2012guided}; f) Our smoothing result. Both our filter and guided filter \cite{he2012guided} can preserve good edges while removing noise.}
\label{exp_fig_smoothing}    
\end{figure}

\begin{figure}[!ht]
\centering 
\subfigure[\textbf{Target $I$}]{
\begin{minipage}{2.3cm}
\centering
\includegraphics[width=1.1\linewidth,left]{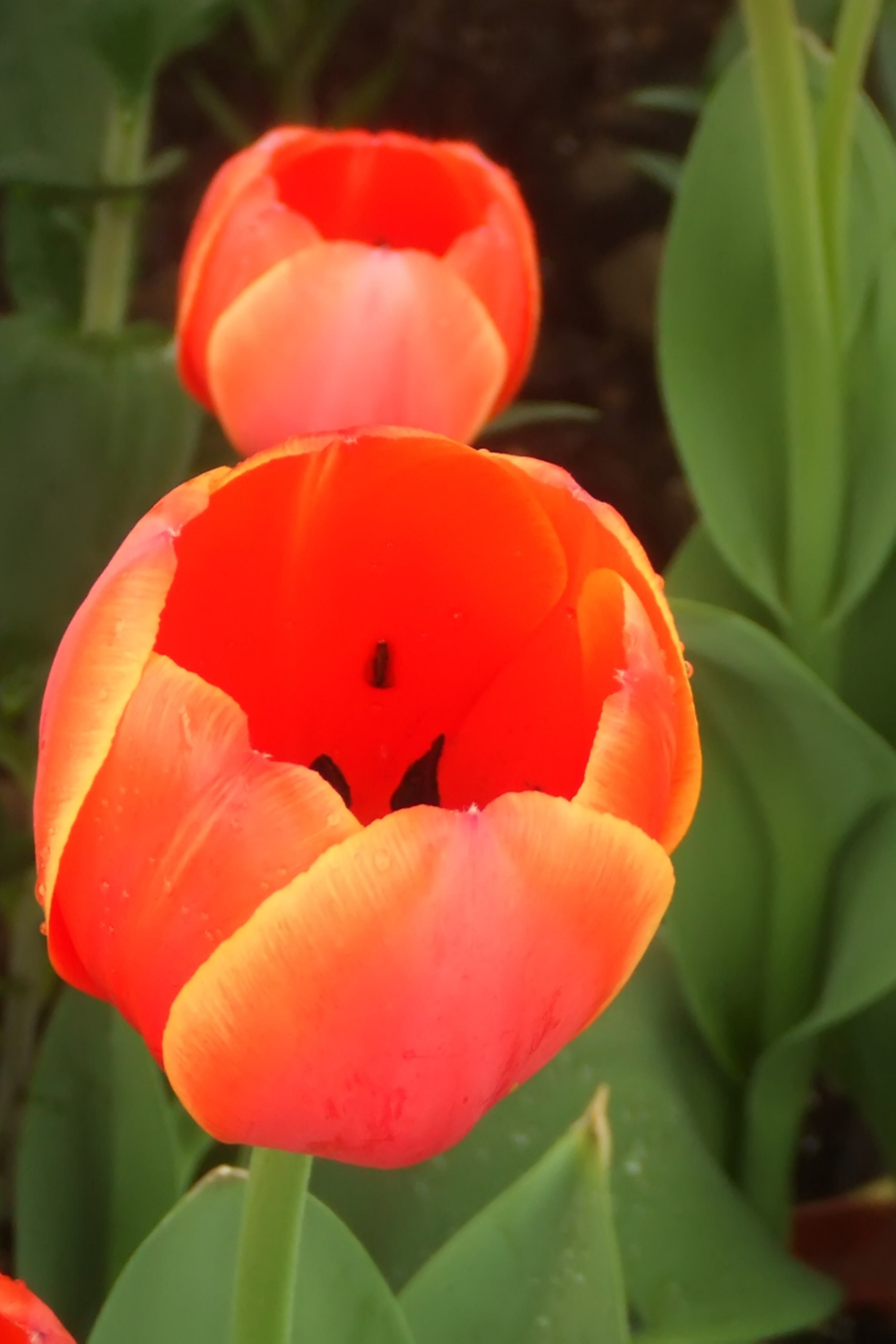}
\end{minipage}
}
\subfigure[\textbf{GF \cite{he2012guided}}]{
\begin{minipage}{2.3cm}
\centering
\includegraphics[width=1.1\linewidth,left]{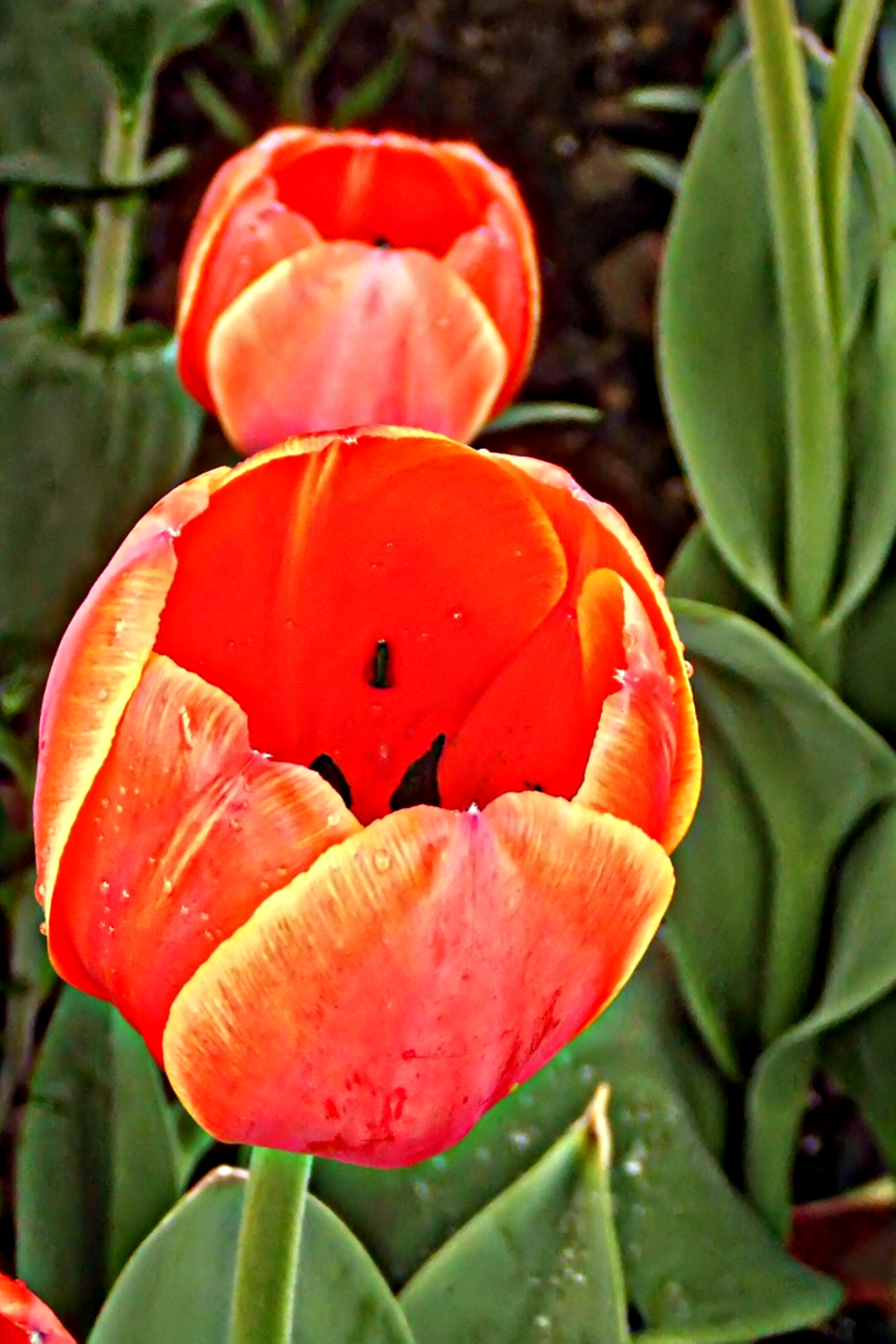}
\end{minipage}
}
\subfigure[\textbf{\textit{This paper}}]{
\begin{minipage}{2.3cm}
\centering
\includegraphics[width=1.1\linewidth,left]{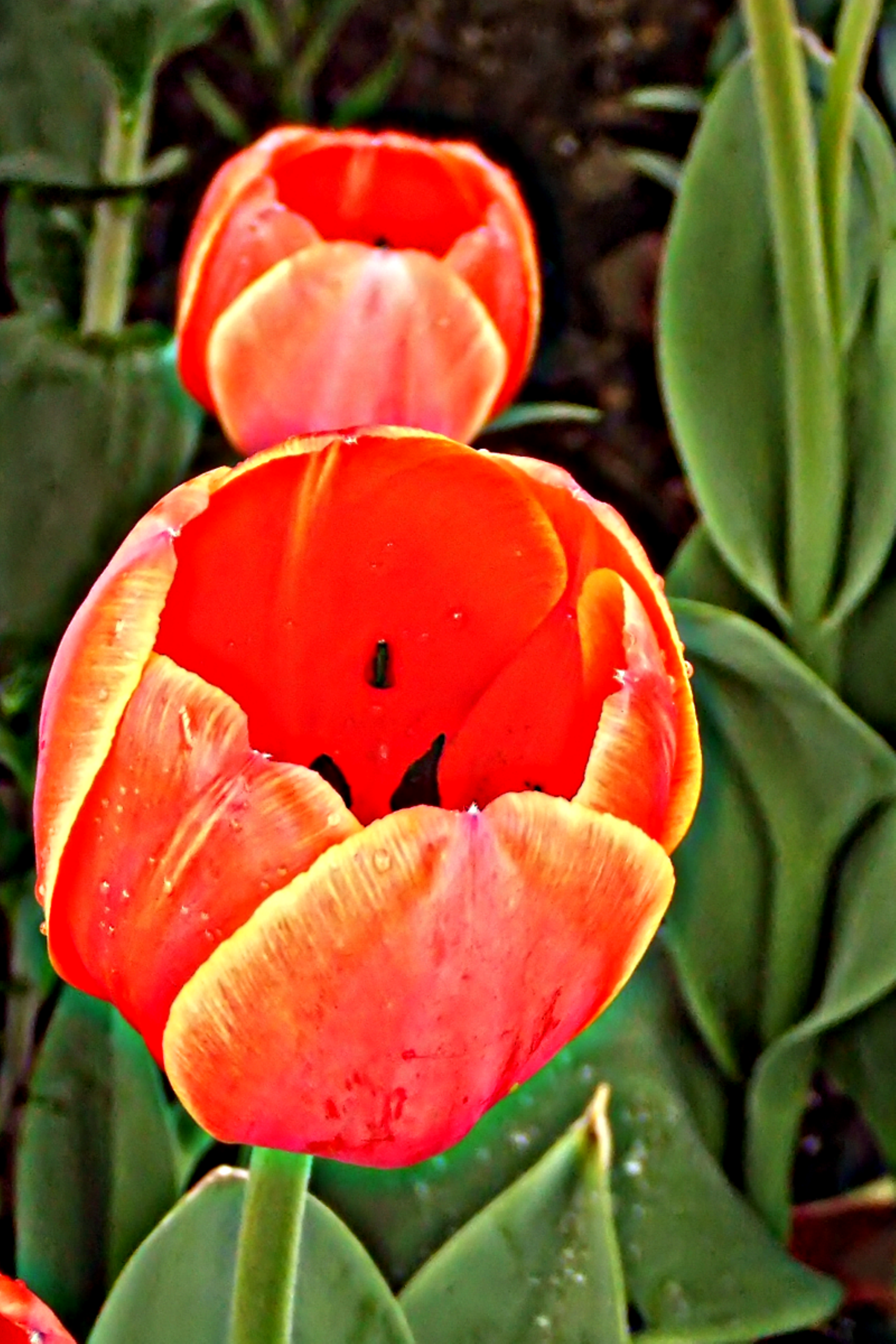}
\end{minipage}
}
\caption{\textbf{Detail enhancement.} The parameters are $r=16$, $\epsilon=0.1^2$. Our filter without learning, as defined in Eq. (\ref{eq_new_form3}), performs as good as the guided filter \cite{he2012guided}.
}
\label{exp_fig_enhancement}    
\end{figure}

\begin{figure}[!ht]
\centering 
\subfigure[\textbf{Target $I$}]{
\begin{minipage}{1.85cm}
\centering
\includegraphics[width=1.1\linewidth,left]{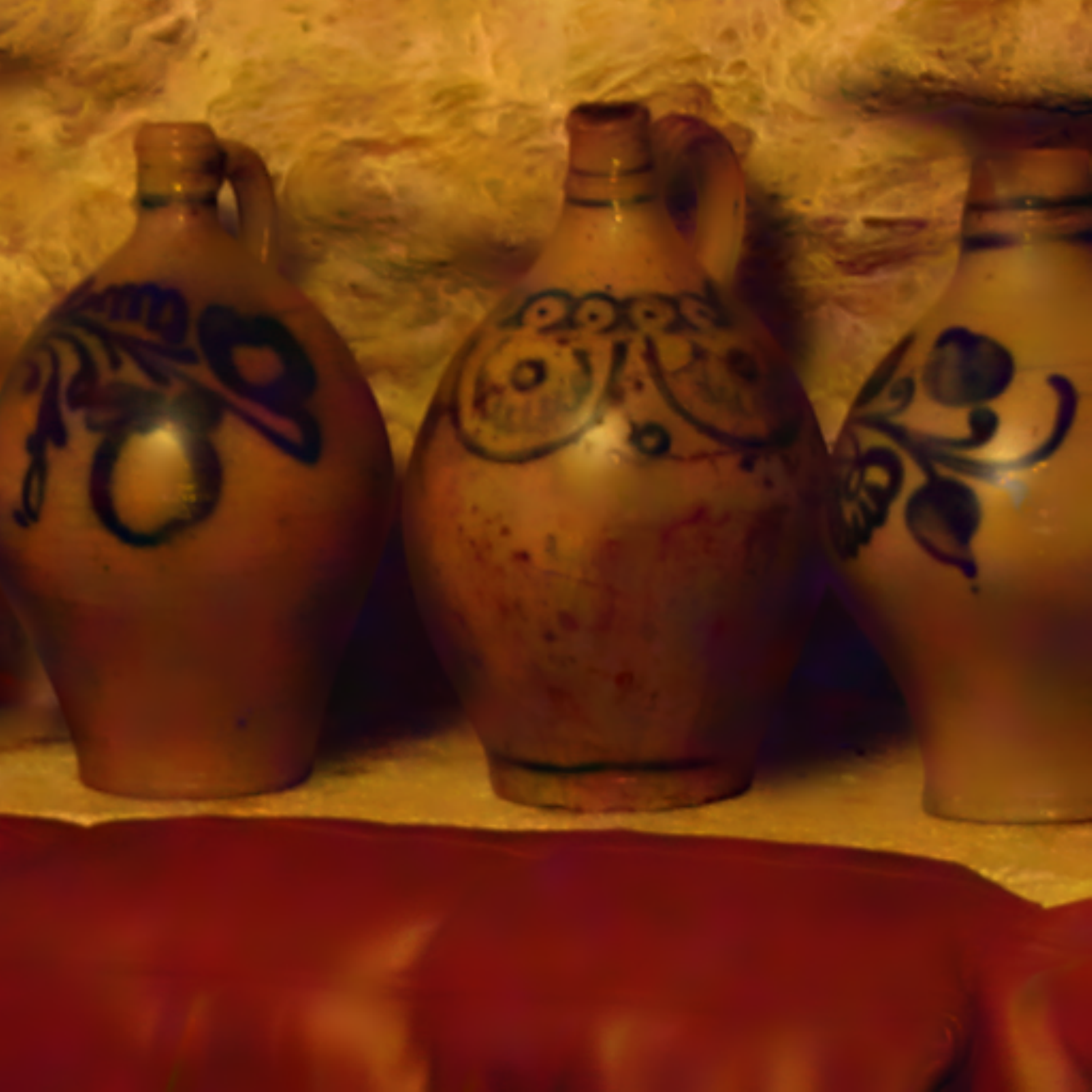}
\end{minipage}
}
\subfigure[\textbf{Guidance $G$}]{
\begin{minipage}{1.85cm}
\centering
\includegraphics[width=1.1\linewidth,left]{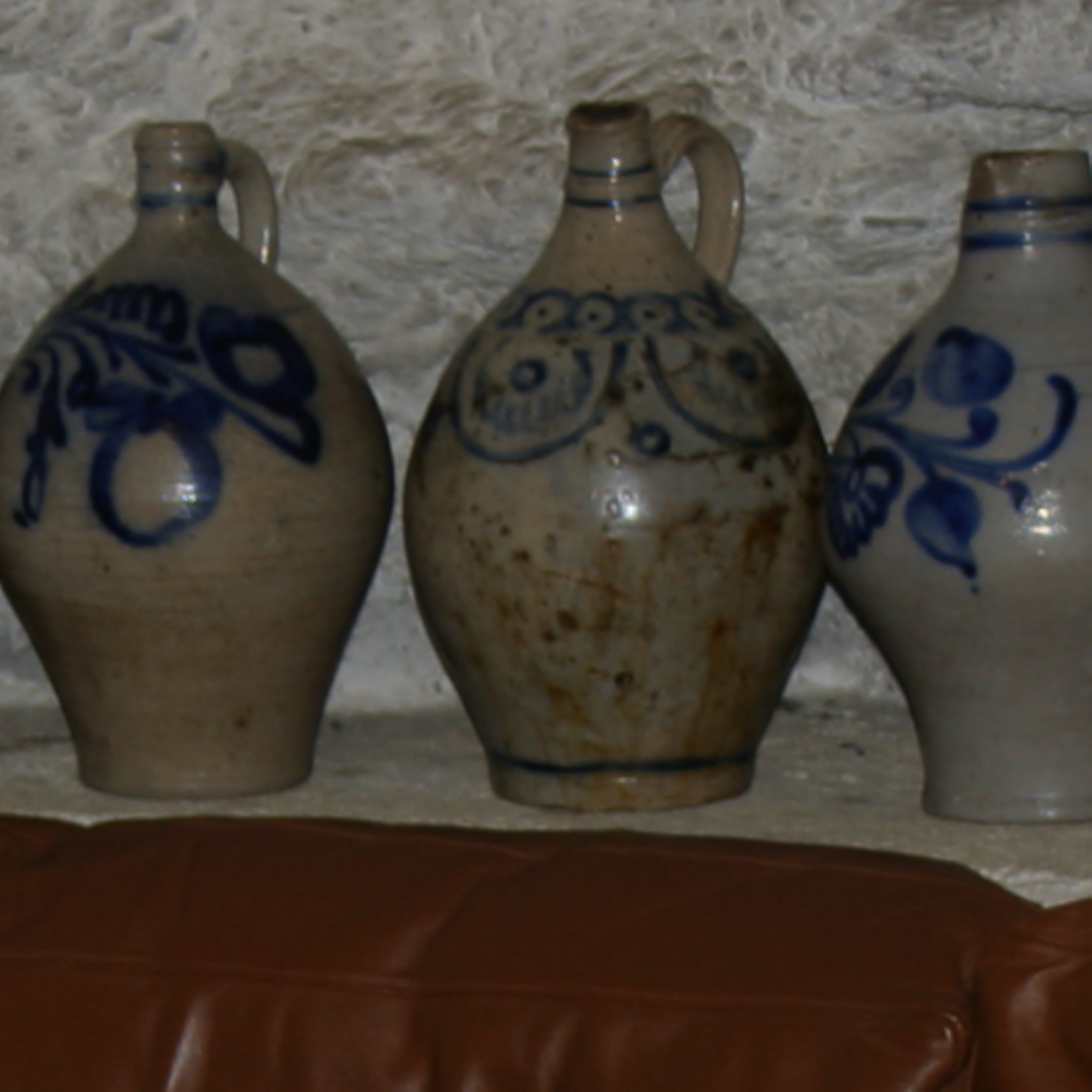}
\end{minipage}
}
\subfigure[\textbf{GF \cite{he2012guided}}]{
\begin{minipage}{1.85cm}
\centering
\includegraphics[width=1.1\linewidth,left]{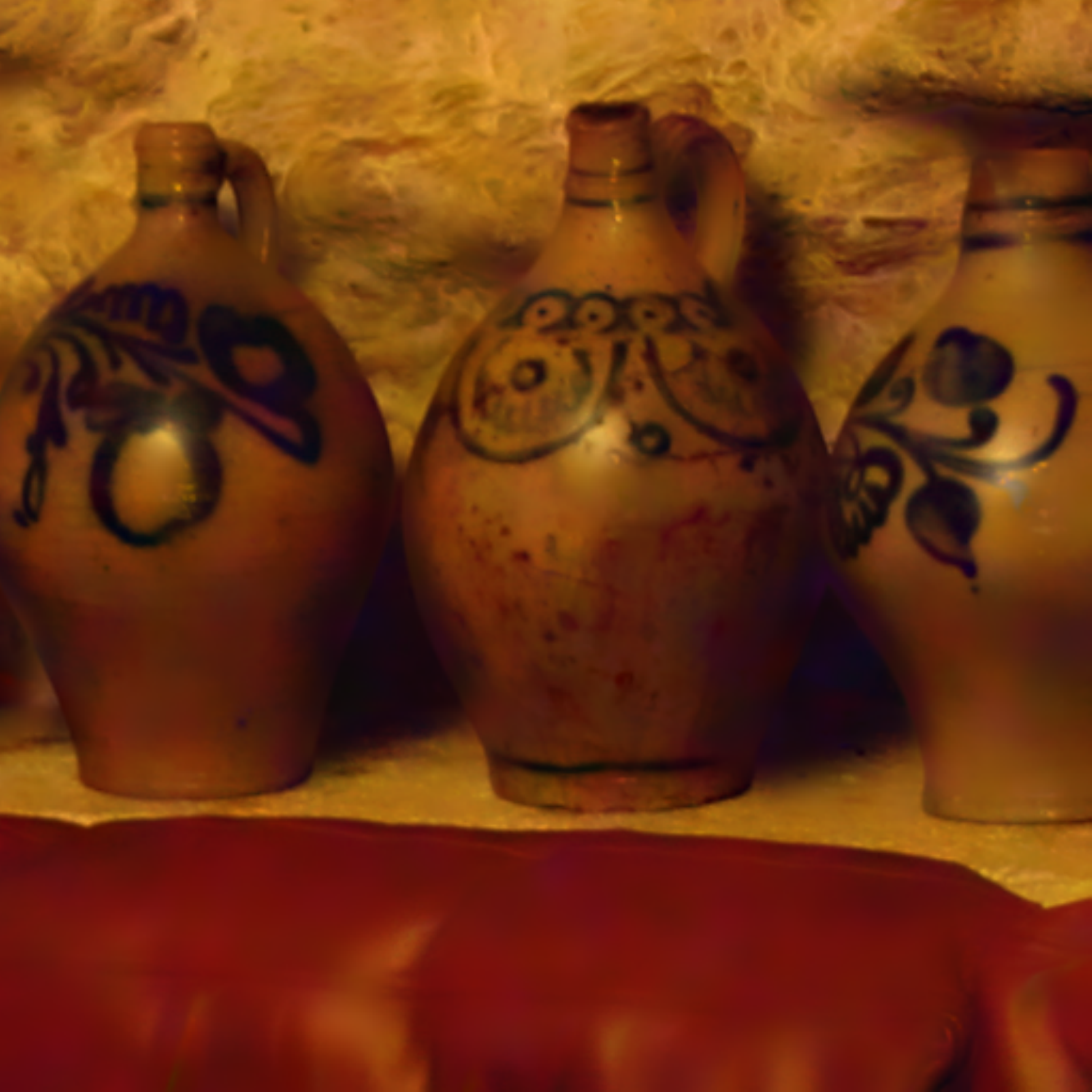}
\end{minipage}
}
\subfigure[\textbf{\textit{This paper}}]{
\begin{minipage}{1.85cm}
\centering
\includegraphics[width=1.1\linewidth,left]{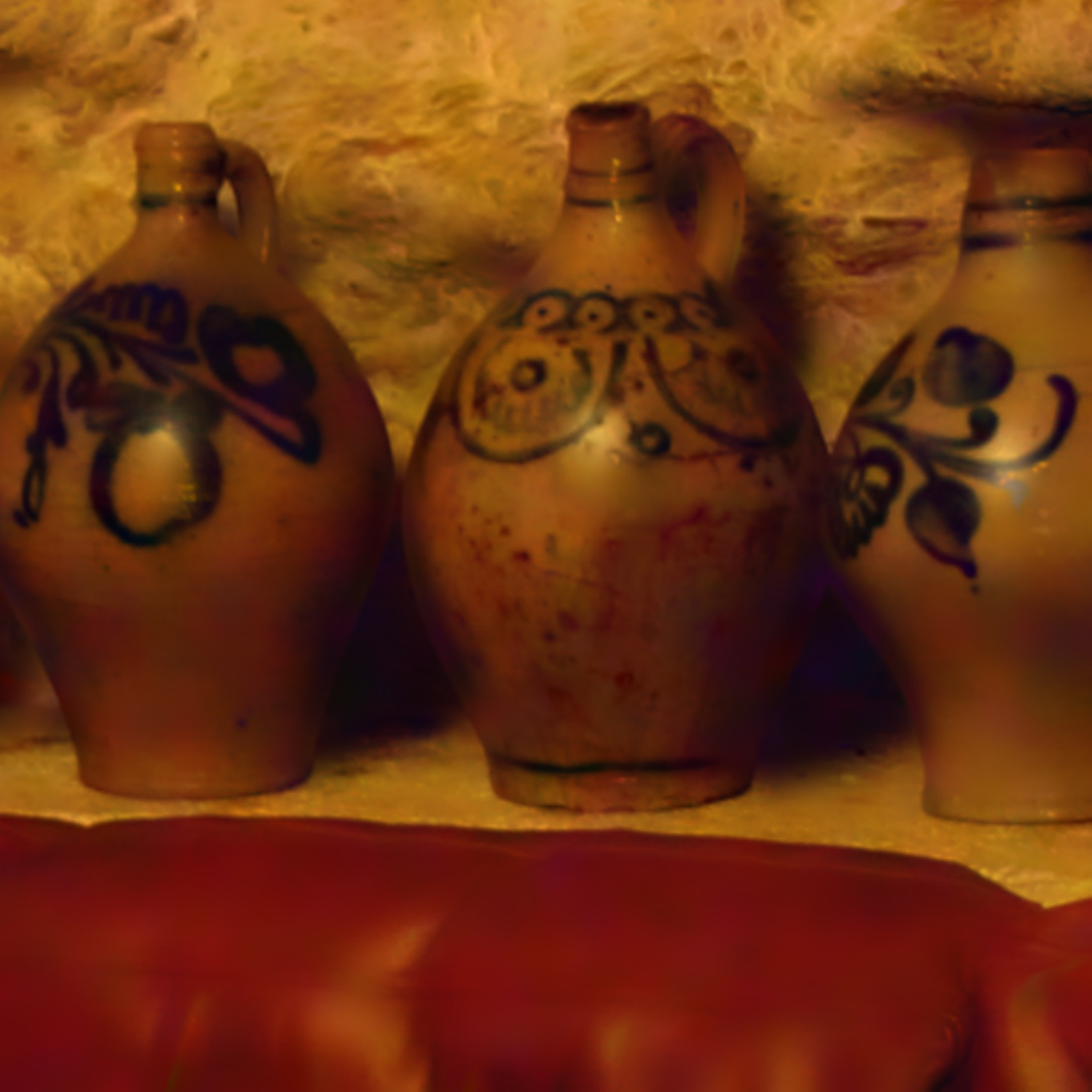}
\end{minipage}
}
\caption{\textbf{Flash/no-flash denoising.} The parameters are $r=8$, $\epsilon=0.2^2$. Our filter without learning, as defined in Eq. (\ref{eq_new_form3}), performs as good as the guided filter \cite{he2012guided}.
}
\label{exp_fig_flash_denoising}    
\end{figure}

\textbf{Qualitative results.} The first example performs edge-preserving smoothing on a gray-scale image. The second example is about detail enhancement. For both, the target image and guided image are identical. Fig. \ref{exp_fig_smoothing} and Fig. \ref{exp_fig_enhancement} show the results of filtering, where we can see that our filter performs as good as the guided filter \cite{he2012guided} in preserving structures. In the third example, we denoise a no-flash image under the guidance of its flash version to verify the effect of structure-transferring. The denoising results of our filter and the guided filter \cite{he2012guided} in Fig. \ref{exp_fig_flash_denoising} are consistent and don't have gradient reversal artifacts.

\textbf{Quantitative results.}
Next, we compare our filter with the guided filter (GF) \cite{he2012guided} and the weighted guided filter (WGF) \cite{li2014weighted} for natural image denoising on BSDS500 and depth upsampling on NYU Depth V2. There are two hyperparameters, $r$ and $\epsilon$ in GF and WGF. Grid-search is used to find the optimal hyperparameters. The results in Table \ref{exp_tab1} show that our filter performs at least as good as the guided filter \cite{he2012guided} and the weighted guided filter \cite{li2014weighted}, indicating that our formulation makes sense as a guided filter.
\begin{table}[t]
\small
\caption{\textbf{Quantitative results} for image denoising on BSDS500 and depth upsampling on NYU Depth V2. When the functions $a$ of the guided filter \cite{he2012guided} and the weighted guided filter \cite{li2014weighted} are used for $f_a$, our filter is denoted by ``Ours + GF" and ``Ours + WGF", respectively. Our filters perform at least as good as the baselines.}
\centering
\resizebox{1\linewidth}{!}{
\begin{tabular}{@{}lccccccc@{}}
\toprule
 & \multicolumn{3}{c}{\textbf{Denoising (PSNR)} $\uparrow$} & \multicolumn{3}{c}{\textbf{Upsampling (RMSE)} $\downarrow$}\\
\cmidrule(lr){2-4} \cmidrule(lr){5-7}
& $\sigma=15$ & $\sigma=25$ & $\sigma=50$ & $ 4\times $ & $ 8\times$ & $ 16\times $ \\
\hline
Bicubic/Input & 24.61 & 20.17 & 14.15 & 8.21  &  14.03 & 22.48   \\
GF \cite{he2012guided} &29.16&26.47&23.82 &7.25  &  12.38 & 19.86  \\
Ours + GF &29.24&26.59&23.84 &7.18  &  \textbf{12.28} & \textbf{19.75}  \\
\hline
WGF \cite{li2014weighted} &\textbf{29.40}&26.92&23.98 &\textbf{7.17}  &  12.33 & 19.79  \\
Ours + WGF &29.35&\textbf{26.96}&\textbf{23.99} &7.18  &  12.30 & 19.76  \\
\bottomrule
\end{tabular}}
\label{exp_tab1}
\end{table}
\subsection{Unsharp-mask guided filtering with learning}
\label{sec:exp2}
Next, we assess the benefit of our formulation when $f_a$ is learned by a neural network. We compare to four baselines: (\textit{i}) DMSG \cite{hui2016depth}, (\textit{ii}) DGF \cite{wu2018fast} (\textit{iii}) DJF \cite{li2019joint}, and (\textit{iv}) SVLRM \cite{pan2019spatially}. The experiments are performed on NYU Depth V2 \cite{silberman2012indoor} for depth upsampling ($16 \times$) and depth denoising ($\sigma = 50$). We compare these baselines separately. For each comparison, the network for $f_a$ is the same as the network used in the compared method. We use a box mean filter with radius $r=8$ to obtain $\mathcal{F}_L(I)$ and $\mathcal{F}_L(G)$. For depth upsampling ($16 \times$), we first upsample the low-resolution depth image by bicubic interpolation to obtain the target resolution for $\mathcal{F}_L(I)$. We also use the upsampled depth image as the input of the network, following \cite{li2019joint,pan2019spatially,wu2018fast}. One exception is the comparison with DMSG \cite{hui2016depth} which uses the original low-resolution depth image as the input of the network.
\begin{figure}[t!]
\centering 
\subfigure[\textbf{Upsampling}]{
\begin{minipage}{4.9cm}
\centering
\includegraphics[width=1.1\linewidth,left]{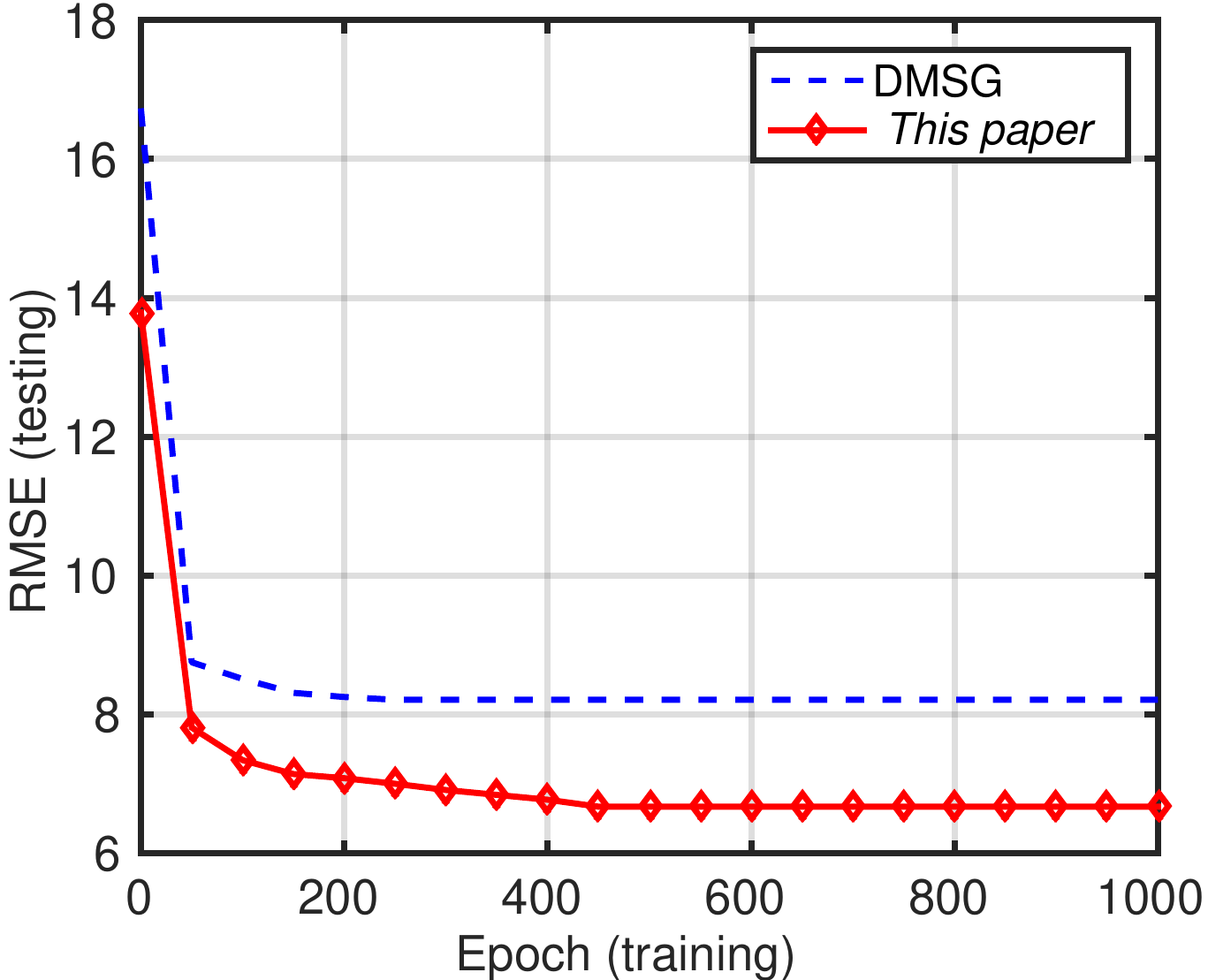}
\end{minipage}
}
\subfigure[\textbf{Guidance image}]{
\begin{minipage}{3.5cm}
\centering
\includegraphics[width=1.1\linewidth,left]{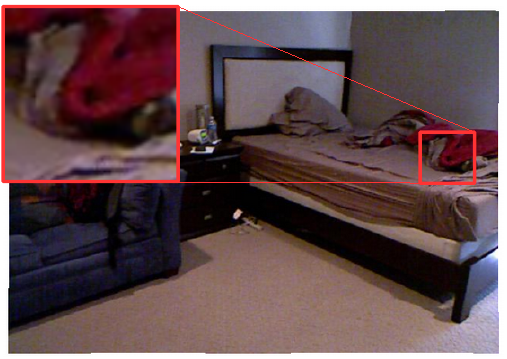}
\end{minipage}
}
\subfigure[\textbf{GT depth}]{
\begin{minipage}{3.5cm}
\centering
\includegraphics[width=1.1\linewidth,left]{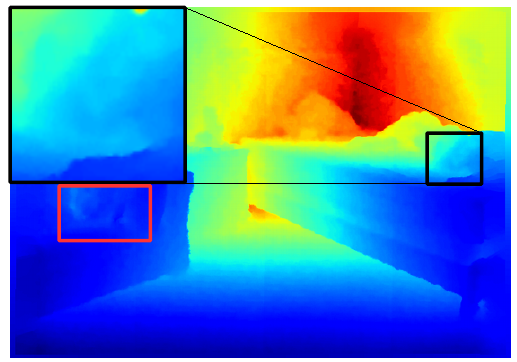}
\end{minipage}
}
\subfigure[\textbf{DMSG depth}]{
\begin{minipage}{3.5cm}
\centering
\includegraphics[width=1.08\linewidth,left]{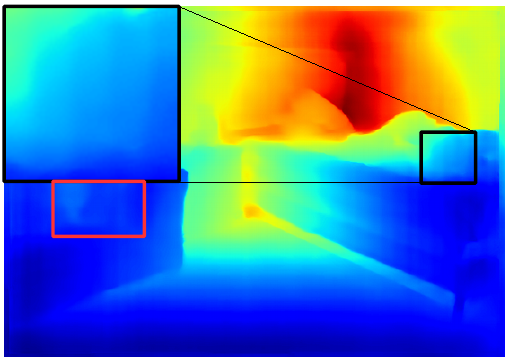}
\end{minipage}
}
\subfigure[\textbf{\textit{Our depth}}]{
\begin{minipage}{3.5cm}
\centering
\includegraphics[width=1.1\linewidth,left]{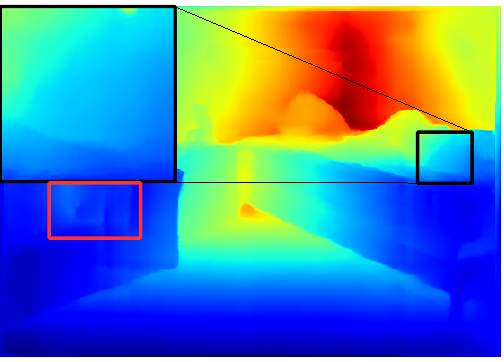}
\end{minipage}
}
\subfigure[\textbf{DMSG $(w_l*D_l)^{\uparrow }$}]{
\begin{minipage}{3.5cm}
\centering
\includegraphics[width=1.1\linewidth,left]{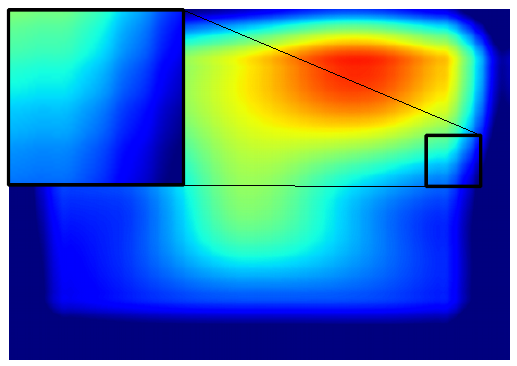}
\end{minipage}
}
\subfigure[\textbf{\textit{Our $\mathcal{F}_L(I)$}}]{
\begin{minipage}{3.5cm}
\centering
\includegraphics[width=1.1\linewidth,left]{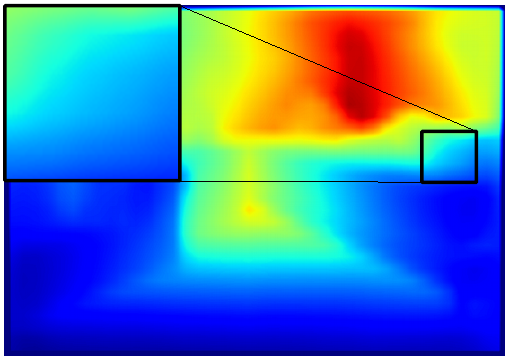}
\end{minipage}
}
\caption{\textbf{Comparison with DMSG \cite{hui2016depth}.} Compared to DGF, our approach better recovers finer edges as shown in the regions marked by the red boxes, and avoids producing artifacts as shown in the regions marked by the black boxes.
}
\label{exp_fig_dmsg}    
\end{figure}

\textbf{Comparison with DMSG \cite{hui2016depth}.} Fig. \ref{exp_fig_dmsg} (a) demonstrates our approach achieves better upsampling results than DMSG \cite{hui2016depth} in terms of RMSE. DMSG performs depth upsampling based on spectral decomposition. Specifically, a low-resolution depth image is first decomposed into low-frequency components and high-frequency components. The low-frequency components are directly upsampled by bicubic interpolation. The high-frequency components are upsampled by learning two convolutional networks. The first network is used to obtain multi-scale guidance. The other one performs multi-scale joint upsampling. The difference between our method and DMSG mainly lies in two aspects. First, we don't use the first network and just use the second network for amount function $f_a$ to explicitly perform structure transfer instead of directly predicting the filtered output image. We find that our approach avoids halo effects more successfully, as shown in Fig. \ref{exp_fig_dmsg} (d) and (e). Second, Hui \etal use a Gaussian filter to smooth the low-resolution target depth image when generating its low-frequency components. After that, they upsample the low-frequency components by bicubic interpolation. However, this step is likely to produce artifacts, as shown in Fig. \ref{exp_fig_dmsg} (f). Since the network learning focuses on upsampling high-frequency components, the artifacts still remain in the final upsampling output, as shown in Fig. \ref{exp_fig_dmsg} (d). By contrast, we first upsample the low-resolution target depth image before smoothing. By doing so, the artifacts generated by bicubic interpolation can be removed by smoothing, as shown in Fig. \ref{exp_fig_dmsg} (g).
\begin{figure}[!ht]
\centering 
\subfigure[\textbf{Upsampling}]{
\begin{minipage}{3.5cm}
\centering
\includegraphics[width=1.08\linewidth,left]{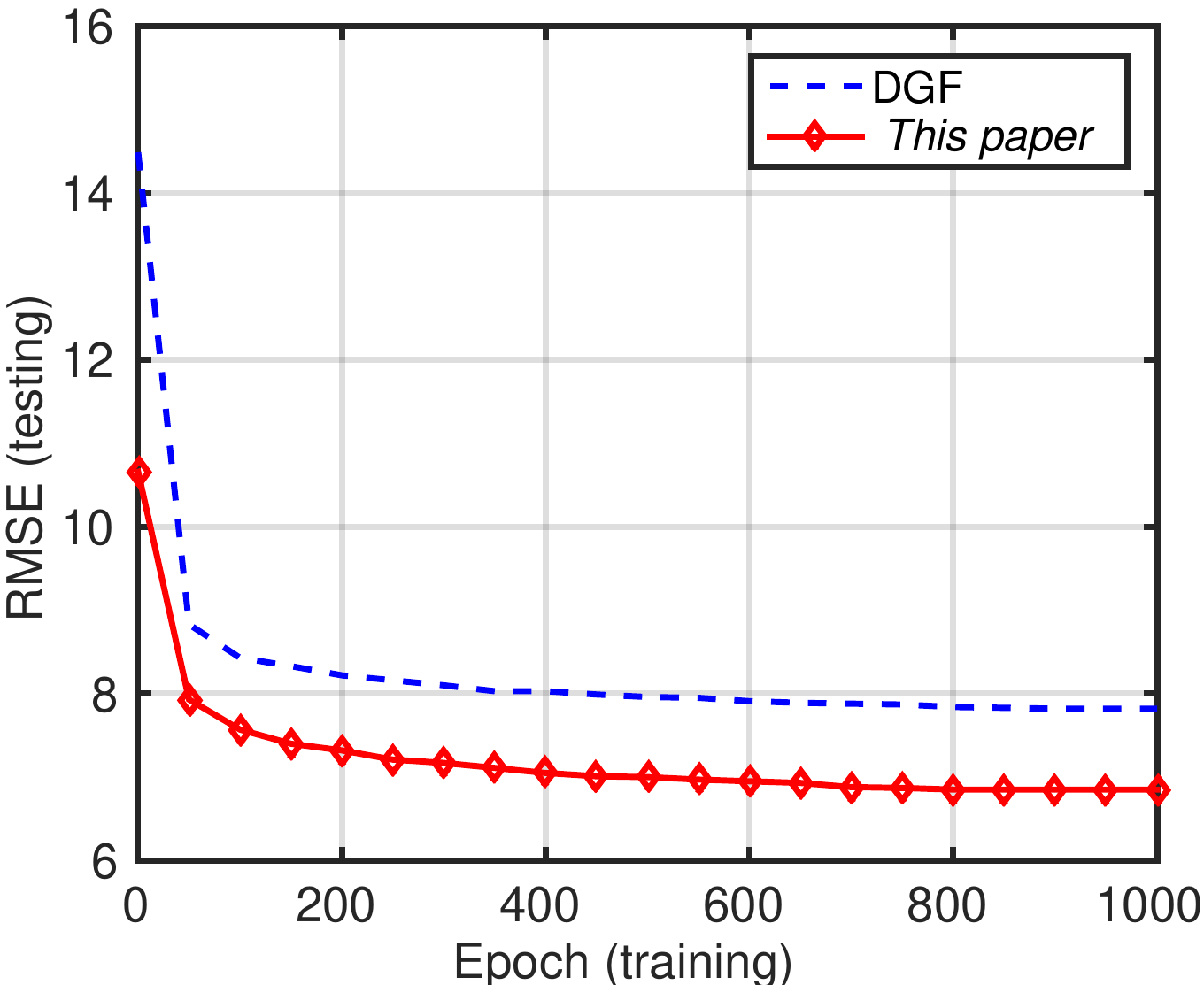}
\end{minipage}
}
\subfigure[\textbf{Denoising}]{
\begin{minipage}{3.5cm}
\centering
\includegraphics[width=1.08\linewidth,left]{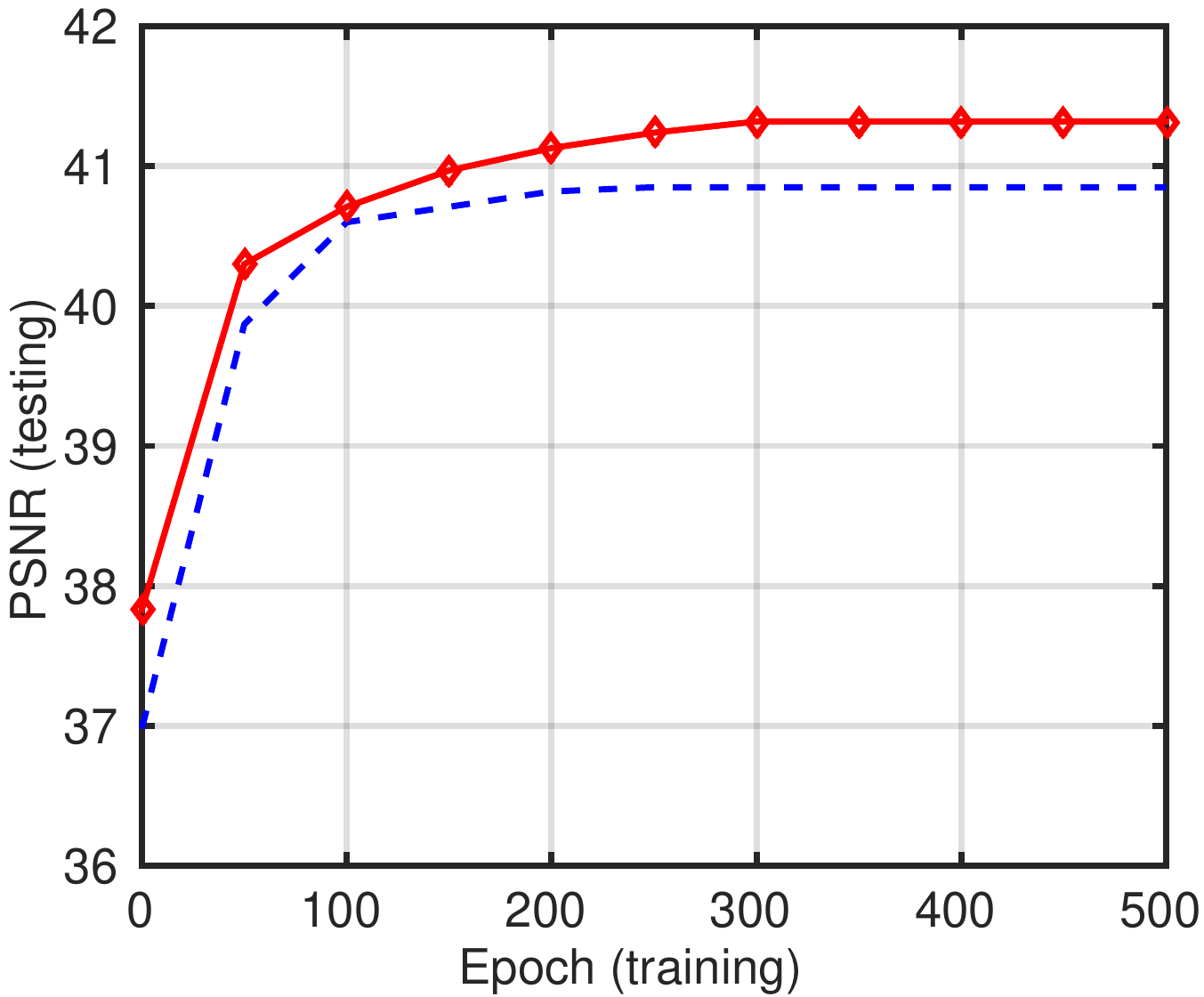}
\end{minipage}
}
\subfigure[\textbf{Guidance image}]{
\begin{minipage}{3.5cm}
\centering
\includegraphics[width=1.08\linewidth,left]{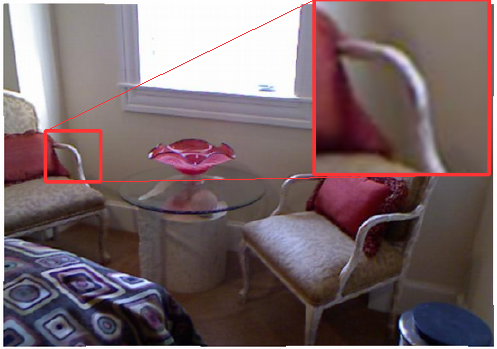}
\end{minipage}
}
\subfigure[\textbf{GT depth}]{
\begin{minipage}{3.5cm}
\centering
\includegraphics[width=1.08\linewidth,left]{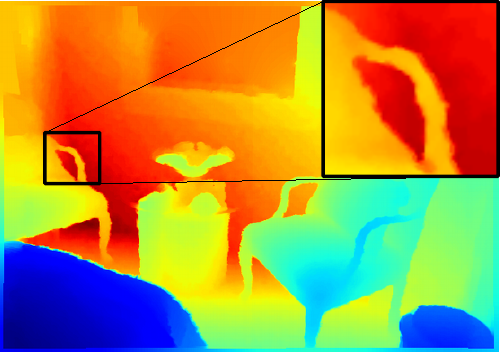}
\end{minipage}
}
\subfigure[\textbf{DGF depth}]{
\begin{minipage}{3.5cm}
\centering
\includegraphics[width=1.08\linewidth,left]{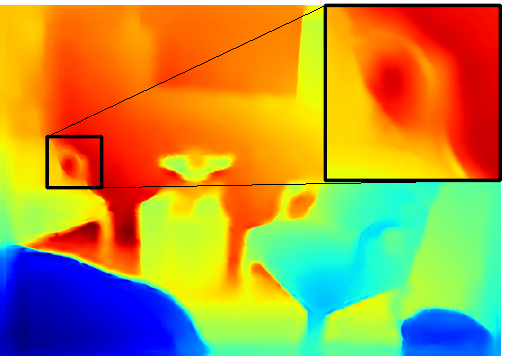}
\end{minipage}
}
\subfigure[\textbf{\textit{Our depth}}]{
\begin{minipage}{3.5cm}
\centering
\includegraphics[width=1.08\linewidth,left]{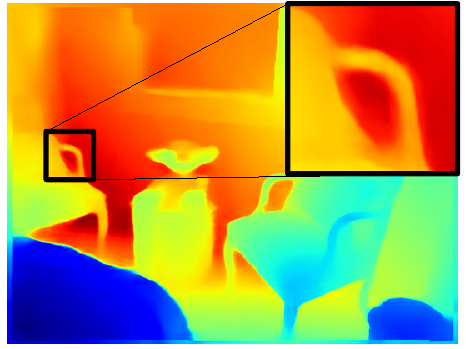}
\end{minipage}
}
\subfigure[\textbf{DGF $\bar{a}$}]{
\begin{minipage}{3.5cm}
\centering
\includegraphics[width=1.08\linewidth,left]{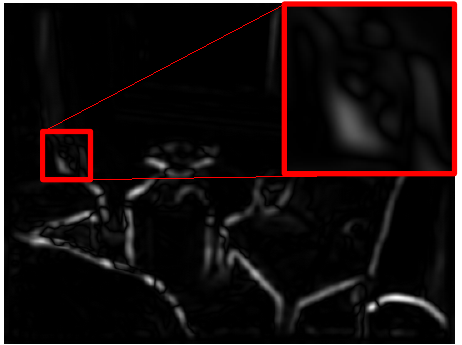}
\end{minipage}
}
\subfigure[\textbf{\textit{Our $f_a$}}]{
\begin{minipage}{3.5cm}
\centering
\includegraphics[width=1.08\linewidth,left]{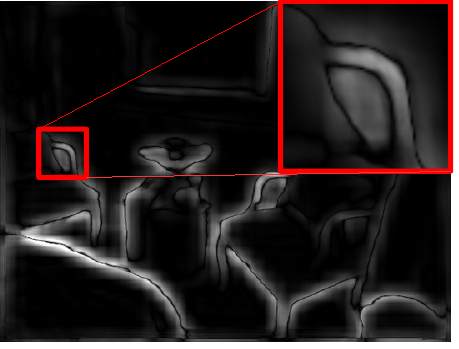}
\end{minipage}
}
\caption{\textbf{Comparison with DGF \cite{wu2018fast}.} The parameters of the guided filter \cite{he2012guided} used in DGF are $r=4$, $\epsilon=0.1^2$. Our learned amount function performs better on structure-transferring than the manually designed one as shown in the region marked by red boxes of (g) and (e). Thus, our filter reduces the over-smoothing of important edges as shown in the region marked by black boxes of (e) and (f).}
\label{exp_fig_dgf}    
\end{figure}

\textbf{Comparison with DGF \cite{wu2018fast}.} Wu \etal \cite{wu2018fast} learn two networks to amend the guidance and target images before feeding them to the guided filter \cite{he2012guided}. The learned guidance and target images fit the guided filter \cite{he2012guided} better than the original ones. However, DGF still suffers from the halo problem since its final filtering output is generated by the guided filter \cite{he2012guided}. Our approach performs better than DGF \cite{wu2018fast} for both upsampling and denoising tasks, as demonstrated in Fig. \ref{exp_fig_dgf} (a) and (b). As shown in Fig. \ref{exp_fig_dgf} (e) and (g), the important edges are unavoidable to be smoothed because the structure-transferring is performed in an undesirable fashion. By contrast, our approach performs better on preserving and transferring important structures, as shown in \ref{exp_fig_dgf} (f) and (h), as the amount function $f_a$ is learned in a pixel-adaptive way through a deep neural network instead of designed manually.
\begin{figure}[t!]
\centering 
\subfigure[\textbf{Upsampling}]{
\begin{minipage}{3.5cm}
\centering
\includegraphics[width=1.08\linewidth,left]{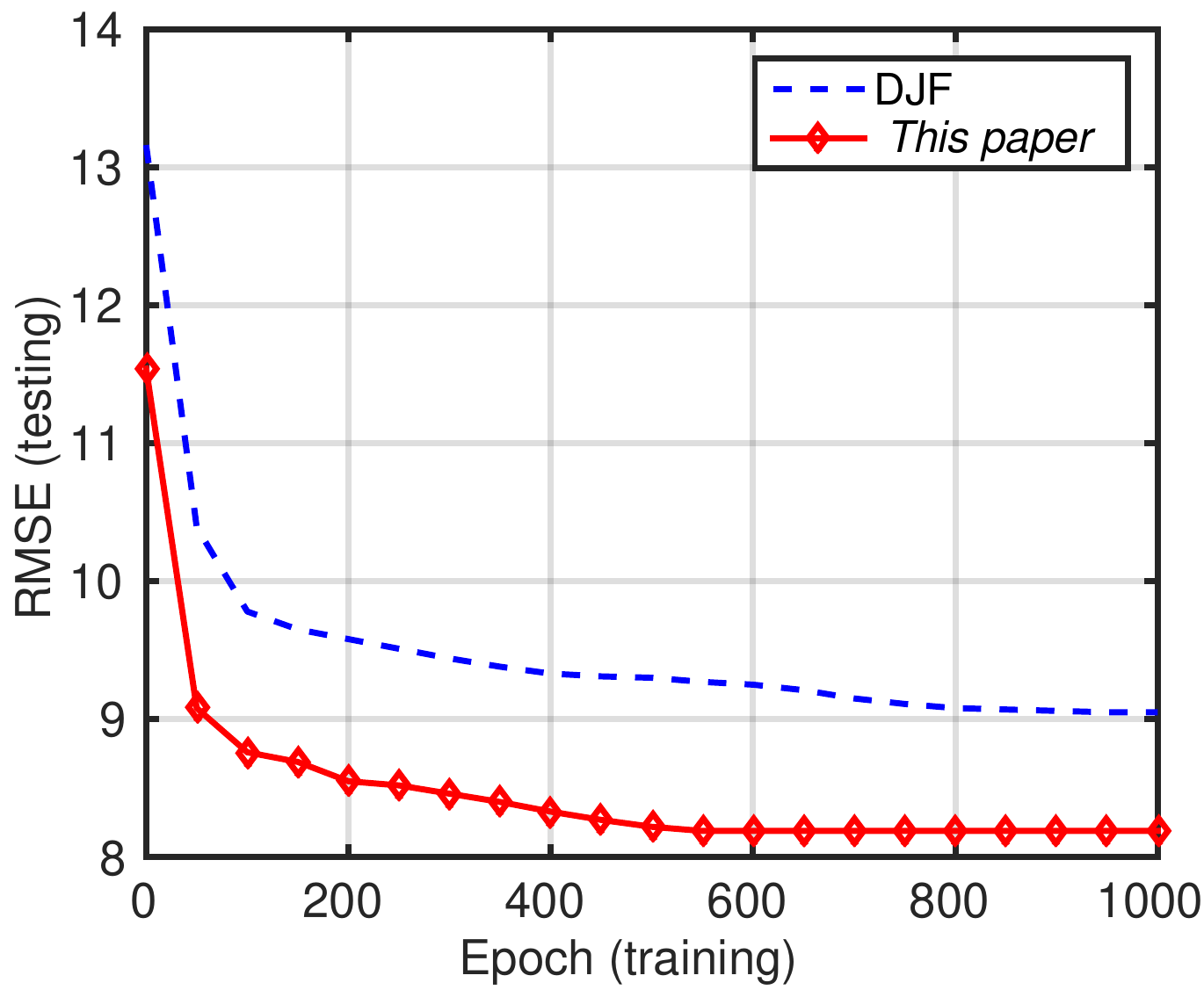}
\end{minipage}
}
\subfigure[\textbf{Denoising}]{
\begin{minipage}{3.5cm}
\centering
\includegraphics[width=1.08\linewidth,left]{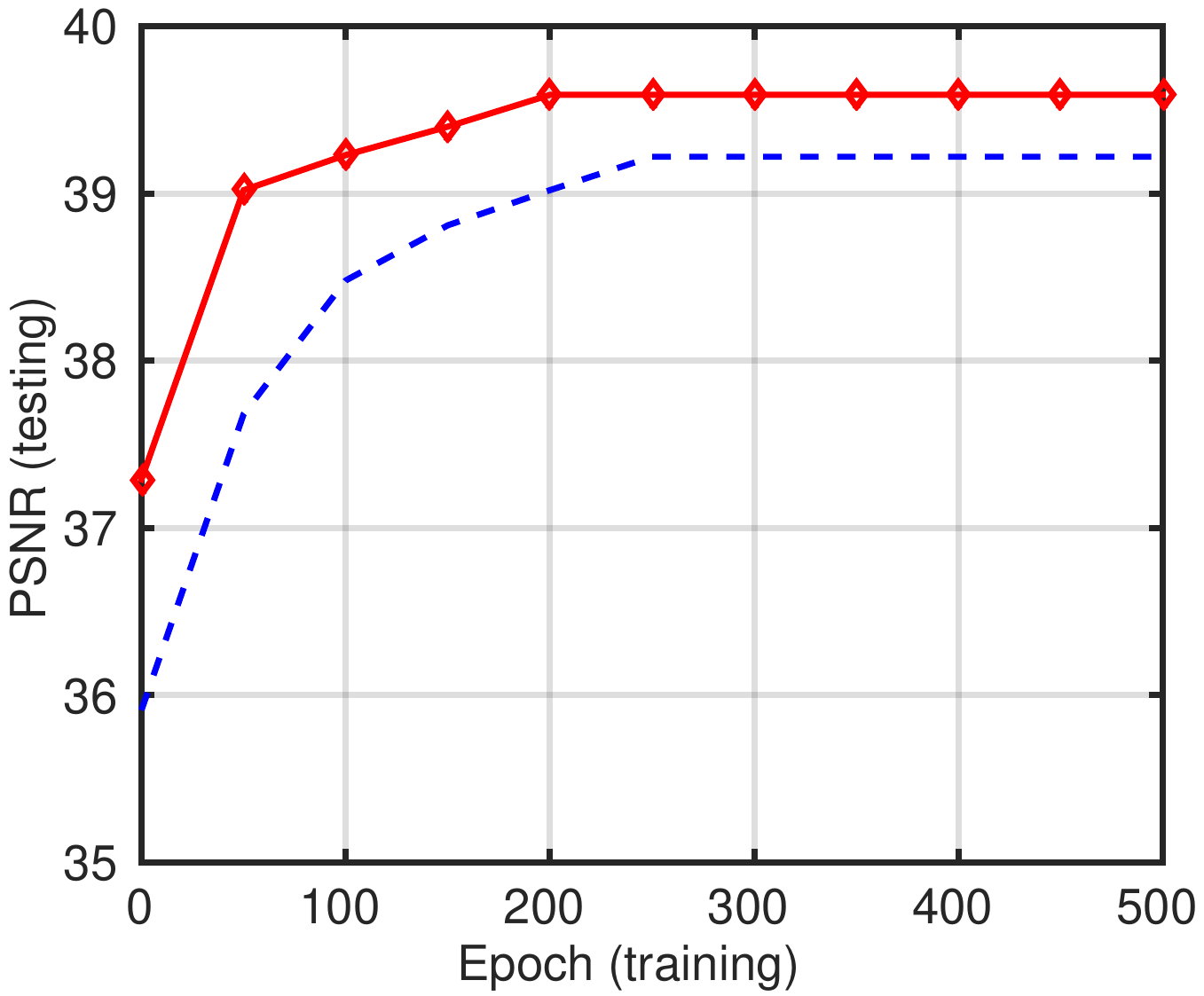}
\end{minipage}
}
\subfigure[\textbf{Guidance image}]{
\begin{minipage}{3.5cm}
\centering
\includegraphics[width=1.08\linewidth,left]{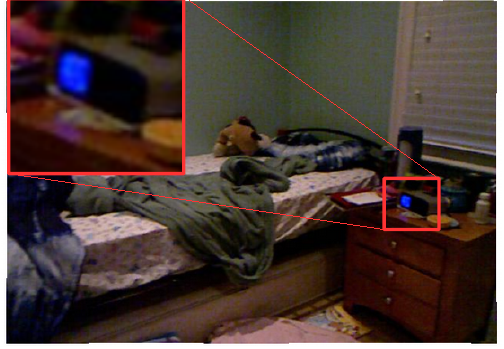}
\end{minipage}
}
\subfigure[\textbf{GT depth}]{
\begin{minipage}{3.5cm}
\centering
\includegraphics[width=1.08\linewidth,left]{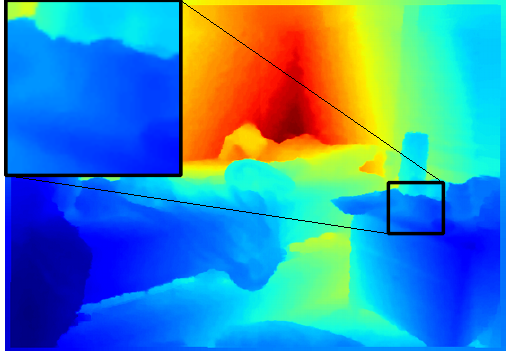}
\end{minipage}
}
\subfigure[\textbf{DJF depth}]{
\begin{minipage}{3.5cm}
\centering
\includegraphics[width=1.08\linewidth,left]{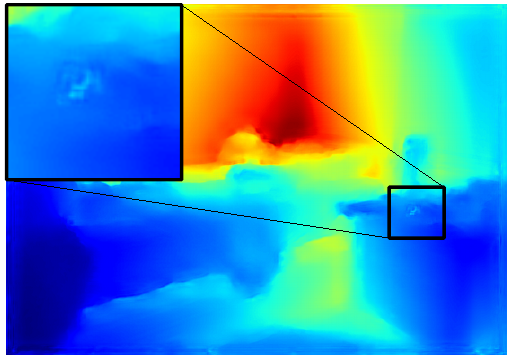}
\end{minipage}
}
\subfigure[\textbf{\textit{Our depth}}]{
\begin{minipage}{3.5cm}
\centering
\includegraphics[width=1.08\linewidth,left]{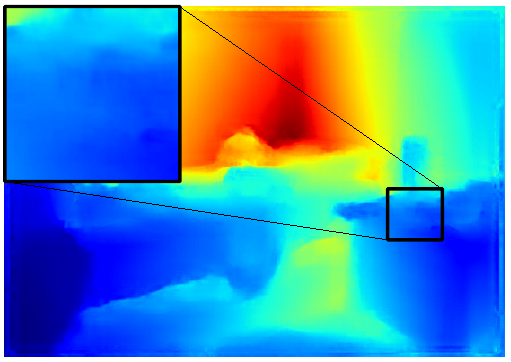}
\end{minipage}
}
\caption{\textbf{Comparison with DJF \cite{li2019joint}.} Our approach avoids unwanted structures transferred from the guidance image to the target image as shown in the regions marked by the black boxes, leading to more desirable filtering results than DJF.
}
\label{exp_fig_djf}    
\end{figure}

\textbf{Comparison with DJF \cite{li2019joint}.} The network in our method directly uses the unsharp masks of the target image and the guided image as inputs, and learns to estimate the amount function $f_a$ for explicitly deciding how to transfer the desired structure from the guidance image to the target image. By contrast, the network in DJF \cite{li2019joint} uses the original guidance image and target image as inputs, and directly predicts filtered output relying on feature fusion. The implicit structure transfer is likely to cause slow convergence and the unwanted contents to be transferred from guidance image to target image, as shown in Fig. \ref{exp_fig_djf}. From Fig. \ref{exp_fig_djf} (a) and (b), we can see our approach convergences faster and achieves better filtering performance than DJF \cite{li2019joint} on both upsampling and denoising tasks.
\begin{figure*}[t!]
\centering 
\subfigure[\textbf{Upsampling}]{
\begin{minipage}{3.25cm}
\centering
\includegraphics[width=1.1\linewidth,left]{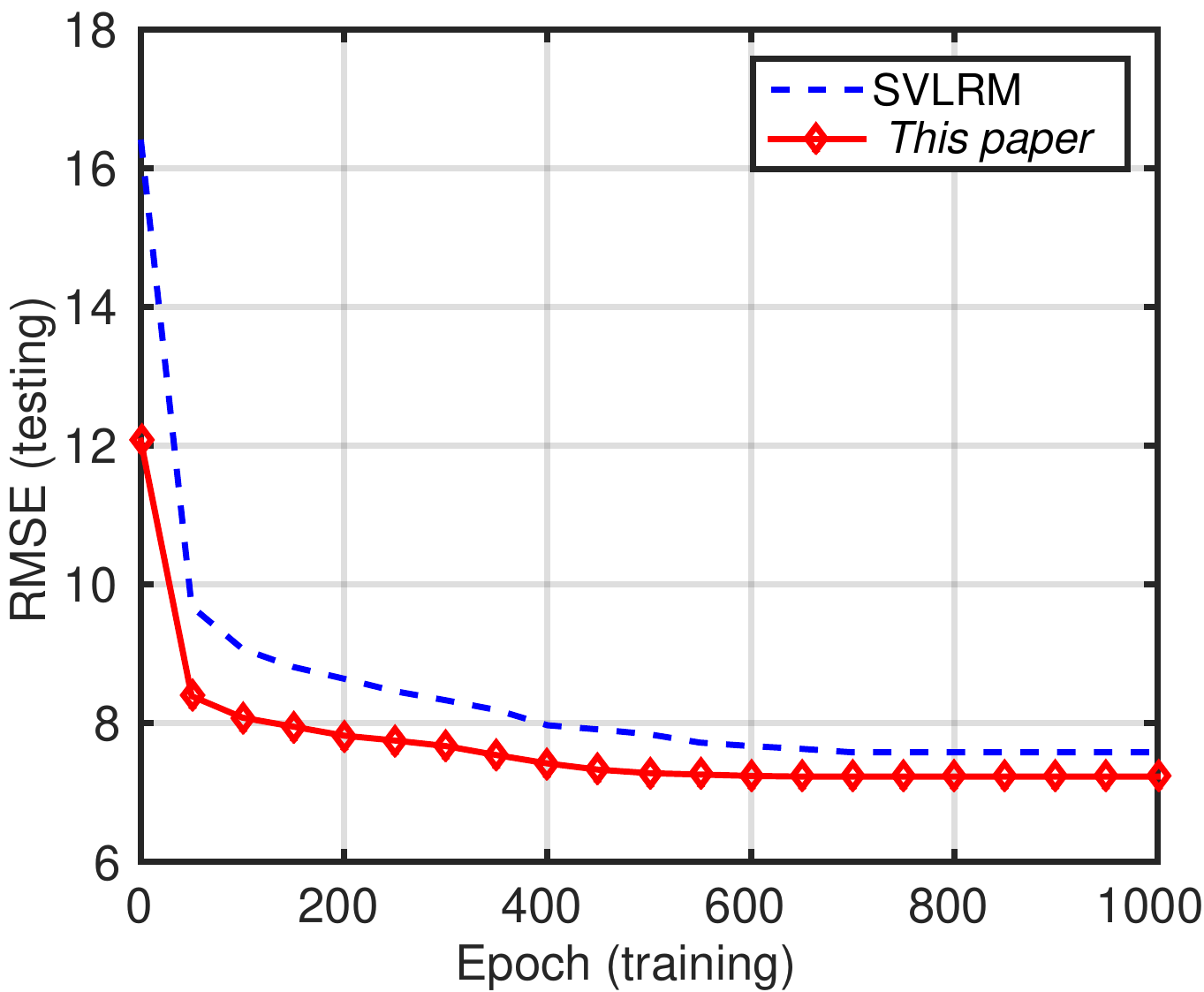}
\end{minipage}
}
\subfigure[\textbf{Guidance image}]{
\begin{minipage}{3.25cm}
\centering
\includegraphics[width=1.1\linewidth,left]{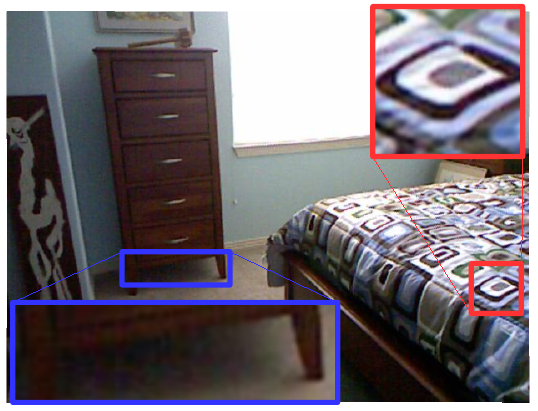}
\end{minipage}
}
\subfigure[\textbf{SVLRM depth}]{
\begin{minipage}{3.25cm}
\centering
\includegraphics[width=1.1\linewidth,left]{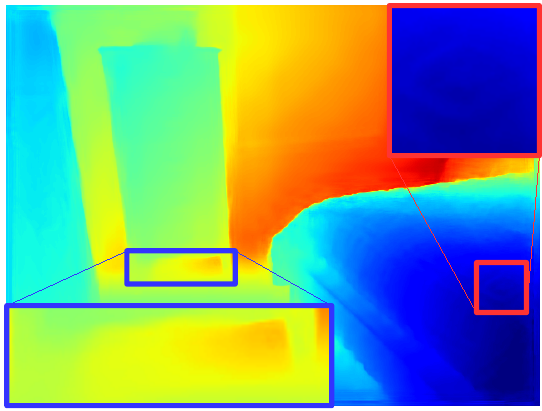}
\end{minipage}
}
\subfigure[\textbf{SVLRM $f_{\alpha}$}]{
\begin{minipage}{3.25cm}
\centering
\includegraphics[width=1.1\linewidth,left]{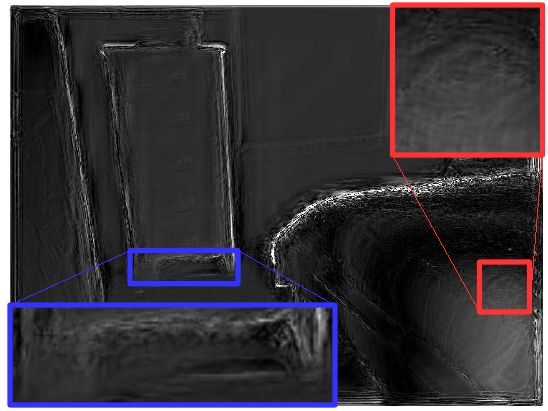}
\end{minipage}
}
\subfigure[\textbf{SVLRM $f_{\beta}$}]{
\begin{minipage}{3.25cm}
\centering
\includegraphics[width=1.1\linewidth,left]{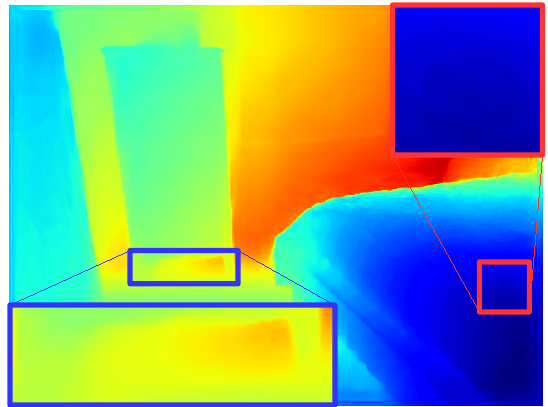}
\end{minipage}
}
\subfigure[\textbf{Denoising}]{
\begin{minipage}{3.25cm}
\centering
\includegraphics[width=1.1\linewidth,left]{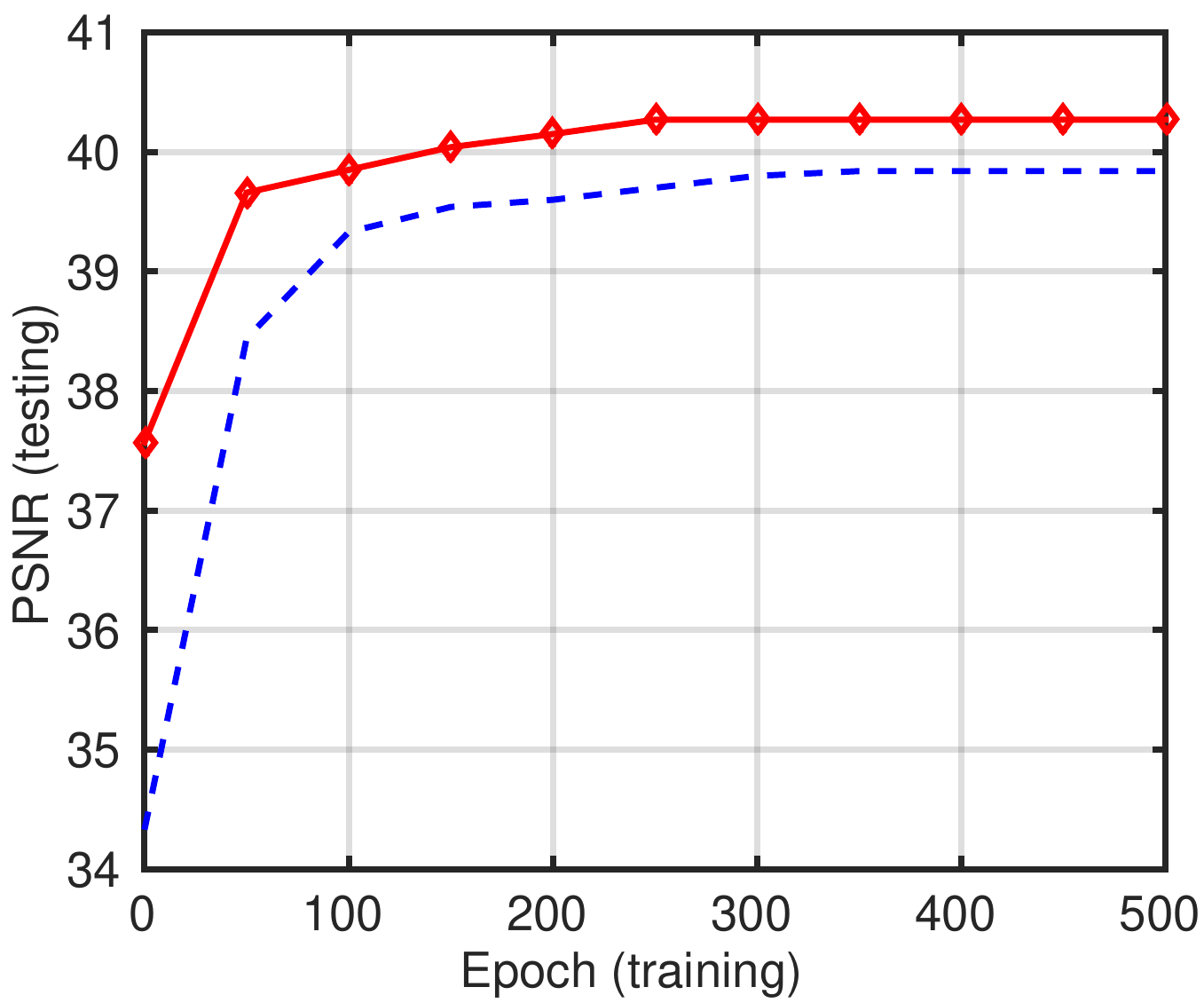}
\end{minipage}
}
\subfigure[\textbf{GT depth}]{
\begin{minipage}{3.25cm}
\centering
\includegraphics[width=1.1\linewidth,left]{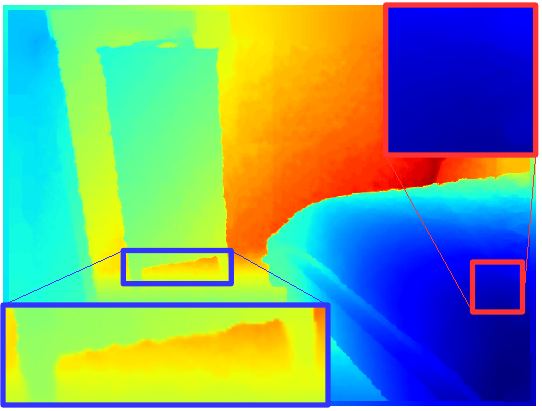}
\end{minipage}
}
\subfigure[\textbf{\textit{Our depth}}]{
\begin{minipage}{3.25cm}
\centering
\includegraphics[width=1.1\linewidth,left]{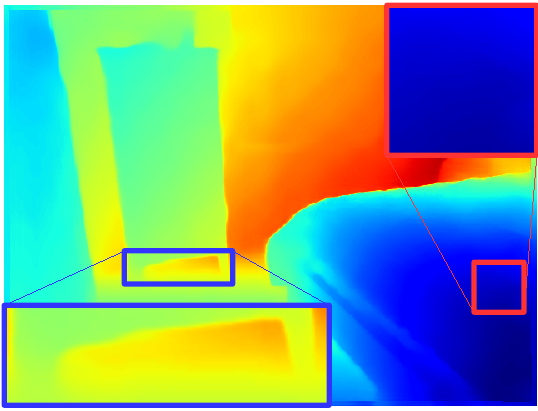}
\end{minipage}
}
\subfigure[\textbf{\textit{Our $f_a$}}]{
\begin{minipage}{3.25cm}
\centering
\includegraphics[width=1.1\linewidth,left]{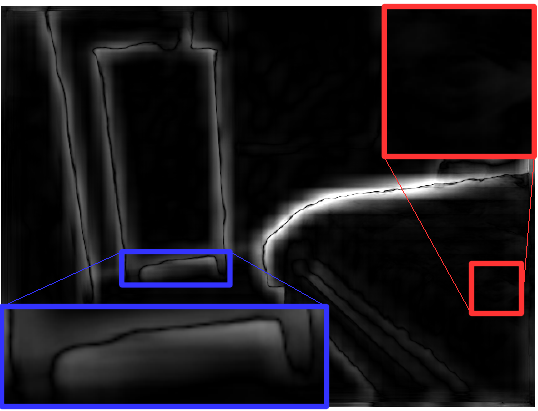}
\end{minipage}
}
\subfigure[\textbf{\textit{Our $\mathcal{F}_L(I)$}}]{
\begin{minipage}{3.25cm}
\centering
\includegraphics[width=1.1\linewidth,left]{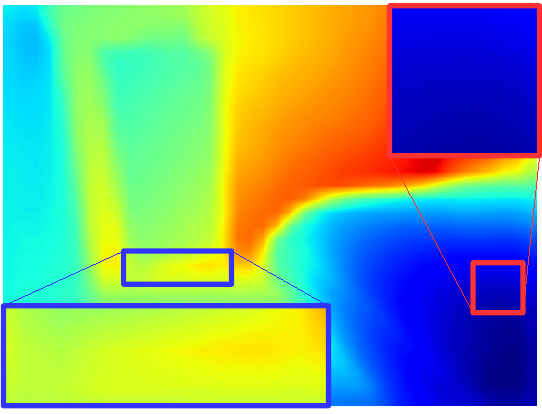}
\end{minipage}
}
\caption{\textbf{Comparison with SVLRM \cite{pan2019spatially}.} In SVLRM, $f_{\alpha}$ and $f_{\beta}$ are likely to learn the same structure information. In this case, $f_{\alpha}$ can't transfer the structure desired by $f_{\beta}$, resulting from over-smoothing of edges as shown in the regions marked by the blue boxes of (d) and (e). When the filtering output is determined by $f_{\beta}$, SVLRM behaves like DJF \cite{li2019joint}, causing the transfer of unwanted contents from guidance image to target image as shown in the regions marked by the red boxes of (c-e). By contrast, our approach resolve the problems of SVLRM by explicitly learning structure-transferring, leading to more desirable filtering results as shown in regions marked by the blue and red boxes of (h-j).
}
\label{exp_fig_svlrm}    
\end{figure*}

\textbf{Comparison with SVLRM \cite{pan2019spatially}.} Lastly, we compare to the state-of-the-art in deep guided filtering, namely SVLRM \cite{pan2019spatially}. Here, we analyze two drawbacks of SVLRM \cite{pan2019spatially}, as illustrated in Fig. \ref{exp_fig_svlrm} (c-e). First, $f_{\alpha}(I,G)$ and $f_{\beta}(I,G)$ is likely to learn similar structure information. This is because they share the same training dependencies; such as input, network architecture and objective function. As a result, $f_{\alpha}(I,G)$ can't transfer the desired structure from guidance $G$ to the output image of $f_{\beta}(I,G)$. Second, SVLRM behaves like DJF \cite{li2019joint} when the filtering performance is determined by $f_{\beta}(I,G)$. The implicit joint filtering causes slow convergence and the unwanted structures are transferred. By contrast, our approach focuses on estimating the amount function $f_a$ for explicit structure transfer, leading to more desirable filtering results, as illustrated in Fig. \ref{exp_fig_svlrm} (h-j). Fig. \ref{exp_fig_svlrm} (a) and (b) demonstrate the better performance of our approach compared to SVLRM \cite{pan2019spatially}, for both upsampling and denoising.

\textbf{Amount function $f_a$.} When we estimate the amount function $f_a$ through a convolutional neural network, the network architecture plays an important role in filtering performance. We have compared our filtering formulation with several baselines when $f_a$ is estimated by different networks used in these baselines. Generally, we found deep networks perform better than shallow networks, \eg~the network of SVLRM with a depth of 12 achieves an RMSE of 7.23 for upsampling ($16 \times$) and a PSNR of 40.27 for denoising ($\sigma = 50$), better than the RMSE of 8.19 and the PSNR of 39.59 obtained by the network of DJF with a depth of 6. One explanation for this is the fact that the deep network has more ability to express complex functions than shallow ones.

In Eq.(\ref{eq_new_form4}), we use the unsharp masks of the target image and guidance image as the input of the amount function network $f_a$. The raw target image and guidance image can also be the input. Next, we perform an experiment to study which input performs better. The network of DJF \cite{li2019joint} is used for $f_a$. On NYU Depth V2, using the unsharp mask as input achieves an RMSE of 8.19 for upsampling ($16 \times$) and a PSNR of 39.59 for denoising ($\sigma = 50$), better than the RMSE of 8.64 and the PSNR of 39.31 obtained by using the raw image as input. We find that using the unsharp mask as input not only achieves better filtering performance, but also convergences faster because network learning can focus on extracting the desired structure without the interference of redundant signals from the smooth basis of the image.

\begin{figure}
\centering 
\includegraphics[width=0.8\linewidth]{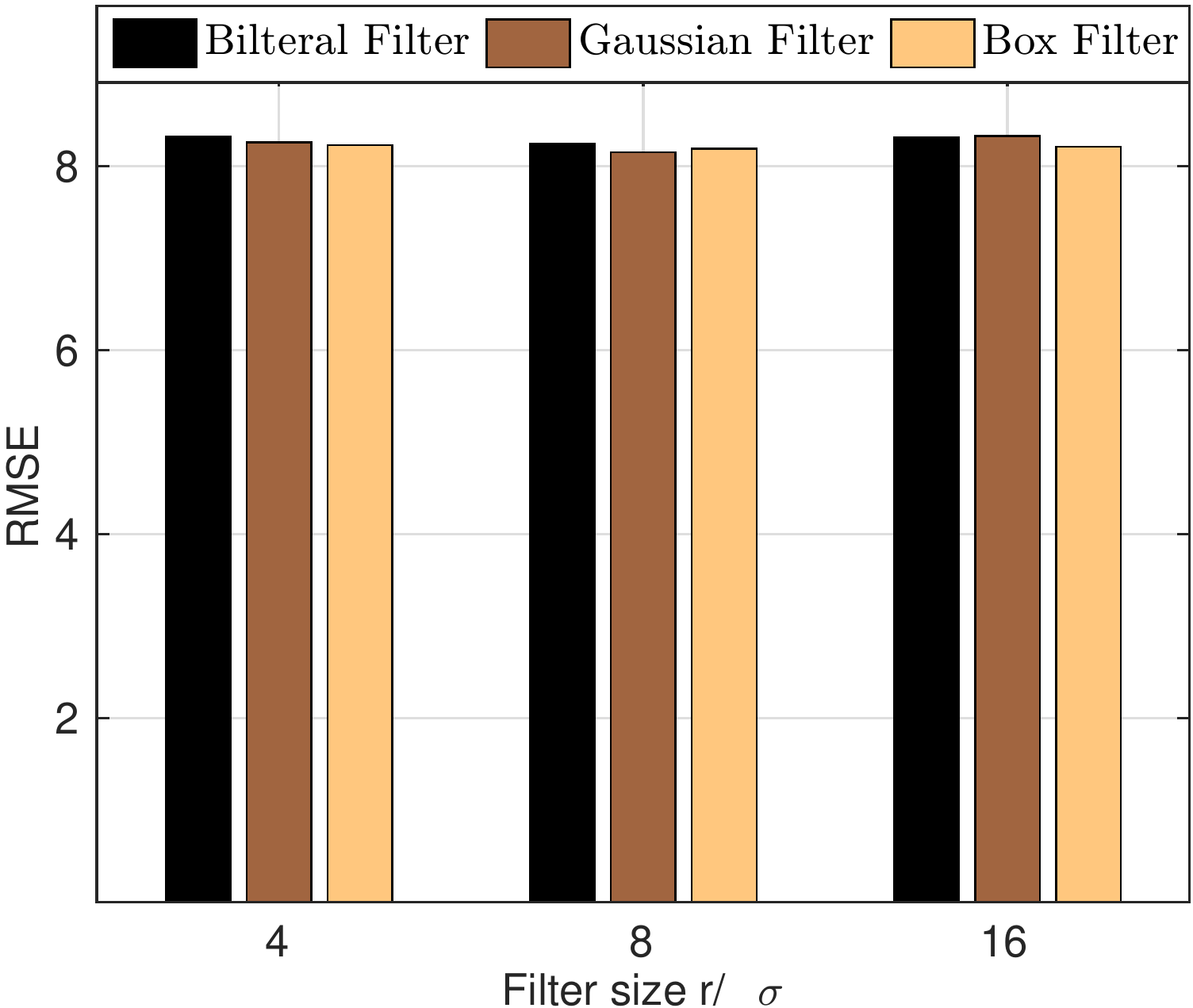}
\caption{\textbf{The effect of smoothing filter $\mathcal{F}_L$} on NYU Depth V2 for depth upsampling ($16 \times$). Our method is robust across smoothing filter type and size. }
\label{fig_dis}  
\end{figure}

\textbf{Smoothing filter $\mathcal{F}_L$.} To obtain the unsharp masks $G_m$ and $I_m$, we need a smoothing filter for $\mathcal{F}_L(I)$ and $\mathcal{F}_L(G)$. Next, we explore how the smoothing process affects the final filtering performance, we compare three different smoothing filters with different hyper-parameters on NYU Depth V2 for 16$\times$ depth upsampling. The hyper-parameter is the filtering size $r$ for the box filter, or the Gaussian variance $\sigma$ for the Gaussian and bilateral filters. We use three different values: $(4,8,16)$. The network of DJF \cite{li2016deep} is used for $f_a$. As shown in Fig. \ref{fig_dis}, our method is robust across filter type and size. We opt for the box filter with a filtering size of 8 throughout our experiments because it is simple, efficient and effective.
\subsection{Successive filtering network}
\label{sec:exp3}
Next, we investigate the effect of the network we designed for our successive filtering formulation. There are multiple amount blocks used in the successive filtering network. We explore how the number of amount blocks $L$ affects the filtering performance on NYU Depth V2 for depth upsampling (16$\times$) and depth denoising ($\sigma=50$). $\mathcal{F}_L$ is a box mean filter with radius $r=8$. For depth upsampling ($16 \times$), we first upsample the low-resolution depth image by bicubic interpolation to obtain the target resolution for $\mathcal{F}_L (I)$ and network learning. We make two observations from the results shown in Table \ref{tab_ablation_net}. First, our model's filtering performance is consistently improved when increasing $L$ from 1 to 5. Second, we can obtain multiple (L) filtering results by training a single network. Each filtering result is as good as the result obtained by an independently trained network. The model's filtering performance doesn't improve a lot when $L$ is increased from $4$ to $5$. Thus, we opt for $L=4$ in the following experiments. 
\begin{table*}[!ht]
\small
\caption{\textbf{Ablation studies for our network} on NYU Depth V2 for depth upsampling ($ 16\times $, RMSE) and denoising ($\sigma=50$, PSNR). our model's filtering performance is consistently improved when increasing $L$ from 1 to 5.}
\centering
\resizebox{1\linewidth}{!}{
\begin{tabular}{@{}lcccccccccccc@{}}
\toprule
 & \multicolumn{2}{c}{\textbf{L=1}} & \multicolumn{2}{c}{\textbf{L=2}} & \multicolumn{2}{c}{\textbf{L=3}} & \multicolumn{2}{c}{\textbf{L=4}} & \multicolumn{2}{c}{\textbf{L=5}}  \\
\cmidrule(lr){2-3} \cmidrule(lr){4-5} \cmidrule(lr){6-7} \cmidrule(lr){8-9} \cmidrule(lr){10-11}
& Upsampling $\downarrow$ & Denoising $\uparrow$& Upsampling $\downarrow$ & Denoising $\uparrow$ & Upsampling $\downarrow$ & Denoising $\uparrow$ & Upsampling $\downarrow$ & Denoising $\uparrow$ & Upsampling $\downarrow$ & Denoising $\uparrow$ \\
%& $16\times$ & $\sigma=50$ & $16\times$  & $\sigma=50$ \\
\hline
$\hat{I}^(1)$& 7.97 &40.08 & 8.03 & 39.95 & 8.05 &39.92 & 8.09 & 39.87 & 8.16 & 39.81 \\
$\hat{I}^(2)$& n.a. & n.a. & 6.76 & 40.93 & 6.75 &40.87 & 6.77 & 40.85 & 6.87 & 40.79 \\
$\hat{I}^(3)$& n.a. & n.a. & n.a. & n.a.  & 6.33 &41.27 & 6.28 & 41.25 & 6.32 & 41.21 \\
$\hat{I}^(4)$& n.a. & n.a. & n.a. & n.a.  & n.a. & n.a. & 6.07 & 41.53  & 6.09 & 41.49\\
$\hat{I}^(5)$& n.a. & n.a. & n.a. & n.a.  & n.a. & n.a. & n.a. & n.a.  & 6.02 & 41.61\\
\bottomrule
\end{tabular}}
\label{tab_ablation_net}
\end{table*}
\subsection{Performance analysis.} 
\label{sec:exp4}
Next, we analyze the performance of different deep guided filtering methods from three aspects: run-time performance, model parameters and filtering accuracy. For our methods, we use the successive filtering network with $L=4$. Thus, we can obtain four filtering models by training a single network, indicated by Ours($\hat{I}^{(1)}$), Ours($\hat{I}^{(2)}$), Ours($\hat{I}^{(3)}$) and Ours($\hat{I}^{(4)}$). We perform upsampling ($16 \times$) with all methods on the testing datasets (499 RGB/depth pairs) of NYU Depth V2. We perform all the testings on the same machine with an Intel Xeon E5-2640 2.20GHz CPU and an Nvidia GTX 1080 Ti GPU. The average run-time performance on 499 images with the size of $640 \times 480$ is reported in GPU mode with TensorFlow. From Table \ref{tab_analysis}, we can see that Ours($\hat{I}^{(1)}$) has the fewest parameters (17 k), and Ours($\hat{I}^{(4)}$) achieves the best filtering accuracy (6.07 RMSE). Ours($\hat{I}^{(1)}$) achieves a competitive average run-time performance (31 ms) compared to the best one achieved by DJF \cite{li2016deep} (29 ms).
\begin{table}[!ht]
\small
\caption{\textbf{Performance analysis.} on NYU Depth V2 for depth upsampling ($ 16\times $). Our filtering models achieve competitive performance in terms of run-time, model parameters and filtering accuracy.}
\centering
\resizebox{1\linewidth}{!}{
\begin{threeparttable}
\begin{tabular}{@{}lcccccccccccc@{}}
\toprule
 & Run-time (ms) $\downarrow$ & Parameters (k)$\downarrow$ & Accuracy (RMSE)$\downarrow$ \\
\hline
DMSG$^{\dagger}$& 36 &534 & 8.21  \\
DJF$^{\dagger}$& \textbf{29} &40 & 9.05  \\
DGF$^{\dagger}$& 34 &32 & 7.82  \\
SVLRM$^{\dagger}$& 47 &371 & 7.58  \\\midrule
Ours($\hat{I}^{(1)}$)& 31 &\textbf{17} & 8.09 \\
Ours($\hat{I}^{(2)}$)& 45 &38 & 6.77  \\
Ours($\hat{I}^{(3)}$)& 58 &59 & 6.28    \\
Ours($\hat{I}^{(4)}$)& 66 &85 & \textbf{6.07} \\
\bottomrule
\end{tabular}
  \end{threeparttable}
}
\footnotesize{$^{\dagger}$Results from our reimplementation under the same settings as this work.}
\label{tab_analysis}
\end{table}

\subsection{Depth and flow upsampling}
\label{sec:exp5}
\begin{figure*}
\centering 
\subfigure[\textbf{Guidance}]{
\begin{minipage}{4.1cm}
\centering
\includegraphics[width=1.05\linewidth,left]{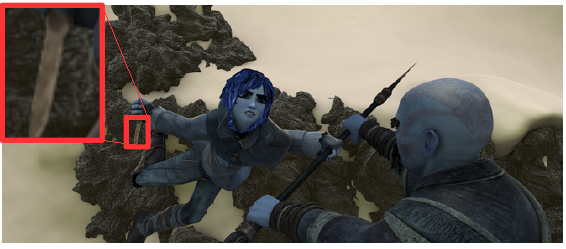}
\end{minipage}
}
\subfigure[\textbf{Ground truth}]{
\begin{minipage}{4.1cm}
\centering
\includegraphics[width=1.05\linewidth,left]{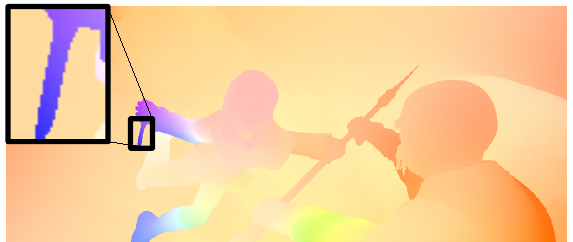}
\end{minipage}
}
\subfigure[\textbf{DMSG \cite{hui2016depth}}]{
\begin{minipage}{4.1cm}
\centering
\includegraphics[width=1.05\linewidth,left]{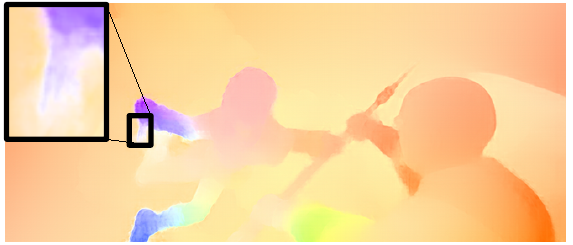}
\end{minipage}
}
\subfigure[\textbf{DGF \cite{wu2018fast}}]{
\begin{minipage}{4.1cm}
\centering
\includegraphics[width=1.05\linewidth,left]{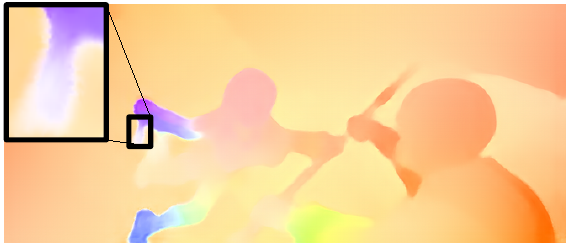}
\end{minipage}
}
\subfigure[\textbf{DJF \cite{li2019joint}}]{
\begin{minipage}{4.1cm}
\centering
\includegraphics[width=1.05\linewidth,left]{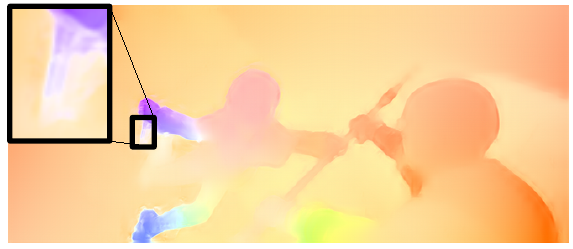}
\end{minipage}
}
\subfigure[\textbf{SVLRM \cite{pan2019spatially}}]{
\begin{minipage}{4.1cm}
\centering
\includegraphics[width=1.05\linewidth,left]{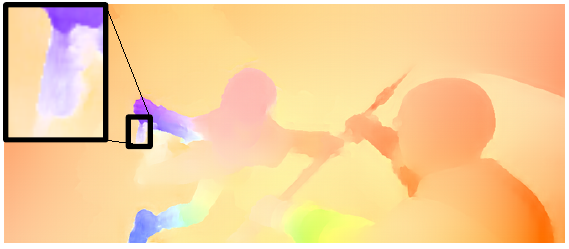}
\end{minipage}
}
\subfigure[\textbf{PAC \cite{pan2019spatially}}]{
\begin{minipage}{4.1cm}
\centering
\includegraphics[width=1.05\linewidth,left]{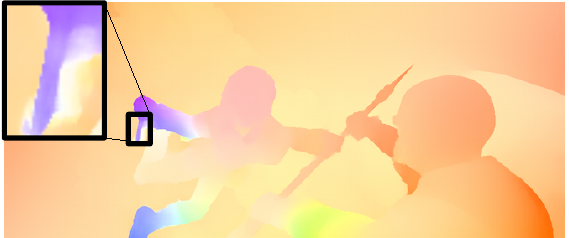}
\end{minipage}
}
\subfigure[\textbf{Ours}]{
\begin{minipage}{4.1cm}
\centering
\includegraphics[width=1.05\linewidth,left]{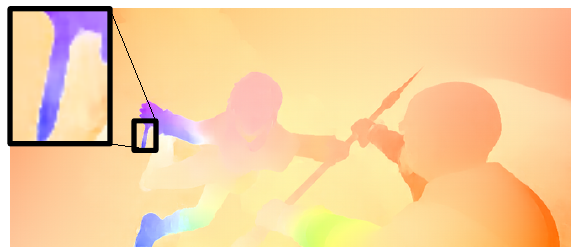}
\end{minipage}
}
\caption{\textbf{Optical flow upsampling} ($16 \times$) on Sintel. We are able to maintain sharp and thin edges.}
\label{fig_flow_16}    
\end{figure*}

Tables \ref{tab_up_depth} and \ref{tab_up_flow} show results for upsampling a depth image or optical flow image, under the guidance of its RGB image. 
We have noted that the existing works use different training settings and evaluation protocols. For fair comparison, we reimplement the main baseline methods under
our experimental settings. Our filter performs well, especially on Sintel and the larger upsampling scales on NYU Depth. Different from the related works \cite{hui2016depth, wu2018fast,li2019joint,pan2019spatially,Su_2019_CVPR,AlBahar_2019_ICCV}, our model learns an amount function $f_a$ to explicitly decide how to transfer the desired structure from guidance image to target image. Thus, our model can be more effective and efficient to learn the desired output.  Fig.~\ref{fig_flow_16} show our ability to better recover finer edges. %Fig.~\ref{fig_depth_16} and

\begin{table}[t]
\small
\caption{\textbf{Depth upsampling} for $ 2\times$, $ 4\times$, $8 \times$ and $16 \times$ on NYU Depth V2. The depth values are measured in centimeter, and a boundary with 6 pixels is excluded for evaluation. We outperform alternative filters for almost all settings.}
\centering
\resizebox{1\linewidth}{!}{
\begin{threeparttable}
\begin{tabular}{@{}lccccccccccccccccccc@{}}
\toprule
& \multicolumn{2}{c}{\textbf{$ 2\times $}} & \multicolumn{2}{c}{\textbf{$ 4\times $}} & \multicolumn{2}{c}{\textbf{$ 8\times $}} & \multicolumn{2}{c}{\textbf{$ 16\times $}} \\
\cmidrule(lr){2-3} \cmidrule(lr){4-5} \cmidrule(lr){6-7} \cmidrule(lr){8-9}
& RMSE $\downarrow$ & SSIM $\uparrow$ & RMSE $\downarrow$ & SSIM $\uparrow$ & RMSE $\downarrow$ & SSIM $\uparrow$ & RMSE $\downarrow$ & SSIM $\uparrow$ \\
\hline
%Bicubic &-&- &8.16&- &14.22&- &22.32&-\\
%GF \cite{he2012guided} &-&- &7.32&- &13.62&- &22.03&-\\
DMSG \cite{hui2016depth} &-&- &3.78&- &6.37&- &11.16&- \\
DJF \cite{li2019joint} &-&- &3.38&-  &5.86&- &10.11&- \\
bFT \cite{AlBahar_2019_ICCV} &-&- &3.35&- &5.73&- &9.01&- \\
PAC \cite{Su_2019_CVPR} &-&- &2.39&- &4.59&- &8.09&- \\
FWM \cite{xulearning} &-&- &2.16&- &4.32&- &7.66&- \\
SVLRM \cite{pan2019spatially} &-&- &\textbf{1.74}&- &5.59&- &7.23 &-\\
\hline
DMSG$^{\dagger}$ &2.12&0.9957 &3.43&0.9864 &4.19&0.9814 &8.21&0.9607\\
DGF$^{\dagger}$ &2.29&0.9940 &3.18&0.9897  &4.78&0.9776 &7.82&0.9568\\
DJF$^{\dagger}$ &1.37&0.9972 &2.85&0.9934  &4.48&0.9801 &9.05&0.9548 \\
SVLRM$^{\dagger}$ &1.28&0.9975 &2.62&0.9946 &3.96&0.9835 &7.58&0.9616 \\
\hline
Ours($\hat{I}^{(1)}$) &2.02&0.9963 &2.90&0.9925 &4.23&0.9839 &8.09&0.9563  \\
Ours($\hat{I}^{(2)}$) &1.65&0.9971 &2.61&0.9938  &3.82&0.9851 &6.77&0.9657  \\
Ours($\hat{I}^{(3)}$) &1.34&0.9974 &2.40&0.9940  &3.65&0.9857 &6.28&0.9690  \\
Ours($\hat{I}^{(4)}$) &\textbf{1.21}&\textbf{0.9976} &2.33&\textbf{0.9949}  &\textbf{3.58}&\textbf{0.9863} & \textbf{6.07}&\textbf{0.9706}\\
\bottomrule
\end{tabular}
  \end{threeparttable}
}
\footnotesize{$^{\dagger}$Results from our reimplementation under the same settings as this work.}
\label{tab_up_depth}
\end{table}
\begin{table}[t]
\small
\caption{\textbf{Flow upsampling} for $ 2\times$, $ 4\times$, $8 \times$ and $16 \times$ on Sintel. We outperform alternative filters for almost all settings.}
\centering
\resizebox{1\linewidth}{!}{
\begin{threeparttable}
\begin{tabular}{@{}lccccccccccccc@{}}
\toprule
& \multicolumn{2}{c}{\textbf{$ 2\times $}} & \multicolumn{2}{c}{\textbf{$ 4\times $}} & \multicolumn{2}{c}{\textbf{$ 8\times $}} & \multicolumn{2}{c}{\textbf{$ 16\times $}}\\
\cmidrule(lr){2-3} \cmidrule(lr){4-5} \cmidrule(lr){6-7} \cmidrule(lr){8-9}
& EPE $\downarrow$ & SSIM $\uparrow$ & EPE $\downarrow$ & SSIM $\uparrow$ & EPE $\downarrow$ & SSIM $\uparrow$ & EPE $\downarrow$ & SSIM $\uparrow$ \\
\hline
%Bicubic &-&- &0.32&- &0.57&- &1.53&-\\
%GF \cite{he2012guided} &-&- &0.42&- &0.72&- &1.38&-\\
DJF \cite{li2019joint} &-&- &0.18&-  &0.44&- &1.04&- \\
PAC \cite{Su_2019_CVPR} &-&- &0.11&- &0.26&- &0.59&- \\
FWM \cite{xulearning} &-&- &0.09&- &0.23&- &0.55&- \\
\hline
DMSG$^{\dagger}$ &0.14&0.9928 &0.24&0.9895 &0.41&0.9811 &0.96&0.9560\\
DGF$^{\dagger}$ &0.11&0.9942 &0.13&0.9934  &0.31&0.9842 &0.78&0.9692\\
DJF$^{\dagger}$ &0.10&0.9951 &0.17&0.9927  &0.43&0.9837 &1.04&0.9547 \\
SVLRM$^{\dagger}$ &0.09&0.9957 &0.16&0.9921 &0.36&0.9845 &0.98&0.9567 \\
\hline
Ours($\hat{I}^{(1)}$) &0.06&0.9988 &0.11&0.9936 &0.36&0.9851 &0.86&0.9684  \\
Ours($\hat{I}^{(2)}$) &0.05&0.9990 &0.07&0.9942  &0.29&0.9859 &0.68&0.9734  \\
Ours($\hat{I}^{(3)}$) &0.03&0.9991 &0.05&0.9943  &0.18&0.9864 &0.52&0.9754  \\
Ours($\hat{I}^{(4)}$) &\textbf{0.03}&\textbf{0.9991} &\textbf{0.04}&\textbf{0.9947}  &\textbf{0.16}&\textbf{0.9867} & \textbf{0.45}&\textbf{0.9773}\\
\bottomrule
\end{tabular}
  \end{threeparttable}
}
\footnotesize{$^{\dagger}$Results from our reimplementation under the same settings as this work.}
\label{tab_up_flow}
\end{table}
\begin{figure*}[!ht]
\centering 
\subfigure[\textbf{Guidance}]{
\begin{minipage}{4.1cm}
\centering
\includegraphics[width=1.05\linewidth,left]{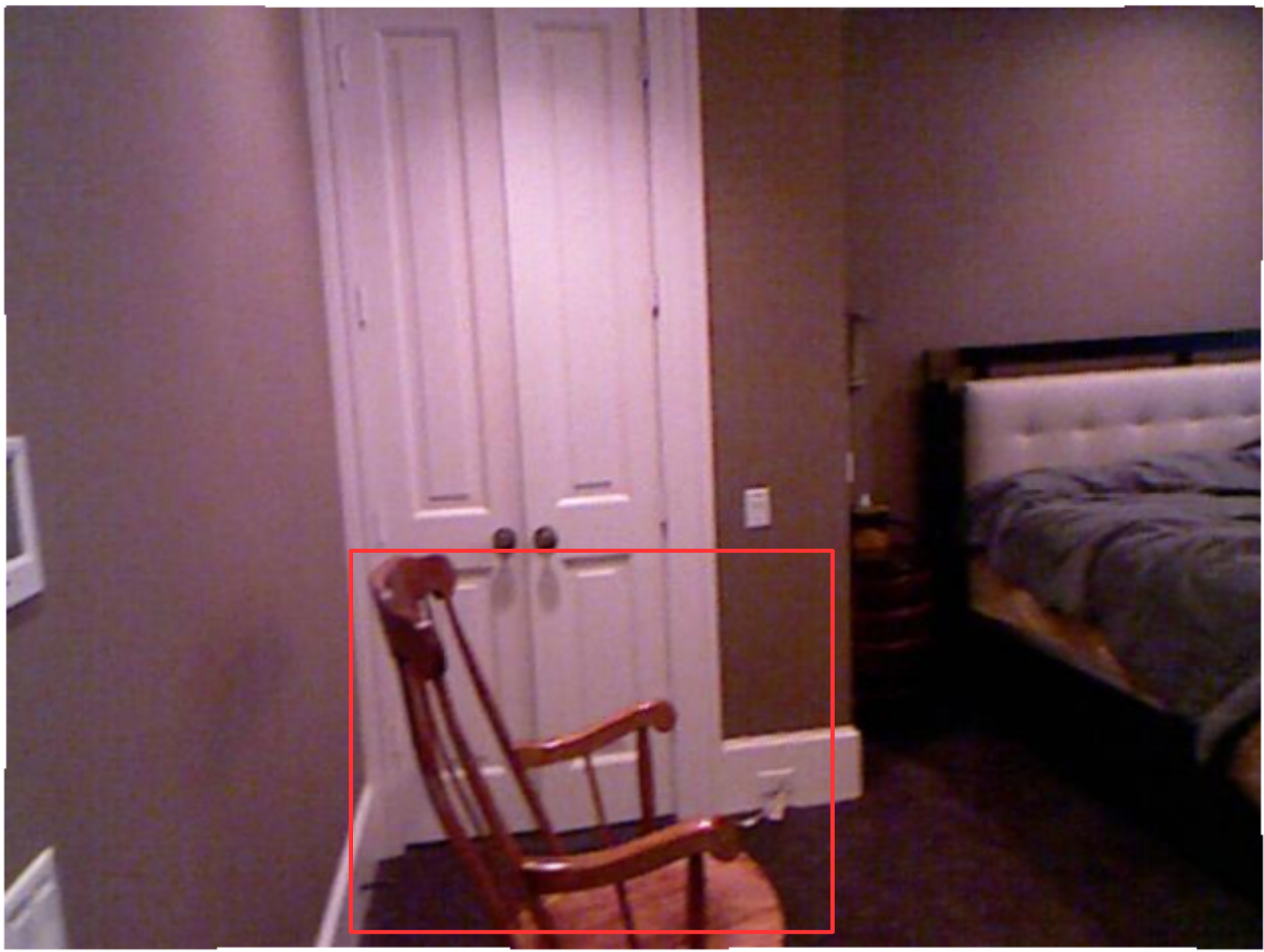}
\end{minipage}
}
\subfigure[\textbf{Target image}]{
\begin{minipage}{4.1cm}
\centering
\includegraphics[width=1.05\linewidth,left]{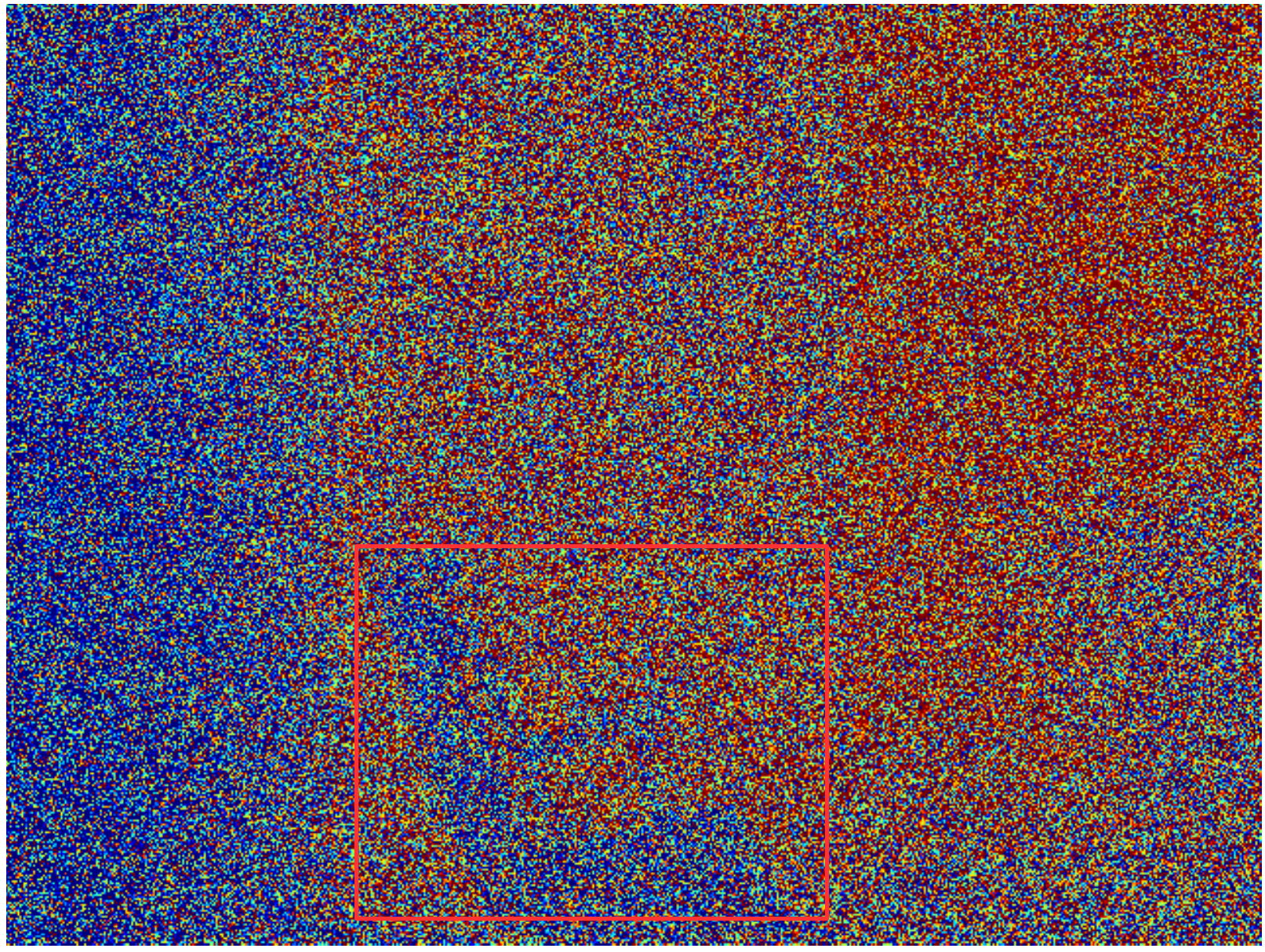}
\end{minipage}
}
\subfigure[\textbf{Ground truth}]{
\begin{minipage}{4.1cm}
\centering
\includegraphics[width=1.05\linewidth,left]{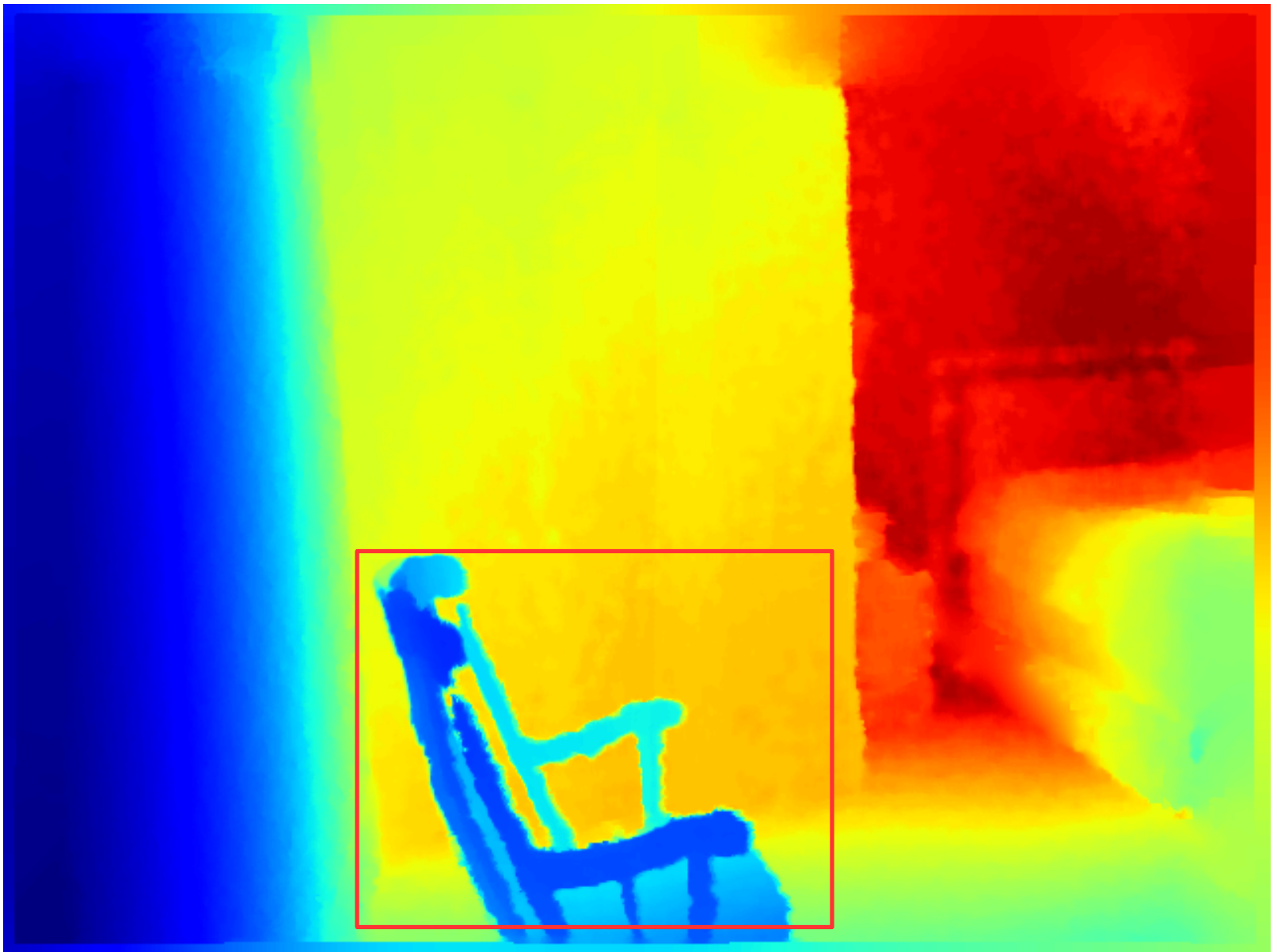}
\end{minipage}
}
\subfigure[\textbf{GF \cite{he2012guided}}]{
\begin{minipage}{4.1cm}
\centering
\includegraphics[width=1.05\linewidth,left]{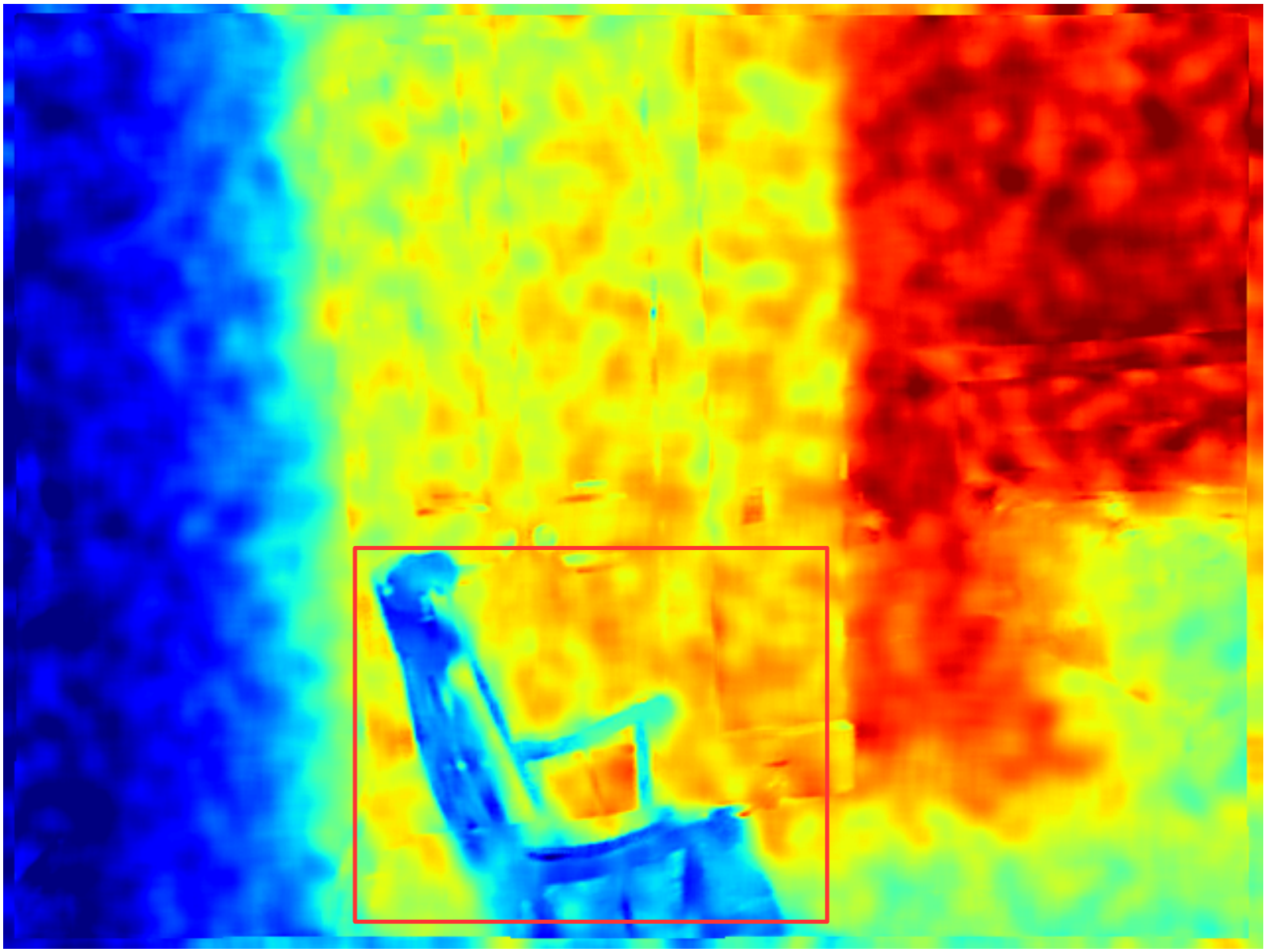}
\end{minipage}
}
\subfigure[\textbf{DGF \cite{wu2018fast}}]{
\begin{minipage}{4.1cm}
\centering
\includegraphics[width=1.05\linewidth,left]{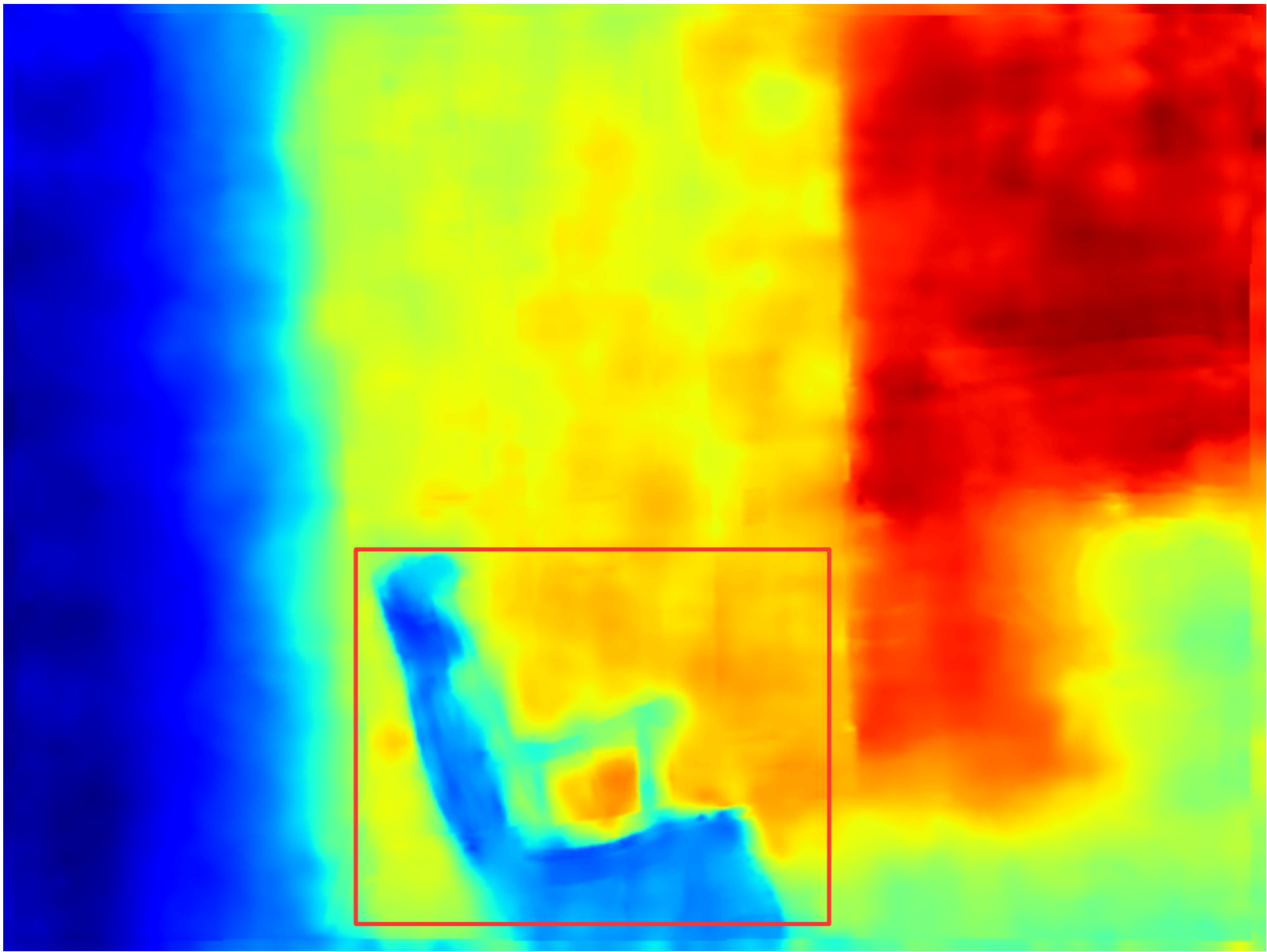}
\end{minipage}
}
\subfigure[\textbf{DJF \cite{li2019joint}}]{
\begin{minipage}{4.1cm}
\centering
\includegraphics[width=1.05\linewidth,left]{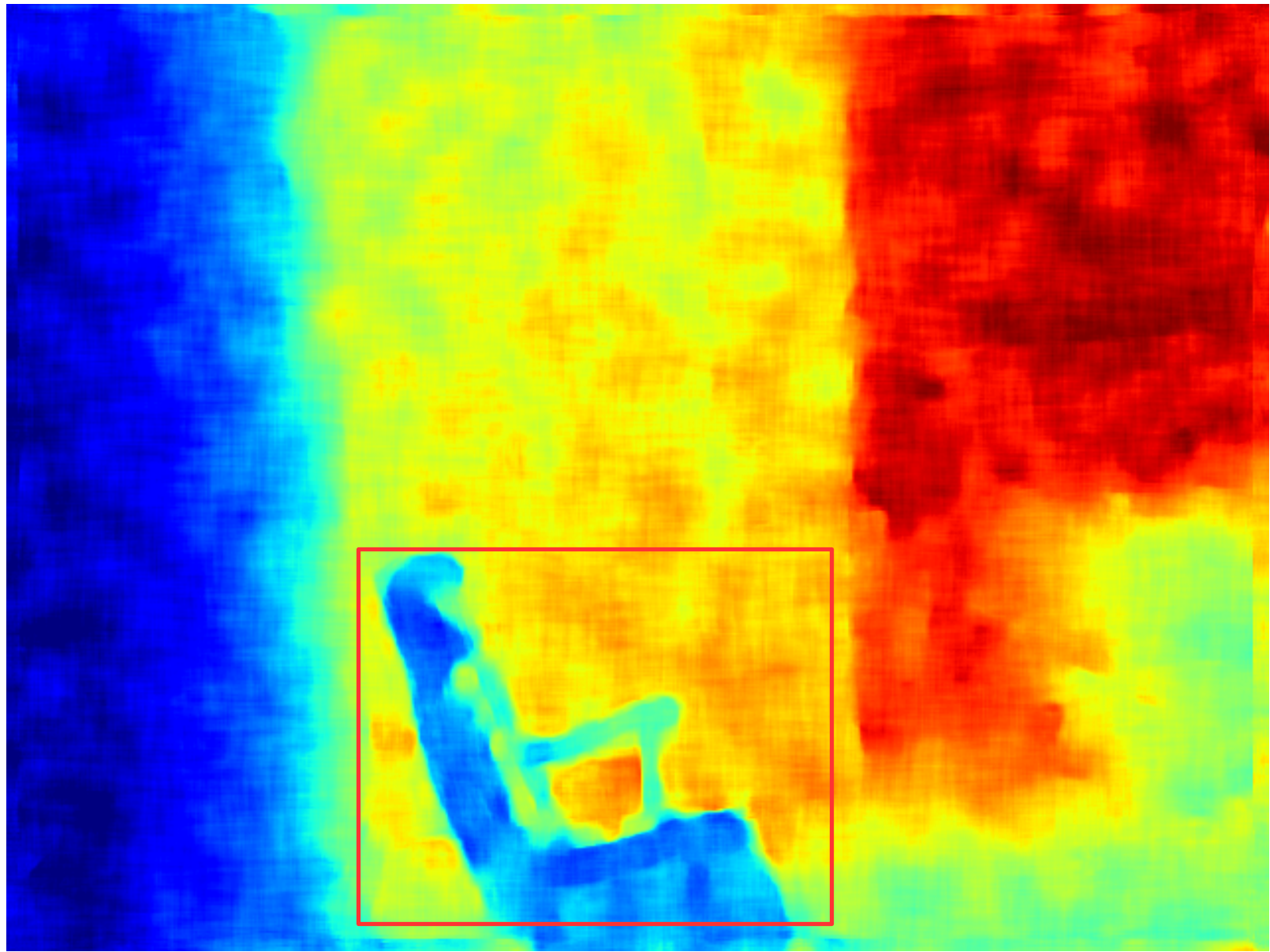}
\end{minipage}
}
\subfigure[\textbf{SVLRM \cite{pan2019spatially}}]{
\begin{minipage}{4.1cm}
\centering
\includegraphics[width=1.05\linewidth,left]{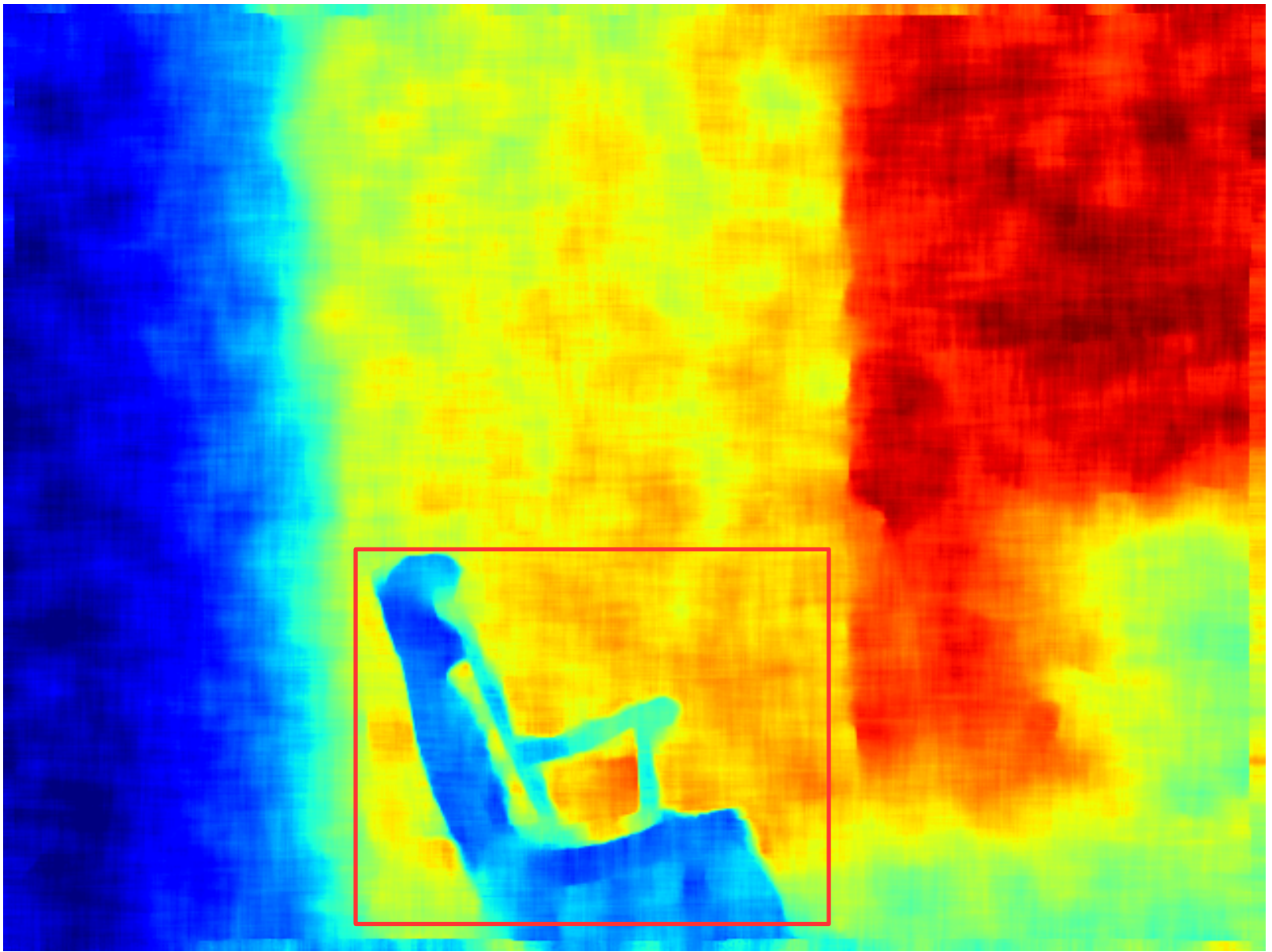}
\end{minipage}
}
\subfigure[\textbf{Ours($\hat{I}^{(4)}$)}]{
\begin{minipage}{4.1cm}
\centering
\includegraphics[width=1.05\linewidth,left]{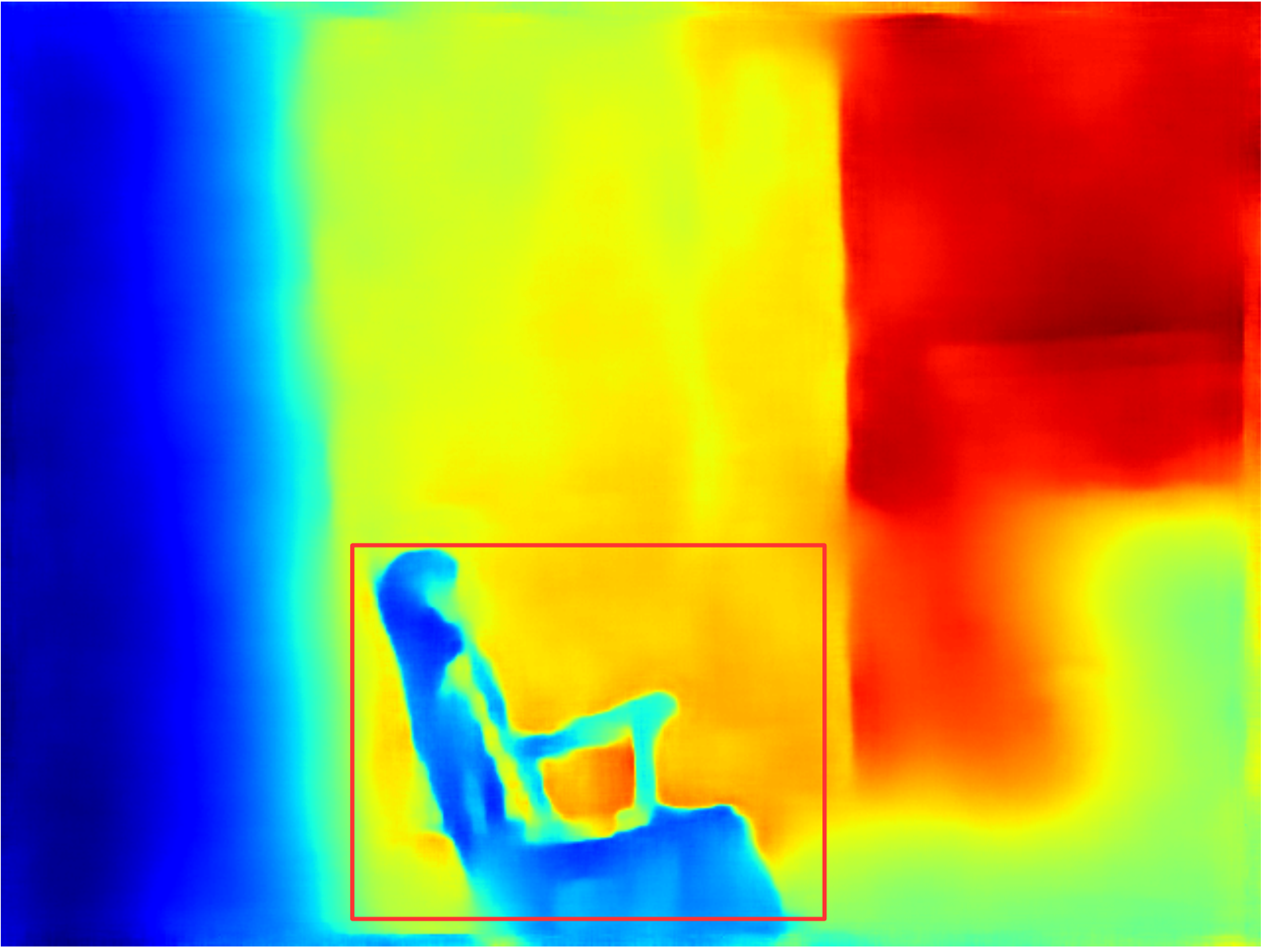}
\end{minipage}
}
\caption{\textbf{Depth denoising result} ($\sigma=50$) on NYU depth v2. Our filter better preserves edges and removes noise. }
\label{fig_depth_50}    
\end{figure*}
\begin{figure*}[!ht]
\centering 
\subfigure[\textbf{Target image}]{
\begin{minipage}{4.1cm}
\centering
\includegraphics[width=1.05\linewidth,left]{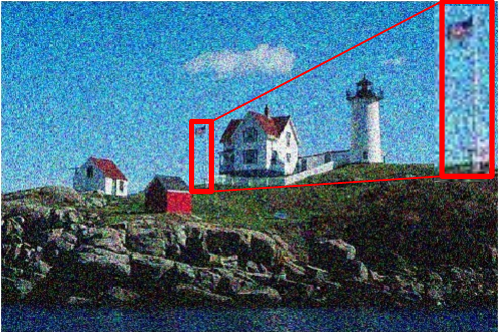}
\end{minipage}
}
\subfigure[\textbf{Ground truth}]{
\begin{minipage}{4.1cm}
\centering
\includegraphics[width=1.05\linewidth,left]{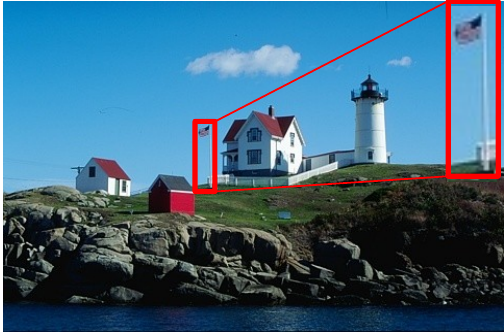}
\end{minipage}
}
\subfigure[\textbf{GF \cite{he2012guided}}]{
\begin{minipage}{4.1cm}
\centering
\includegraphics[width=1.05\linewidth,left]{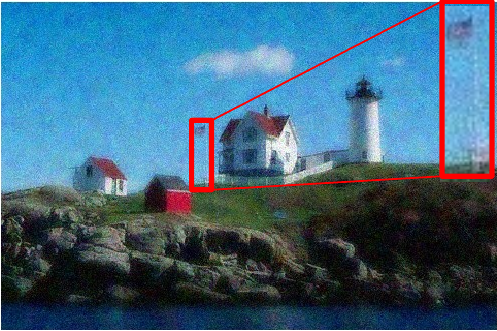}
\end{minipage}
}
\subfigure[\textbf{DGF \cite{wu2018fast}}]{
\begin{minipage}{4.1cm}
\centering
\includegraphics[width=1.05\linewidth,left]{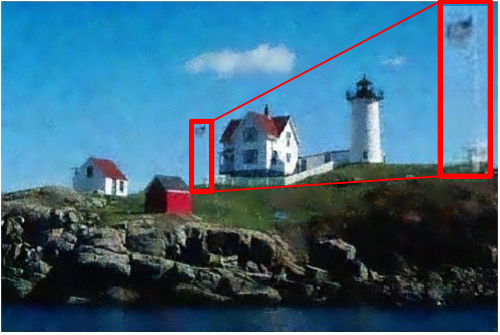}
\end{minipage}
}
\subfigure[\textbf{DnCNN \cite{zhang2017beyond}}]{
\begin{minipage}{4.1cm}
\centering
\includegraphics[width=1.05\linewidth,left]{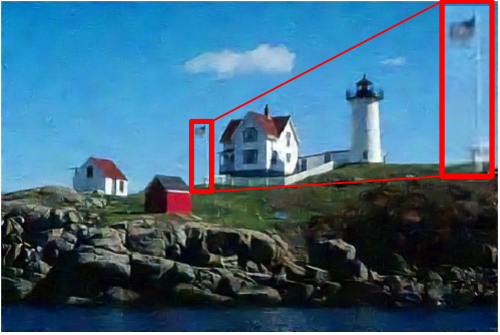}
\end{minipage}
}
\subfigure[\textbf{DJF \cite{li2019joint}}]{
\begin{minipage}{4.1cm}
\centering
\includegraphics[width=1.05\linewidth,left]{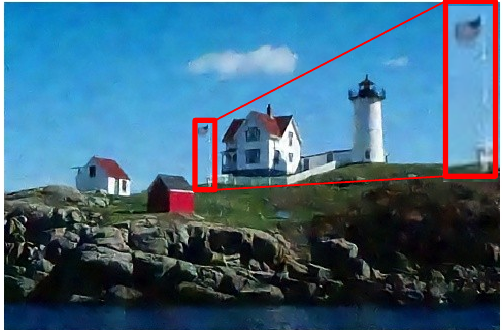}
\end{minipage}
}
\subfigure[\textbf{SVLRM \cite{pan2019spatially}}]{
\begin{minipage}{4.1cm}
\centering
\includegraphics[width=1.05\linewidth,left]{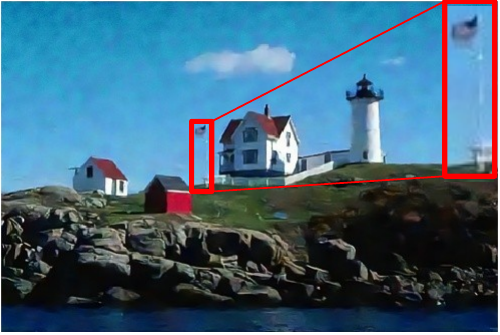}
\end{minipage}
}
\subfigure[\textbf{\textit{This paper}}]{
\begin{minipage}{4.1cm}
\centering
\includegraphics[width=1.05\linewidth,left]{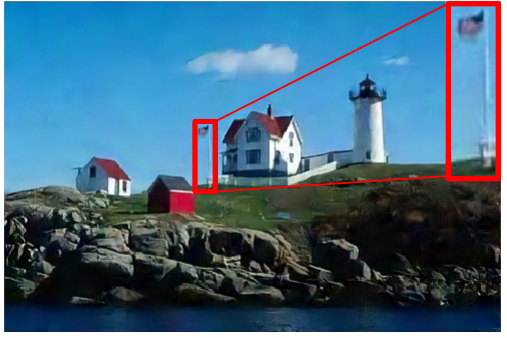}
\end{minipage}
}
\caption{\textbf{Denoising result} ($\sigma=50$) on BSDS500. Our filter better preserves finer edges with fewer noise artifacts.}
\label{fig_rgb_50}   
\end{figure*}
\subsection{Depth and natural image denoising}
\label{sec:exp6}
Our formulation also allows for standard filtering without a guidance image by simply making $G$ identical to $I$ in Eq.~(\ref{eq_successive}). Intuitively, such a setup defines a structure-preservation filter, while the guided variant defines a structure-transfer filter. Next, we evaluate the ability to remove Gaussian noise from depth and natural images. For depth image denoising, its RGB image is used as guidance.
For natural image denoising, the target and guidance image are the same RGB image. The quantitative results are shown in Tables \ref{tab_denoise_nyu} and \ref{tab_denoise_bsd}. We obtain the best PSNR and SSIM scores on both datasets for all three noise levels. Fig. \ref{fig_depth_50} and Fig. \ref{fig_rgb_50} highlight our ability to better remove noise and preserve finer edges compared to other filters.
\begin{table}[t]
\small
\caption{\textbf{Depth image denoising} on NYU Depth V2. Our filter achieves the best results for all settings.}
\centering
\resizebox{1\linewidth}{!}{
\begin{threeparttable}
\begin{tabular}{@{}lcccccccccc@{}}
\toprule
 & \multicolumn{2}{c}{\textbf{$ \sigma=15 $}} & \multicolumn{2}{c}{\textbf{$ \sigma=25 $}} & \multicolumn{2}{c}{\textbf{$ \sigma=50 $}}\\
\cmidrule(lr){2-3} \cmidrule(lr){4-5} \cmidrule(lr){6-7} 
& PSNR $\uparrow$ & SSIM $\uparrow$ & PSNR $\uparrow$ & SSIM $\uparrow$ & PSNR $\uparrow$ & SSIM $\uparrow$ \\
\hline
%GF$^{\dagger}$ &43.16&-  &41.25&- &37.25&-  \\
DGF$^{\dagger}$ &45.52&0.9650  &43.96&0.9579 &40.15&0.9422  \\
DJF$^{\dagger}$ &46.06&0.9633  &43.58&0.9462 &39.24&0.8826 \\
SVLRM$^{\dagger}$ &47.35&0.9722 &44.38&0.9524 &39.84&0.8891  \\
\hline
Ours($\hat{I}^{(1)}$) &46.02&0.9627 &43.62&0.9474 &39.87&0.9112   \\
Ours($\hat{I}^{(2)}$) &46.92&0.9704 &44.53&0.9586 &40.85&0.9310  \\
Ours($\hat{I}^{(3)}$) &47.30&0.9732 &44.93&0.9635 &41.25&0.9422  \\
Ours($\hat{I}^{(4)}$) &\textbf{47.45}&\textbf{0.9750}  &\textbf{45.09}&\textbf{0.9664} &\textbf{41.53}&\textbf{0.9488}  \\
\bottomrule
\end{tabular}
  \end{threeparttable}
}
\footnotesize{$^{\dagger}$Results from our reimplementation under the same settings as this work.}
\label{tab_denoise_nyu}
\end{table}
\begin{table}[t]
\small
\caption{\textbf{Natural image denoising} on BSDS500. Our filter achieves the best results for all settings.}
\centering
\resizebox{1\linewidth}{!}{
\begin{threeparttable}
\begin{tabular}{@{}lcccccccccc@{}}
\toprule
 & \multicolumn{2}{c}{\textbf{$ \sigma=15 $}} & \multicolumn{2}{c}{\textbf{$ \sigma=25 $}} & \multicolumn{2}{c}{\textbf{$ \sigma=50 $}}\\
\cmidrule(lr){2-3} \cmidrule(lr){4-5} \cmidrule(lr){6-7} 
& PSNR $\uparrow$ & SSIM $\uparrow$ & PSNR $\uparrow$ & SSIM $\uparrow$ & PSNR $\uparrow$ & SSIM $\uparrow$ \\
\hline
%GF$^{\dagger}$ &28.37&-  &24.02&- &22.57&-  \\
DnCNN$^{\dagger}$ &33.12&0.9166  &30.48&0.8618 &27.10&0.7495  \\
DGF$^{\dagger}$ &31.89&0.9010  &29.56&0.8409 &26.31&0.7229  \\
DJF$^{\dagger}$ &33.10&0.9168  &30.49&0.8646 &27.09&0.7496 \\
SVLRM$^{\dagger}$ &33.01&0.9202 &30.60&0.8712 &27.37&0.7686  \\
\hline
Ours($\hat{I}^{(1)}$) &33.41&0.9272 &30.80&0.8758 &27.43&0.7716   \\
Ours($\hat{I}^{(2)}$) &33.65&0.9341 &31.05&0.8868 &27.73&0.7829  \\
Ours($\hat{I}^{(3)}$) &33.76&0.9354 &31.16&0.8890 &27.87&0.7895  \\
Ours($\hat{I}^{(4)}$) &\textbf{33.79}&\textbf{0.9361}  &\textbf{31.19}&\textbf{0.8906} &\textbf{27.91}&\textbf{0.7938}  \\
\bottomrule
\end{tabular}
  \end{threeparttable}
}
\footnotesize{$^{\dagger}$Results from our reimplementation under the same settings as this work.}
\label{tab_denoise_bsd}
\end{table}
\begin{table}[!ht]
\small
\caption{\textbf{Cross-modality filtering} for joint upsampling ($ 4\times $) on bw/color and RGB/saliency pairs. Our filter achieves the best results for all settings.}
\centering
\resizebox{1\linewidth}{!}{
\begin{threeparttable}
\begin{tabular}{@{}lcccccc@{}}
\toprule
 & \multicolumn{2}{c}{\textbf{bw/color}} & \multicolumn{2}{c}{\textbf{RGB/saliency}} \\
\cmidrule(lr){2-3} \cmidrule(lr){4-5}
& RMSE $\downarrow$ & SSIM $\uparrow$ & F-measure $\uparrow$ & SSIM $\uparrow$ \\
\hline
%Bicubic/input$^{\star}$ &12.78&-  &0.681&-    \\
GF$^{\dagger}$ &11.51&0.6054  &0.685&0.5365    \\
DGF$^{\dagger}$ &11.17&0.6041  &0.701&0.5431    \\
DJF$^{\dagger}$ &10.96&0.6046 &0.697&0.5378  \\
SVLRM$^{\dagger}$ &10.78&0.6074  &0.699&0.4588    \\
\textbf{Ours} &\textbf{10.39}&\textbf{0.6095}  &\textbf{0.705}&\textbf{0.6087} \\
\bottomrule
\end{tabular}
  \end{threeparttable}
  }
  \footnotesize{$^{\dagger}$Results from our reimplementation under the same settings as this work.}
\label{tab_modality_up}
\end{table}
\begin{table}[!ht]
\small
\caption{\textbf{Cross-modality filtering} for joint denoising ($\sigma=25$) on Flash/no-flash and RGB/NIR pairs. Our filter achieves the best results for all settings.}
\centering
\resizebox{1\linewidth}{!}{
\begin{threeparttable}
\begin{tabular}{@{}lccccc@{}}
\toprule
 & \multicolumn{2}{c}{\textbf{Flash/no-flash}} & \multicolumn{2}{c}{\textbf{RGB/NIR}} \\
\cmidrule(lr){2-3} \cmidrule(lr){4-5}
& PSNR $\uparrow$ & SSIM $\uparrow$ & PSNR $\uparrow$ & SSIM $\uparrow$\\
\hline
%Bicubic/input$^{\star}$ &20.17&-  &20.17&-    \\
GF$^{\dagger}$ &29.43&0.7675  &27.22&0.6971    \\
DGF$^{\dagger}$ &28.11&0.7246 &26.40&0.6436   \\
DJF$^{\dagger}$ &29.53&0.7407 &27.56&0.6826  \\
SVLRM$^{\dagger}$ &30.27&0.7584 &28.33&0.7084   \\
\textbf{Ours} &\textbf{30.76}&\textbf{0.7699}  &\textbf{28.95}&\textbf{0.7226} \\
\bottomrule
\end{tabular}
  \end{threeparttable}
}
\footnotesize{$^{\dagger}$Results from our reimplementation under the same settings as this work.}
\label{tab_modality_denoise}
\end{table}
\subsection{Cross-modality filtering}
\label{sec:exp7}
Finally, we demonstrate that our models trained on one modality can be directly applied to other modalities. Here, we use the models trained with RGB/depth image pairs for the joint upsampling of bw/color and RGB/saliency image pairs, and the joint denoising of RGB/NIR and flash/no-flash image pairs. For our method, we use the model of $Ours(\hat{I}^{(4)})$. Following \cite{li2019joint}, for the multi-channel target image, \ie,~no-flash image, we apply the trained models independently for each channel. For the single-channel guidance image, \ie,~NIR image, we replicate it three times to obtain a 3-channel guidance image.

\textbf{Joint upsampling.}
To speed up the translation from one image to another image, one strategy is to perform translation at a coarse resolution and then upsample the low-resolution solution back to the original one with a joint image upsampling filter. Here, we demonstrate that our models act as joint upsampling filters well on bw/color and RGB/saliency image pairs translation tasks. For bw/color translation, we use 68 bw images from BSD68 dataset \cite{roth2009fields}, and the colorization model proposed by Lizuka \etal \cite{iizuka2016let} is used as translation model. For RGB/saliency translation, we use 1000 RGB image from ECSSD dataset \cite{shi2015hierarchical}, and the saliency region detection model proposed by Hou \etal \cite{hou2017deeply} as translation model. The input images, \ie, bw images and RGB images, are first downsampled by a factor of $4 \times$ using nearest-neighbor interpolation, and then are feed into the translation models to generate the output images. After that, we recover the output images to the original resolution under the guidance of the original input images by various joint upsamling methods. Table \ref{tab_modality_up} shows the quantitative results, and we can see that our model achieves the best performance for both two tasks. The joint upsampling pipeline performs more than two times faster than direct translation on the GPU mode. We also provide the qualitative results in Fig. \ref{fig_bw} and \ref{fig_saliency} to show that the proposed model better recovers finer details.

\textbf{Joint denoising.}
We introduce two datasets for the joint denoising experiments. \textbf{Flash/no-flash} \cite{FNF_Saliency_ECCV2014} consists of 120 image pairs, where the no-flash image is used for denoising under the guidance of its flash version \cite{petschnigg2004digital,he2012guided,yan2013cross, li2019joint}. \textbf{Nirscene1} \cite{BS11} consists of 477 RGB/NIR image pairs in 9 categories, where the RGB image is used for denoising under the guidance of its NIR version \cite{yan2013cross, wang2019near, li2019joint}. Table \ref{tab_modality_denoise} shows that our filter has a better denoising ability than alternatives. Fig.~\ref{fig_flash} and Fig.~\ref{fig_nir} provide qualitative results, which shows that our model better preserves important structures while removing noises.
\section{Conclusion}
In this paper, we have introduced a new and simplified guided filter. With inspiration from unsharp masking, our proposed formulation only requires estimating a single coefficient, in contrast to the two entangled coefficients required in current approaches.
Based on the proposed formulation, we introduce a successive guided filtering network. Our network allows for a trade-off between accuracy and efficiency by choosing different filtering results during inference. 
Experimentally, we find that the proposed filtering method better preserves sharp and thin edges, while avoiding unwanted structures transferred from guidance. 
We furthermore show that our approach is effective for various applications such as upsampling, denoising, and cross-modal filtering.

\afterpage{
\begin{figure}[t!]
\centering 
\subfigure[\textbf{Guidance image}]{
\begin{minipage}{4.08cm}
\centering
\includegraphics[width=1.05\linewidth,left]{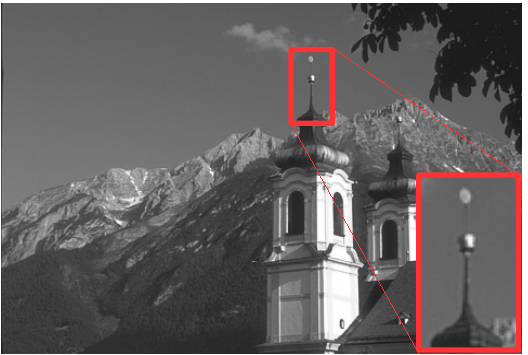}
\end{minipage}
}
\subfigure[\textbf{Target image}]{
\begin{minipage}{4.08cm}
\centering
\includegraphics[width=1.05\linewidth,left]{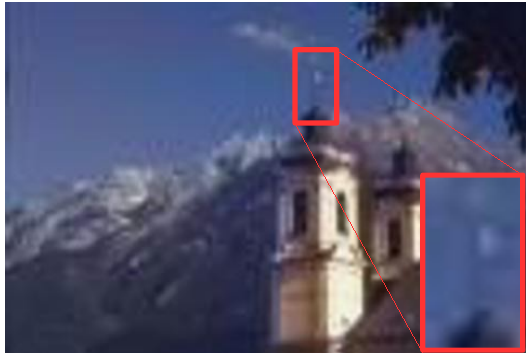}
\end{minipage}
}
\subfigure[\textbf{Ground truth\cite{iizuka2016let}}]{
\begin{minipage}{4.08cm}
\centering
\includegraphics[width=1.05\linewidth,left]{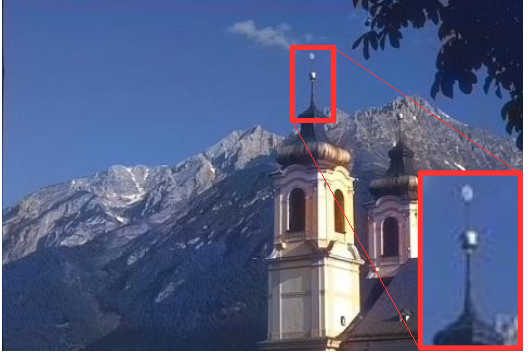}
\end{minipage}
}
\subfigure[\textbf{GF \cite{he2012guided}}]{
\begin{minipage}{4.08cm}
\centering
\includegraphics[width=1.05\linewidth,left]{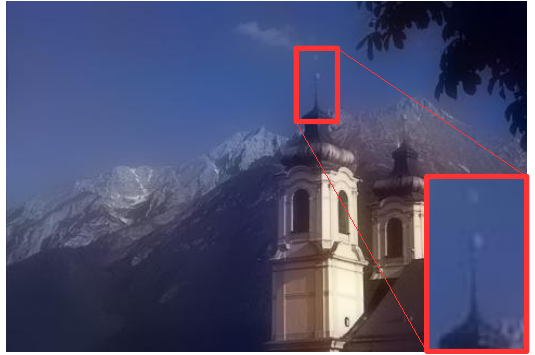}
\end{minipage}
}
\subfigure[\textbf{SVLRM \cite{pan2019spatially}}]{
\begin{minipage}{4.08cm}
\centering
\includegraphics[width=1.05\linewidth,left]{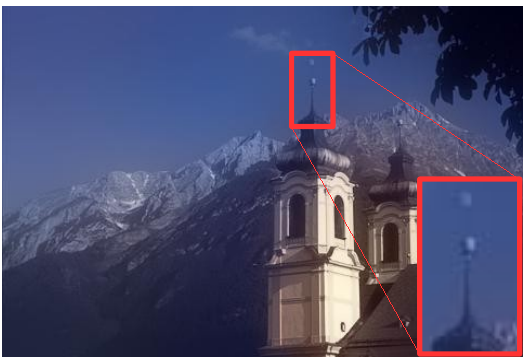}
\end{minipage}
}
\subfigure[\textbf{\textit{This paper}}]{
\begin{minipage}{4.08cm}
\centering
\includegraphics[width=1.05\linewidth,left]{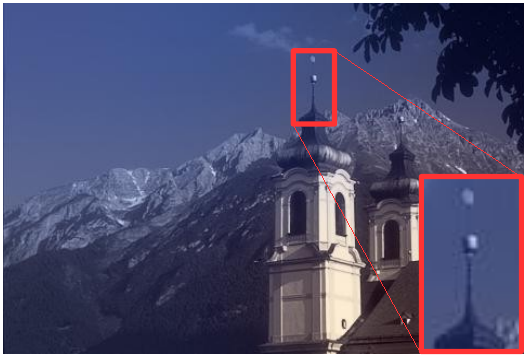}
\end{minipage}
}
\caption{\textbf{Joint upsampling result} ($4 \times$) on bw/color. Our filter recovers more desirable details.}
\label{fig_bw}   
\end{figure}
\begin{figure}[!ht]
\centering 
\subfigure[\textbf{Guidance image}]{
\begin{minipage}{4.03cm}
\centering
\includegraphics[width=1.05\linewidth,left]{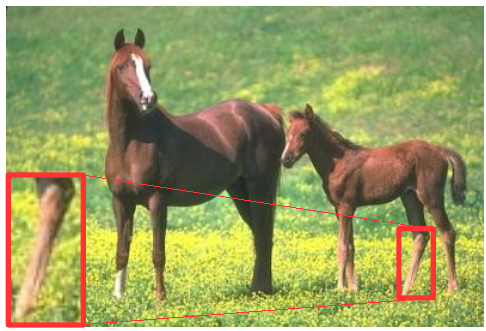}
\end{minipage}
}
\subfigure[\textbf{Target image}]{
\begin{minipage}{4.03cm}
\centering
\includegraphics[width=1.05\linewidth,left]{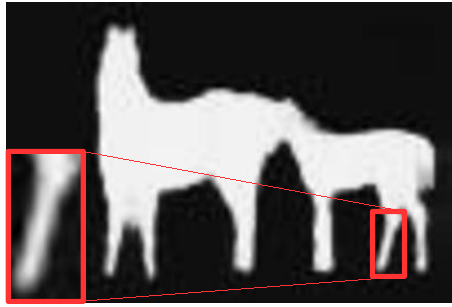}
\end{minipage}
}
\subfigure[\textbf{Ground truth \cite{hou2017deeply}}]{
\begin{minipage}{4.03cm}
\centering
\includegraphics[width=1.05\linewidth,left]{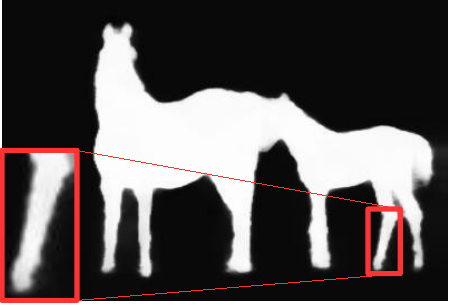}
\end{minipage}
}
\subfigure[\textbf{GF \cite{he2012guided}}]{
\begin{minipage}{4.03cm}
\centering
\includegraphics[width=1.05\linewidth,left]{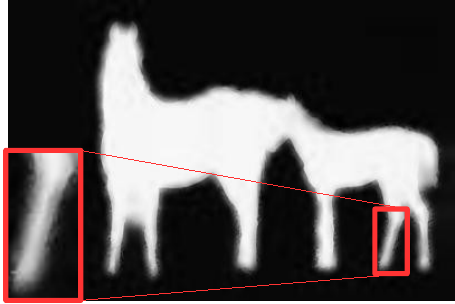}
\end{minipage}
}
\subfigure[\textbf{SVLRM \cite{pan2019spatially}}]{
\begin{minipage}{4.03cm}
\centering
\includegraphics[width=1.05\linewidth,left]{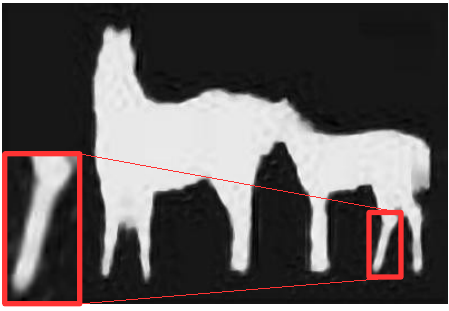}
\end{minipage}
}
\subfigure[\textbf{\textit{This paper}}]{
\begin{minipage}{4.03cm}
\centering
\includegraphics[width=1.05\linewidth,left]{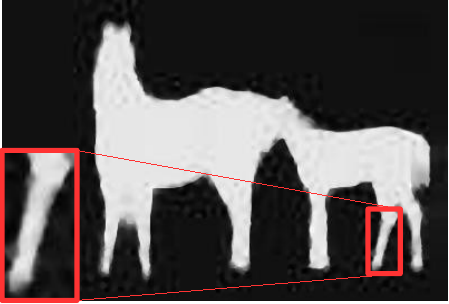}
\end{minipage}
}
\caption{\textbf{Joint upsampling result} ($4 \times$) on RGB/saliency. Our filter recovers sharper edges.}
\label{fig_saliency}   
\end{figure}
\begin{figure}[t!]
\centering 
\subfigure[\textbf{Guidance image}]{
\begin{minipage}{4.1cm}
\centering
\includegraphics[width=1.05\linewidth,left]{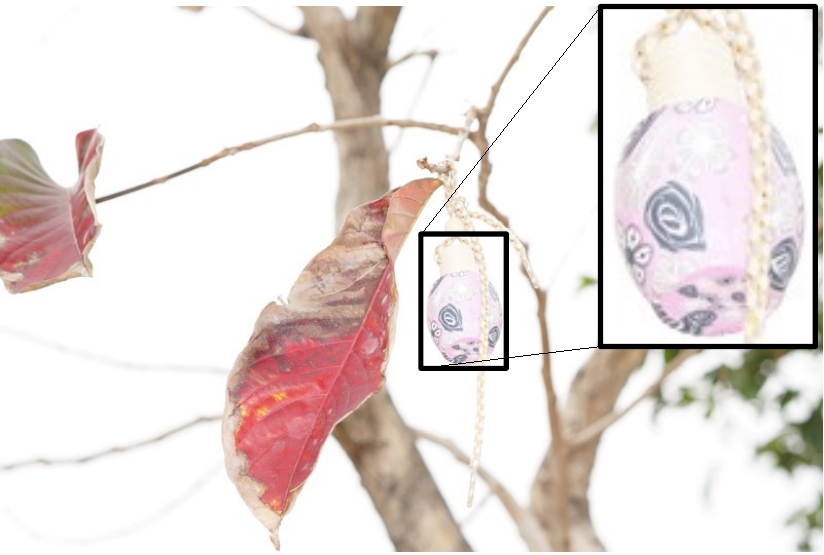}
\end{minipage}
}
\subfigure[\textbf{Target image}]{
\begin{minipage}{4.1cm}
\centering
\includegraphics[width=1.05\linewidth,left]{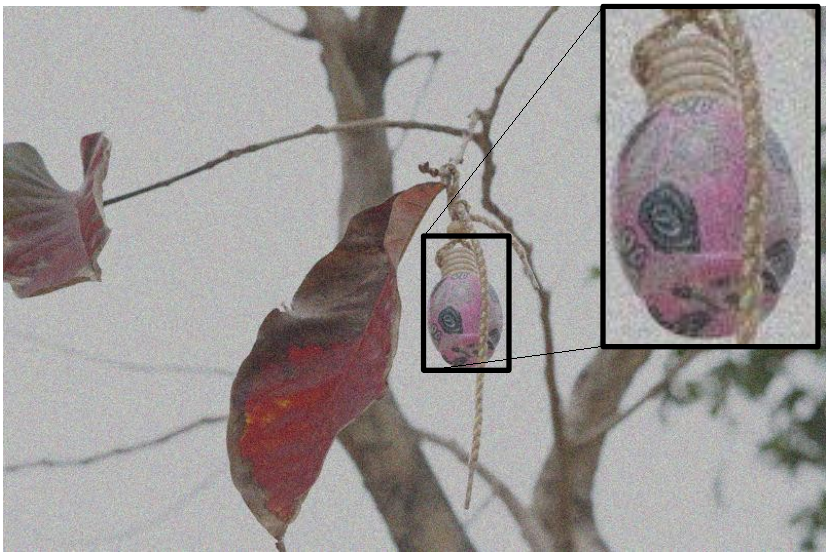}
\end{minipage}
}
\subfigure[\textbf{Ground truth}]{
\begin{minipage}{4.1cm}
\centering
\includegraphics[width=1.05\linewidth,left]{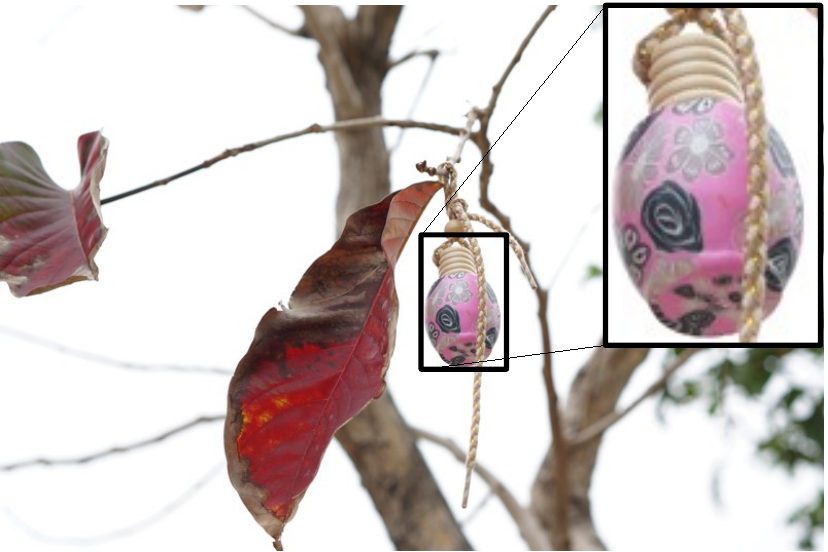}
\end{minipage}
}
\subfigure[\textbf{GF \cite{he2012guided}}]{
\begin{minipage}{4.1cm}
\centering
\includegraphics[width=1.05\linewidth,left]{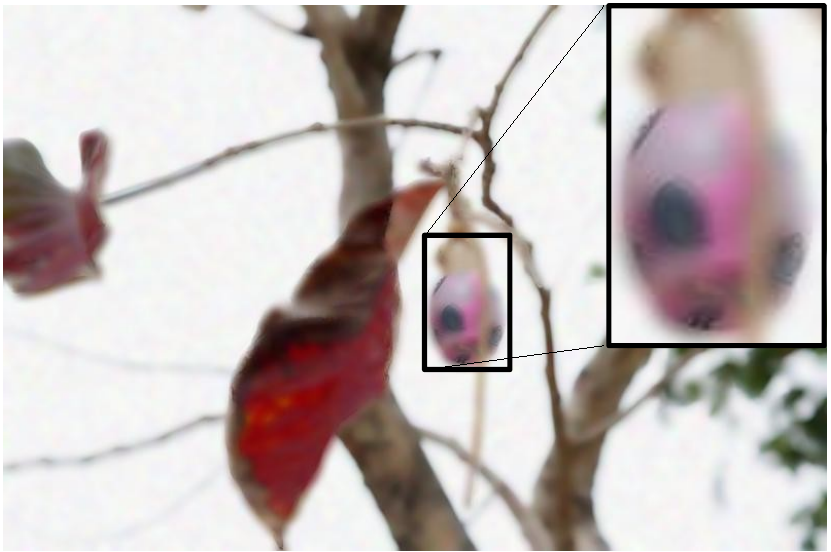}
\end{minipage}
}
\subfigure[\textbf{SVLRM \cite{pan2019spatially}}]{
\begin{minipage}{4.1cm}
\centering
\includegraphics[width=1.05\linewidth,left]{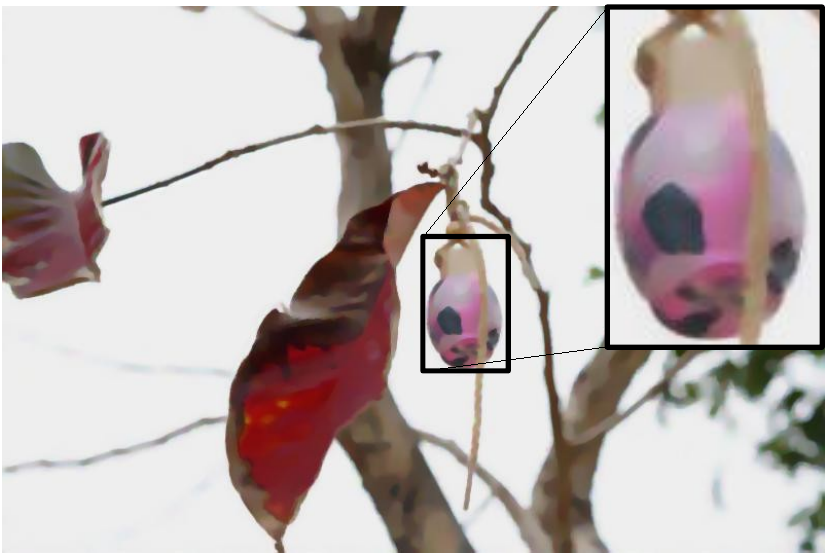}
\end{minipage}
}
\subfigure[\textbf{\textit{This paper}}]{
\begin{minipage}{4.1cm}
\centering
\includegraphics[width=1.05\linewidth,left]{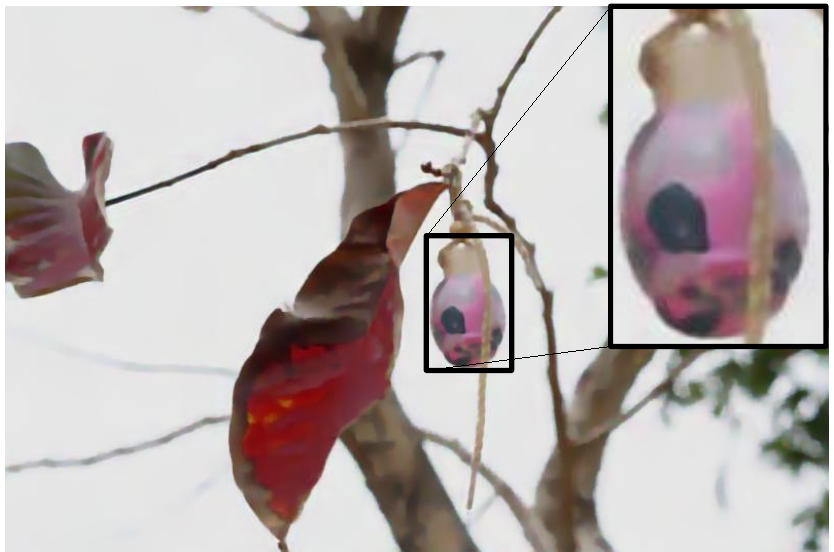}
\end{minipage}
}
\caption{\textbf{Joint denoising result} ($\sigma=25$) on Flash/no-flash. Our filter better removes noises with less blurring.\\}
\label{fig_flash}   
\end{figure}
\begin{figure}[h!]
\centering 
\subfigure[\textbf{Guidance image}]{
\begin{minipage}{4.04cm}
\centering
\includegraphics[width=1.05\linewidth,left]{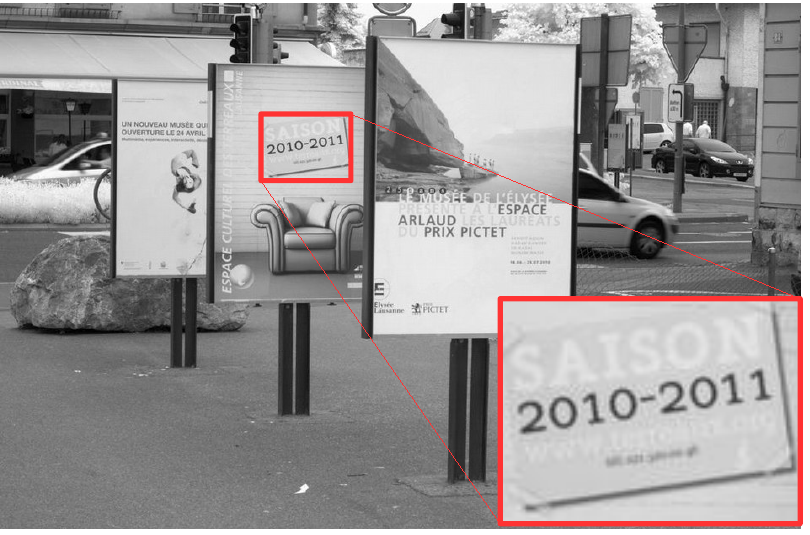}
\end{minipage}
}
\subfigure[\textbf{Target image}]{
\begin{minipage}{4.04cm}
\centering
\includegraphics[width=1.05\linewidth,left]{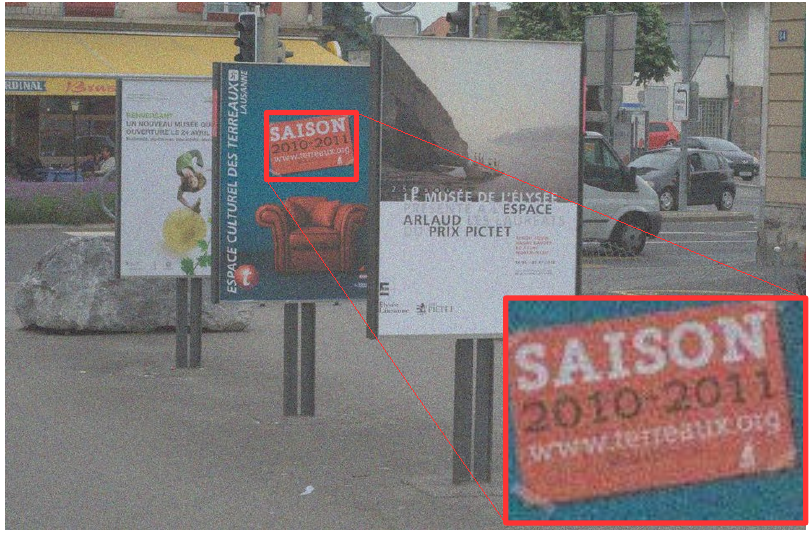}
\end{minipage}
}
\subfigure[\textbf{Ground truth}]{
\begin{minipage}{4.04cm}
\centering
\includegraphics[width=1.05\linewidth,left]{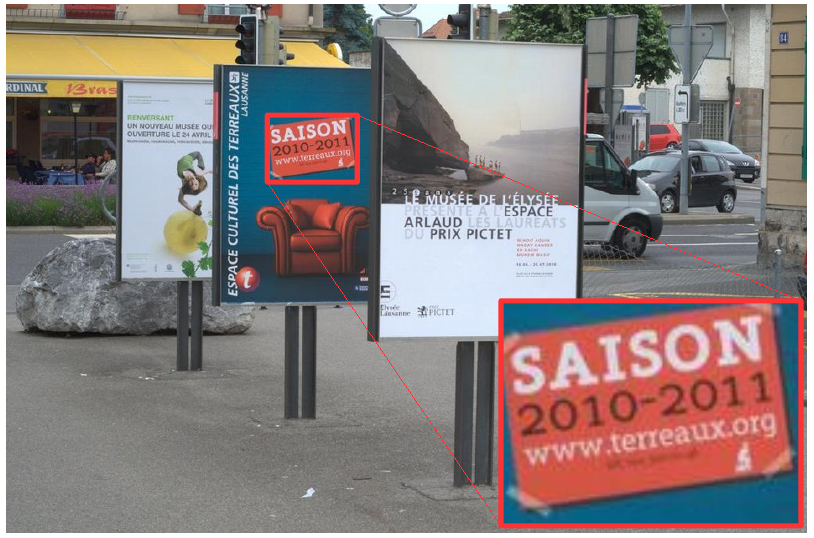}
\end{minipage}
}
\subfigure[\textbf{GF \cite{he2012guided}}]{
\begin{minipage}{4.04cm}
\centering
\includegraphics[width=1.05\linewidth,left]{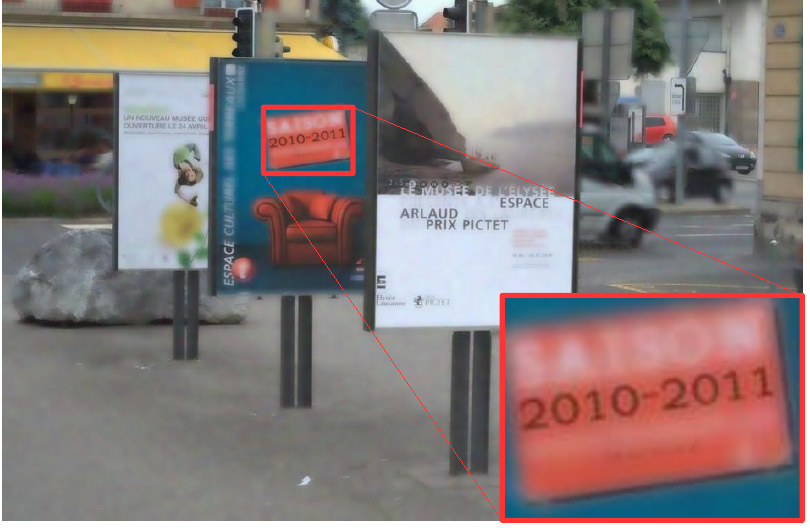}
\end{minipage}
}
\subfigure[\textbf{SVLRM \cite{pan2019spatially}}]{
\begin{minipage}{4.04cm}
\centering
\includegraphics[width=1.05\linewidth,left]{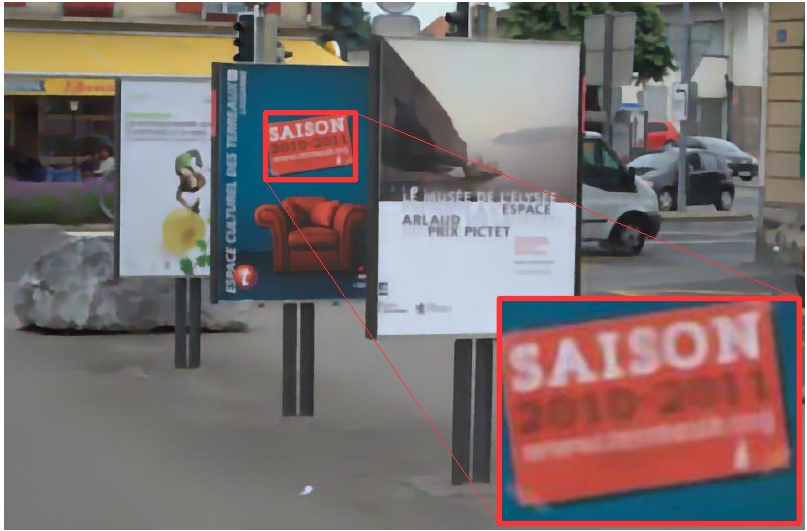}
\end{minipage}
}
\subfigure[\textbf{\textit{This paper}}]{
\begin{minipage}{4.04cm}
\centering
\includegraphics[width=1.05\linewidth,left]{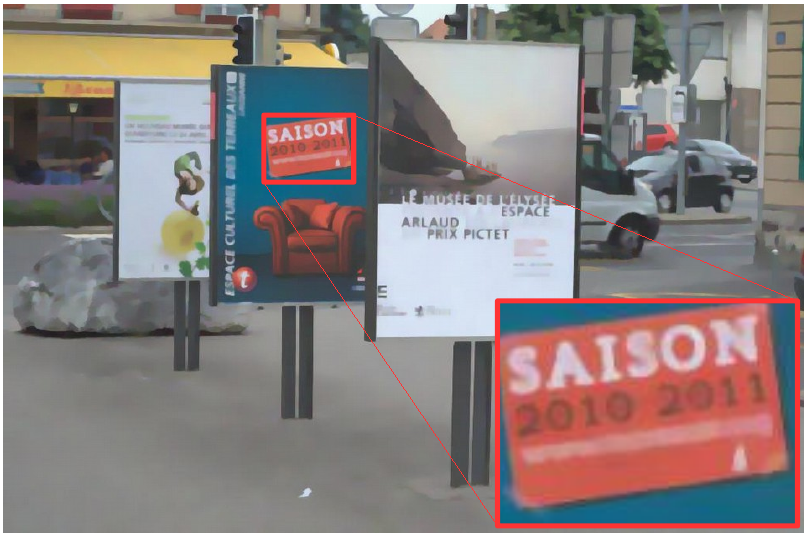}
\end{minipage}
}
\caption{\textbf{Joint denoising result} ($\sigma=25$) on RGB/NIR. Our filter better preserves finer details with fewer noises.}
\label{fig_nir}   
\end{figure}
}
%
%

%\input{conclusion}

%\appendices

%\section{}

%% use section* for acknowledgment
%\section*{Acknowledgment}


\ifCLASSOPTIONcaptionsoff
  \newpage
\fi

\bibliographystyle{IEEEtran.bst}
%\bibliography{egbib}

\end{document}